\documentclass{article}
\usepackage{arxiv}

\usepackage{url}
\usepackage{booktabs}
\usepackage{times}
\usepackage{epsfig}
\usepackage{float}
\usepackage[font=small,textfont=it]{caption}
\usepackage[toc,page]{appendix}
\usepackage{filecontents}
\usepackage{mathtools, nccmath}
\usepackage{lipsum}
\DeclareMathOperator*{\argmin}{argmin}
\DeclareMathOperator*{\argmax}{argmax}

\usepackage{microtype}
\usepackage{nicefrac}
\usepackage{amsmath}
\usepackage{amssymb}
\usepackage{textcomp}
\usepackage{pbox}
\usepackage{enumitem}
\usepackage{array}
\usepackage{bm}
\usepackage{graphicx}
\usepackage{dsfont}
\usepackage{subcaption}
\usepackage{algorithm}
\usepackage[noend]{algpseudocode}
\usepackage[export]{adjustbox}
\usepackage[perpage]{footmisc}
\usepackage{longtable} 

\usepackage[pagebackref=true,breaklinks=true,letterpaper=true,colorlinks,bookmarks=false]{hyperref}

\newcommand{\pascal}[0]{\textsc{Pascal} VOC}

\title{Straight to Shapes++: Real-time Instance Segmentation Made More Accurate}

\author{
	\begin{tabular}{c}
		\begin{tabular}{c@{\hskip 0.5cm}c@{\hskip 0.5cm}c@{\hskip 0.5cm}c@{\hskip 0.5cm}c}
			Laurynas Miksys\thanks{This work was done when L. Miksys was an intern at the Torr Vision Group, Department of Engineering Science, University of Oxford.} $^{,1}$ & Saumya Jetley$^{1,3}$ & Michael Sapienza$^2$ & Stuart Golodetz$^3$ & Philip H.\ S.\ Torr$^{1,3}$
		\end{tabular}
		\\\\
		\begin{tabular}{c}
			\normalsize $^1$ Department of Engineering Science, University of Oxford \\
			\normalsize $^2$ Samsung Research America \\
			\normalsize $^3$ FiveAI Ltd.
		\end{tabular}
		\\\\
		\begin{tabular}{c}
			{\tt\small \{sjetley,smg,phst\}@robots.ox.ac.uk;~\tt\small m.sapienza@samsung.com}
			\vspace{-.8\baselineskip}
		\end{tabular}
	\end{tabular}
}

\begin{document}
\maketitle

\begin{abstract}
Instance segmentation is an important problem in computer vision, with applications in autonomous driving, drone navigation and robotic manipulation. However, most existing methods are not real-time, complicating their deployment in time-sensitive contexts. 
In this work, we extend an existing approach to real-time instance segmentation, called `Straight to Shapes' (STS), which makes use of low-dimensional shape embedding spaces to directly regress to object shape masks. The STS model can run at $35$ FPS on a high-end desktop, but its accuracy is significantly worse than that of offline state-of-the-art methods. 
We leverage recent advances in the design and training of deep instance segmentation models to improve the performance accuracy of the STS model whilst keeping its real-time capabilities intact. 
In particular, we find that parameter sharing, more aggressive data augmentation and the use of structured loss for shape mask prediction all provide a useful boost to the network performance. 
Our proposed approach, `Straight to Shapes++', achieves a remarkable $19.7$ point improvement in mAP (at IOU of $0.5$) over the original method as evaluated on the PASCAL VOC dataset, thus redefining the accuracy frontier at real-time speeds.
Since the accuracy of instance segmentation is closely tied to that of object bounding box prediction, we also study the error profile of the latter
%as per the taxonomy proposed by Hoeim \etal 
and examine the failure modes of our method for future improvements. 

\end{abstract}

\section{\label{ch:1-intro}Introduction}

\begin{figure}[t]
  \centering
  \includegraphics[width=0.7\linewidth]{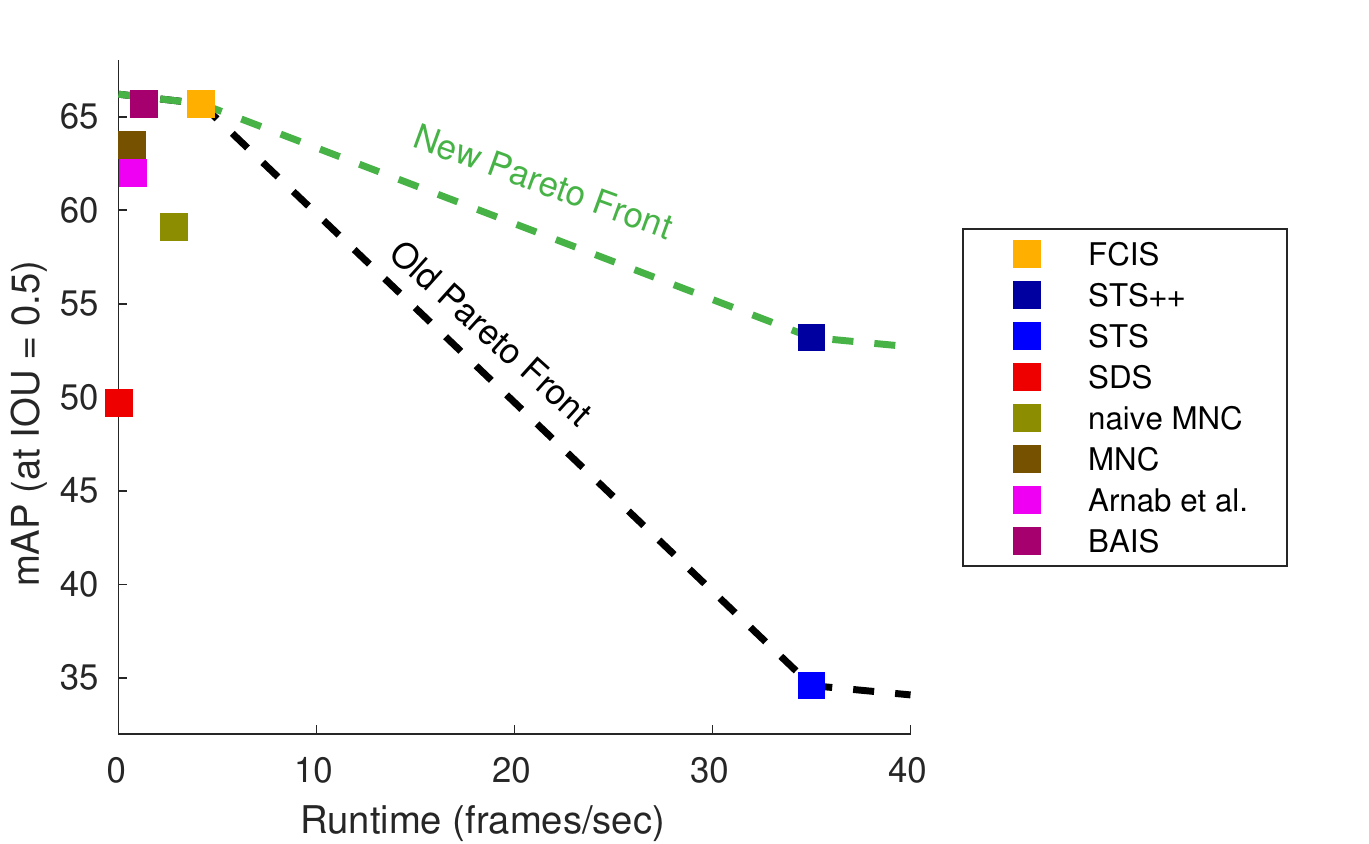}
  \caption[$mAP$ vs.\ runtime trade-off for existing instance segmentation models.]
  {\small $mAP$ vs.\ runtime trade-off as evaluated on PASCAL VOC dataset~\citep{DBLP:journals/ijcv/EveringhamEGWWZ15, Hariharan2011} across a fleet of existing instance segmentation models: FCIS~\citep{DBLP:journals/corr/LiQDJW16}, STS~\citep{sts_jetley16}, SDS~\citep{Hariharan2014}, naive MNC~\citep{DBLP:journals/corr/LiQDJW16}, MNC~\citep{dai2016instance},~\cite{2017cvpr_aarnab}, BAIS~\citep{DBLP:journals/corr/HayderHS16}. STS++ redefines the performance frontier at real-time speeds.}
  \label{fig-perf-illus}
  \vspace{-5mm}
\end{figure}

% WHAT 
% We want to perform instance segmentation. We want to perform it in real time.
Scene understanding deals with the \emph{what} and \emph{where} of objects in a given visual environment.
In cases where the scene is described using a set of RGB images, a rather elementary but challenging task of scene understanding is one of identifying and delineating instances of different categories of interest in those images. Popularly known as instance segmentation, this task has a wide applicability in real-time computer vision applications such as autonomous driving~\citep{Cordts2016Cityscapes}, drone navigation~\citep{zhuvisdrone2018} and robotic manipulation~\citep{schwarz2018rgb}, and the solution methods need to be fast as well as accurate.
%\SG{Is instance segmentation heavily used for drone navigation?}\SJ{~I'd imagine so. For purposes of surveys/counting, for action recognition, pose estimation etc for sports training and assessment the object instance (humans or any other prop) would need to be segmented out.} 
%the above has become a crucial challenge in computer vision. It has wide applicability in real-time tasks such as autonomous driving, drone navigation and robotic manipulation, and the solution methods are required to be fast alongside being accurate.
Towards this, we develop upon a recently proposed instance segmentation model that has demonstrated real-time capabilities but is limited in terms of its performance accuracy~\citep{sts_jetley16}. 
%Thus, we extend a recently proposed instance segmentation model with real-time capabilities -- the Straight to Shapes (STS) model by~\cite{sts_jetley16}. 
%\SG{We need to be careful about how we say this -- if solution methods need to be fast as well as accurate, but no methods are, then how are people tackling the applications at the moment, since they are doing so?} \SJ{Good question! I am not sure how they are tackling this problem for real world applications at the moment. I would imagine that either they are tackling it in a more constrained environment or it is still a topic of open research (check out the dataset called visDrone that has been recently proposed for advancing research in this area).}

% HOW

The Straight to Shapes (STS) model extends the real-time object detector YOLO \citep{Redmon2016} to additionally predict encoded shape representations for multiple object hypotheses.
Their model makes use of continuous shape-based representations and learns to map the input image regions to points in this continuous shape space in order to predict instance-level masks for object categories of interest.  
This use of an intermediate shape embedding space has also been shown to scale to unseen categories at test time that are similar to the training classes, a very useful property when operating in the wild.
However, while their simple extension maintains the inference speed, the inaccuracies in regressing to the real-valued vectors of shape representations result in a sub-par mAP accuracy in comparison to existing instance segmentation methods~\citep{DBLP:journals/corr/LiQDJW16, Hariharan2014, DBLP:journals/corr/LiQDJW16, dai2016instance, 2017cvpr_aarnab, DBLP:journals/corr/HayderHS16}. 
% \SG{This needs rewording a bit.} \SJ{Have made some changes now.}
% the original YOLO architecture trades accuracy for realtime performance in detection, so does the STS model on the more challenging task of instance segmentation.
More recently, \cite{DBLP:journals/corr/LiuAESR15} (with their SSD model) and~\cite{DBLP:journals/corr/RedmonF16} (with YOLO9000), have demonstrated that state-of-the-art quality can be achieved in real-time detection models by making systematic modifications to the underlying network architecture and training procedure. This suggests that similar improvements may be made in instance segmentation. Thus, we extend the existing STS model through the following contributions:

$-$ We review the recent advances in neural network design and training in the context of object detection and instance segmentation tasks.

$-$ We present a revised model called STS++ with an improved mAP accuracy on PASCAL VOC~\citep{DBLP:journals/ijcv/EveringhamEGWWZ15} at real-time speeds, as shown in Figure~\ref{fig-perf-illus}.

%revise the real-time STS model~\cite{sts_jetley16}. This results in a significant performace boost that pushes the pareto front at high runtimes placing us at a very competitive spot.
%\hspace{2mm}$\bullet$ We present a detailed analysis of all individual changes, made either to the network architecture, training setup or the loss function design, and report their comparison in terms of the contribution to the output $mAP$ score.
%\hspace{2mm}$-$ We evaluate the network for the task of zero-shot segmentation for the unseen categories from the MS-COCO~\cite{DBLP:journals/corr/LinMBHPRDZ14} dataset, and demonstrate $\cdots$.

$-$ We analyse the errors that the proposed model makes in predicting the object bounding boxes, in terms of the taxonomy proposed by~\cite{DBLP:conf/eccv/HoiemCD12}, and identify avenues for future improvements.

\iffalse
\MS{The introduction could be re-focused a little to highlight the usefulness a shape-based representation for object detection. We are proposing to move away from a discrete, predefined set of categories, and to move towards a more continuous representation which is more robust and useful, even though the performance on 'traditional cv tasks is lower'. Not all detection failures are equal. If you get inaccurate boundary of an animal, you can still run away from it. If you get results from state-of-the-art methods which fail catastrophically as soon as you test it on a slightly different image, you're dead. STS++ performance degrades gradually not catastrophically. We are trying to change the field - categorisation alone is not scalable}
-- 
\MS{Not just focus the argument on speed, we can use a different network to slow it down and increase accuracy. We should also focus on the basic philosophy and rational of the approach.}
--
\MS{Joining the two networks end-to-end, and adding DT should be included in the contribution section?}
--
\MS{add experiments on zero-shot, and coco}
--
\MS{add reference to 3D-RCNN}
\fi

\begin{figure}[ht]
\begin{minipage}{1\linewidth}
    \centering
    \includegraphics[width=0.7\linewidth]{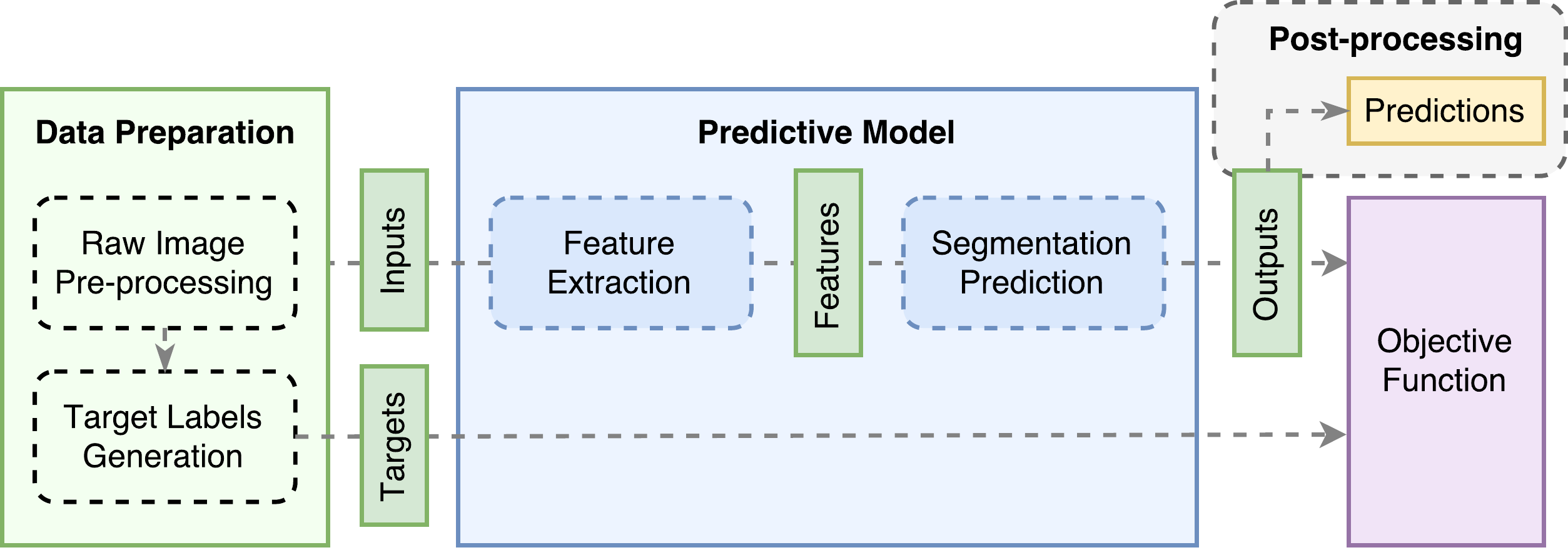}
\end{minipage}

\vspace{5mm}

\begin{minipage}{1\linewidth}
    \centering
    \includegraphics[width=0.65\linewidth]{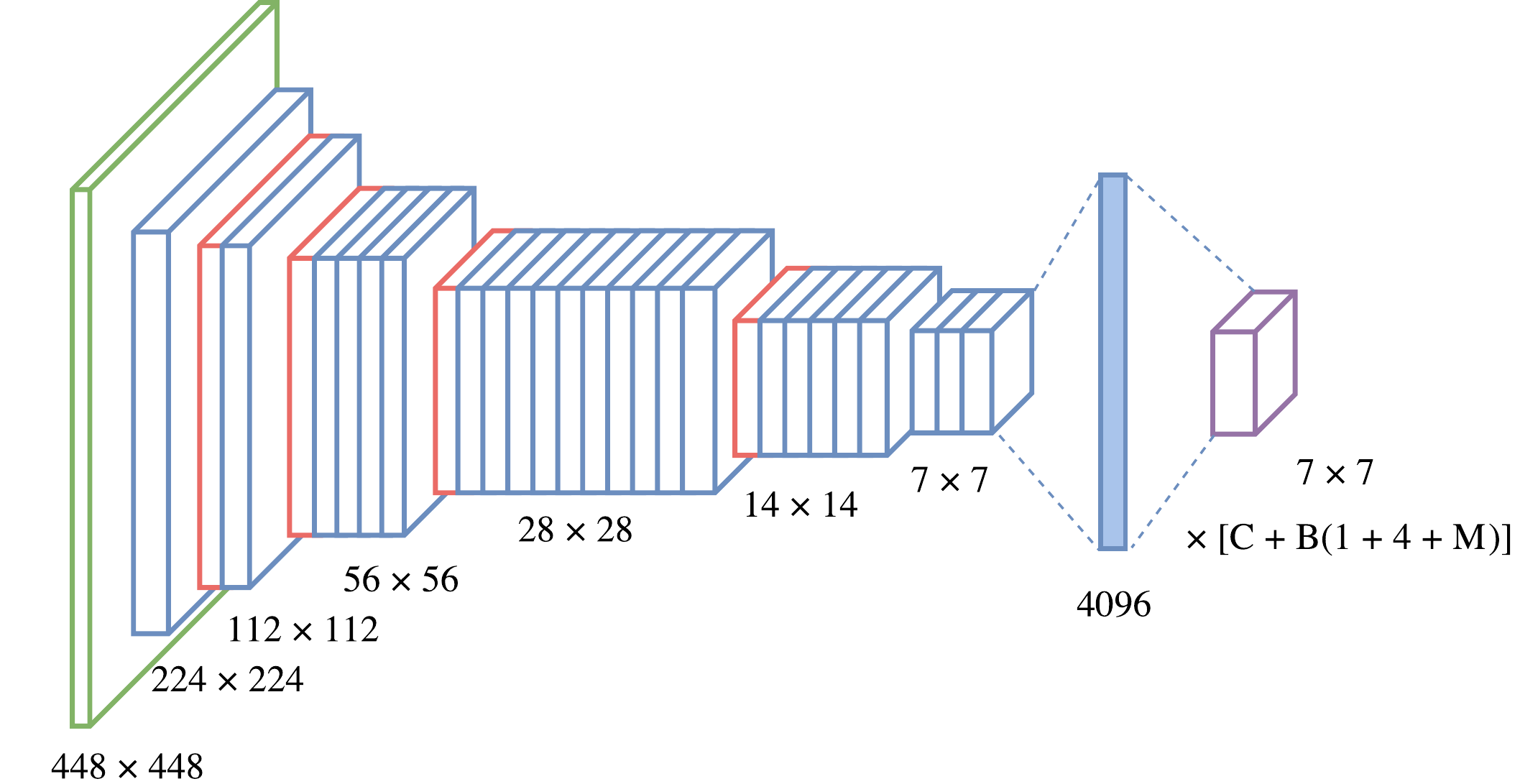}
\end{minipage}
  \caption[Straight-to-Shapes (STS) instance segmentation system.]{Schematic illustration (top) of STS~\citep{sts_jetley16} instance segmentation system, and (bottom) its predictive model. The latter is a deep neural regressor that accepts $448 \times 448$ sized RGB images as input and encodes them into a fixed $4096$-dimensional representation. This representation is then used to predict category, location and shape properties of the underlying object instances. These output properties are matched against ground truth data to train the network via backpropagation. The post-processing step merely takes the predicted shape representations which are then decoded into $2$D shape masks.}
  \label{fig-sts-orig-arc}
\end{figure}

\section{\label{ch:3-relatedwork}Related Work}
%The task of instance segmentation derives naturally from the subjacent tasks of \emph{object detection} and \emph{semantic segmentation} to yield a richer description of a scene as compared to the latter two.
Object detection localises individual objects in an image using tight-fitting bounding boxes and classifies them into one of the pre-defined target categories. However, it overlooks the pose and shape details of the objects.
Semantic segmentation yields a more fine-grained category map at the pixel-level, but does not contain knowledge about the number of individual objects or the boundaries between them when the objects of the same category are adjoining or overlapping.
Instance segmentation attempts to combine the best of both of the above tasks by delineating individual objects at the pixel level.
All the above scene understanding tasks have benefited hugely from the availability of large-scale annotated datasets and the advent of deep learning. In particular, there are many deep convolutional neural network based solutions for both object detection e.g.\ R-CNN~\citep{girshick2014rich}, Faster R-CNN~\citep{ren2015faster}, YOLO~\citep{Redmon2016} and SSD~\citep{DBLP:journals/corr/LiuAESR15} and semantic segmentation e.g.\ FCN~\citep{long2015fully} and CRF as RNN~\citep{zheng2015conditional}.
Following their success, various instance segmentation approaches~\citep{Hariharan2014, dai2015_multitaskcascade, dai2016instance, 2017cvpr_aarnab, DBLP:journals/corr/HayderHS16, DBLP:journals/corr/LiQDJW16,  DBLP:journals/corr/HeGDG17} proposed to combine both top-down detection and bottom-up segmentation results in order to isolate object instances. 

For example, the Simultaneous Detection and Segmentation (SDS) approach of~\cite{Hariharan2014} adapts the R-CNN~\citep{girshick2014rich} detector to the task of instance segmentation. This is achieved  by refining the regions within the object boxes of R-CNN 
%predicted by the Multiscale Combinatorial Grouping (MCG) method of~\cite{arbelaez2014multiscale} 
via a combination of bottom-up cues from the super-pixels of the Multiscale Combinatorial Grouping (MCG) approach of~\cite{arbelaez2014multiscale} and top-down cues from the foreground mask predicted using CNN features. In their approach, the box proposal and mask refinement stages are separately optimised. However, the Multi Network Cascade (MNC) of~\cite{dai2015_multitaskcascade} incorporates all these steps into a single neural network pipeline using a cascaded structure and joint training. A stack of convolutional layers is shared between the three cascaded tasks of bounding box proposal generation, pixel-wise prediction of an instance mask per proposal using region-of-interest (RoI) CNN features, and classification of the output mask into one of $T$ target categories. Similarly, the Boundary-Aware Instance Segmentation (BAIS) model~\citep{DBLP:journals/corr/HayderHS16} uses shared layers for region proposal and RoI feature extraction. This is followed by a pixel-wise prediction of $m$ masks, each corresponding to a specific quantised level of distance transform (DT) values. The DT values directly and redundantly encode the boundary information, which allows the reconstructed instance masks to extend beyond the bounding box and also be robust to prediction noise -- two salient features of this approach.

On the other end, Instance-sensitive FCN~\citep{dai2016instance} adapts the segmentation pipeline of FCN~\citep{long2015fully} to predict $k \times k$ \textit{relative-position} score maps. Here, each pixel value in a given score map captures the pixel's probability of being at the specified relative position w.r.t to the underlying object.
%, where the $(i,j)$ score-map captures the probability of each pixel to belong to the $(i,j)$ block of a $k \times k$ grid position on the underlying object. Thereafter,
An instance assembly module then reconstructs object masks by using dense sliding windows and copying pixel scores for a given position in an object window (say \emph{top-left}) from the associated relative position (\emph{top-left}) map. The instance mask is thus obtained by combining the masks for each relative location within a specific window.
%by copying over the content of $(i,j)$ block from the $(i,j)$ score map. 
This approach is effective at delineating instances, but doesn't associate those instances with semantic categories: it is hence often used in conjunction with R-FCN~\citep{Dai_rfcn} to classify the instance proposals. An end-to-end extension of this instance segmentation and classification pipeline is described as the Fully Convolutional Instance-aware Semantic Segmentation (FCIS) model~\citep{DBLP:journals/corr/LiQDJW16}. Unlike Instance-sensitive FCN~\citep{dai2016instance}, the position-sensitive score maps here are not predicted at the image level but atop the bounding box predictions of a region proposal network (RPN)~\citep{Dai_rfcn}.
% Once the instance masks are assembled, the corresponding CNN features are used for inferring its semantic category. 
The current state-of-the-art model, Mask R-CNN~\citep{DBLP:journals/corr/HeGDG17}, once again extends the RPN pipeline for instance segmentation. The prediction of object bounding boxes and corresponding category scores is followed by the pixel-wise prediction of $T$ binary masks, one for each target category. At test time, the binary mask for the highest-scoring semantic category is selected. The authors remark that the prediction of $T$ class-wise binary masks serves to reduce the competition between pixels to belong to a single category and improves the quality of the output masks. Concurrently, the approach of~\cite{2017cvpr_aarnab} starts from the semantic segmentation maps of CRF as RNN~\citep{zheng2015conditional} and proceeds by allocating each pixel to one of $D$ detections produced by R-FCN~\citep{Dai_rfcn}. They use the CRF framework to optimise this allocation which corresponds to the minimisation of an energy function defined atop the per-pixel semantic score, bounding box score and correlation of the semantic mask with a predefined shape prior.

Two important properties of the above methods are worth noting. Firstly, whilst the accuracy of instance segmentation methods has improved rapidly, runtime has largely remained a secondary concern. The existing best i.e. Mask R-CNN model~\citep{DBLP:journals/corr/HeGDG17} runs at $5$ fps, and cannot trivially be deployed for real-time applications (the cost can be amortised over several frames on a background thread, e.g.\ \cite{Runz2018}, but only if latency is a tolerable issue). Secondly, the bottom-up scheme of pixel-wise instance mask prediction often yields coarse and noisy masks for unseen categories, as noted by~\cite{sts_jetley16}. %The latter, similar to YOLO for the task of detection, performs three tasks in parallel - localisation (via predicting bounding box parameters), classification (via predicting category confidence scores) and instance segmentation (via predicting shape representations). 

In contrast to these methods, STS \citep{sts_jetley16} predicts shape masks via a bottleneck of an encoded, low-dimensional and continuous shape space, which allows test-time generalisation to unseen categories. Unlike all other existing approaches, which cannot run in real time, it is able to run at $35$ FPS on a high-end desktop, making it highly desirable, at least in principle, for use in real-time applications~\citep{Hicks2013}. However, its practical usefulness is compromised in practice by the accuracy of its predictions, which are significantly worse than those that can be achieved by slower approaches.
The goal of this work is to address this problem, inspired by works such as those of~\cite{DBLP:journals/corr/LiuAESR15} and~\cite{DBLP:journals/corr/RedmonF16}, which make systematic modifications to detection networks and achieve improvements in detection accuracy for real-time speeds.

\section{\label{ch:4-method}Methodology}

% What are the details of the system
% How do you proceed - details of modifications
We start by reviewing the original STS model and then discuss the motivation behind the proposed changes and the specifics of those changes.
%, grouped according to the system aspect they modify - model architecture, data preparation scheme, and post-processing mechanism. 
%Our main focus lies on improving the original model's performance while keeping the processing speed under real-time constraint.

\subsection{Original Model}
\label{original_model}
The prediction model of STS~\citep{sts_jetley16} is identical to that of YOLO~\citep{Redmon2016}.
A deep convolutional neural network (comprising 30 layers) encodes an input image of size $448\times448$ into a fixed $4096$ dimensional representation, as shown in Figure~\ref{fig-sts-orig-arc} (bottom). This encoded vector representation is then used for predicting the object instances at all image locations.
The STS model predicts object instances relative to the predefined ${S}\times{S}$ grid for $S=7$.
At every grid location, the model outputs $C=20$ conditional class probabilities and makes $B=2$ class agnostic object instance proposals.
Each proposal consists of a confidence score, center coordinates, dimensions of a bounding box, and the additional \emph{instance shape representation} (of size $M$).
More formally, for every grid cell $1\leq{i}\leq{S^2}$, the model outputs a vector
\begin{equation}
  \hat{\mathbf{y}}_{i}=
  \begin{bmatrix}
    \hat{\mathbf{p}}_{i} & \hat{\mathbf{b}}_{i1} & \dots & \hat{\mathbf{b}}_{iB}
  \end{bmatrix} \text{, where}
\end{equation}
\begin{equation}
  \hat{\mathbf{p}}_{i}=
  \begin{bmatrix}
    \hat{p}_{i1} & \hat{p}_{i2} & \dots & \hat{p}_{iC}
  \end{bmatrix}, \text{ s.t.}
\end{equation}
\begin{equation}
  \hat{p}_{ik}=\mathbb{P}[\textnormal{object of category $k$}~|~\textnormal{object in cell $i$}] \text{, and}
\end{equation}
\begin{equation}
  \hat{\mathbf{b}}_{ij}=
  \begin{bmatrix}
    \hat{c}_{ij} & \hat{x}_{ij} & \hat{y}_{ij} & \hat{\psi}_{ij} & \hat{\omega}_{ij} & \hat{\mathbf{s}}_{ij}
  \end{bmatrix} \text{,}
\end{equation}
are the parameters of the $j^th$ object proposal, where
\begin{equation}
  \hat{c}_{ij}=\mathbb{P}[\textnormal{proposal $j$ in cell $i$ is an object}]\text{,}
\end{equation}
$[\hat{x}_{ij},~\hat{y}_{ij},~\hat{\psi}_{ij},~\hat{\omega}_{ij}]$ are the bounding box parameters, and $\hat{\mathbf{s}}_{ij}=\big[\hat{s}_{ij1}, \hat{s}_{ij2}, \dots \hat{s}_{ijM}\big]$ is the $M$-dimensional shape representation.
%\begin{equation}
%  \hat{\mathbf{s}}_{ij}=
%  \begin{bmatrix}
%    \hat{s}_{ij1} & \hat{s}_{ij2} & \dots & \hat{s}_{ijM}
%  \end{bmatrix}.
%\end{equation}

Note that here $\psi_{ij},~\omega_{ij}$ correspond to the square root of the height and width of the object bounding box, respectively.
This is done to favour correct predictions of small boxes compared to larger boxes, as a small discrepancy in a large-box prediction has a smaller effect on the output accuracy.
Full details of the architecture and loss function design can be found in 
Table \ref{tbl-arc-sts-orig} of Appendix \ref{ch-architectures} and in Appendix \ref{ch-lossfunc} respectively.
%Table 4, Appendix A.1 and Appendix A.3 respectively.

The system explicitly constructs a low-dimensional shape embedding space using a denoising auto-encoder. 
At train time, the deep neural pipeline is adjusted to correctly predict the encoded shape representations $\hat{\mathbf{s}}_{ij}$ corresponding to the underlying scene objects. 
At test time, the pipeline estimates these encoded object shape representations which are then mapped to the space of $2$D binary shape masks using the standalone decoder block of the denoising auto-encoder.
Initial investigation demonstrates that the learned shape space encodes shape information in a semantically meaningful way, where instance masks of objects having similar shape appearances cluster together in the space.
It is a continuous space and allows reconstruction of new and realistic shape masks.
%Thus, at test time, this intermediate shape embedding space allows the overall model to generalise to instances from previously unseen categories.

\subsection{Proposed Revisions}

We present modifications to the existing system to speed up convergence, reduce memory requirements, minimise computational costs and improve the instance segmentation performance.
Following the block diagram of Figure~\ref{fig-sts-orig-arc} (top), we organise the proposed changes into three different groups - changes to the prediction model, the data-preparation step and post-processing operations respectively.

\subsubsection{Changes to Prediction Model}

\textbf{Batch Normalisation:}\hspace{2mm}
Batch normalisation is known to implement regularisation and improve the convergence speed during network training, often resulting in an improved performance at convergence~\citep{DBLP:journals/corr/IoffeS15}.
Its positive effect on deep neural network training is widely reported in the research literature
\citep{he2015deep, DBLP:journals/corr/HeZR015, DBLP:journals/corr/VinyalsTBE14, DBLP:journals/corr/SzegedyVISW15},
and it is demonstrated to improve the original YOLO model in the subsequent work by~\cite{DBLP:journals/corr/RedmonF16}.
We propose adding batch normalisation after every convolutional layer in the network.
Although it increases the total number of learnable parameters in the model, the batch-norm parameters can be combined with convolutional layer's affine transformation once the training is finished.
Thus, it does not affect model's computational cost or speed capabilities during inference.

\begin{figure}
  \begin{center}
    \includegraphics[width=0.7\linewidth]{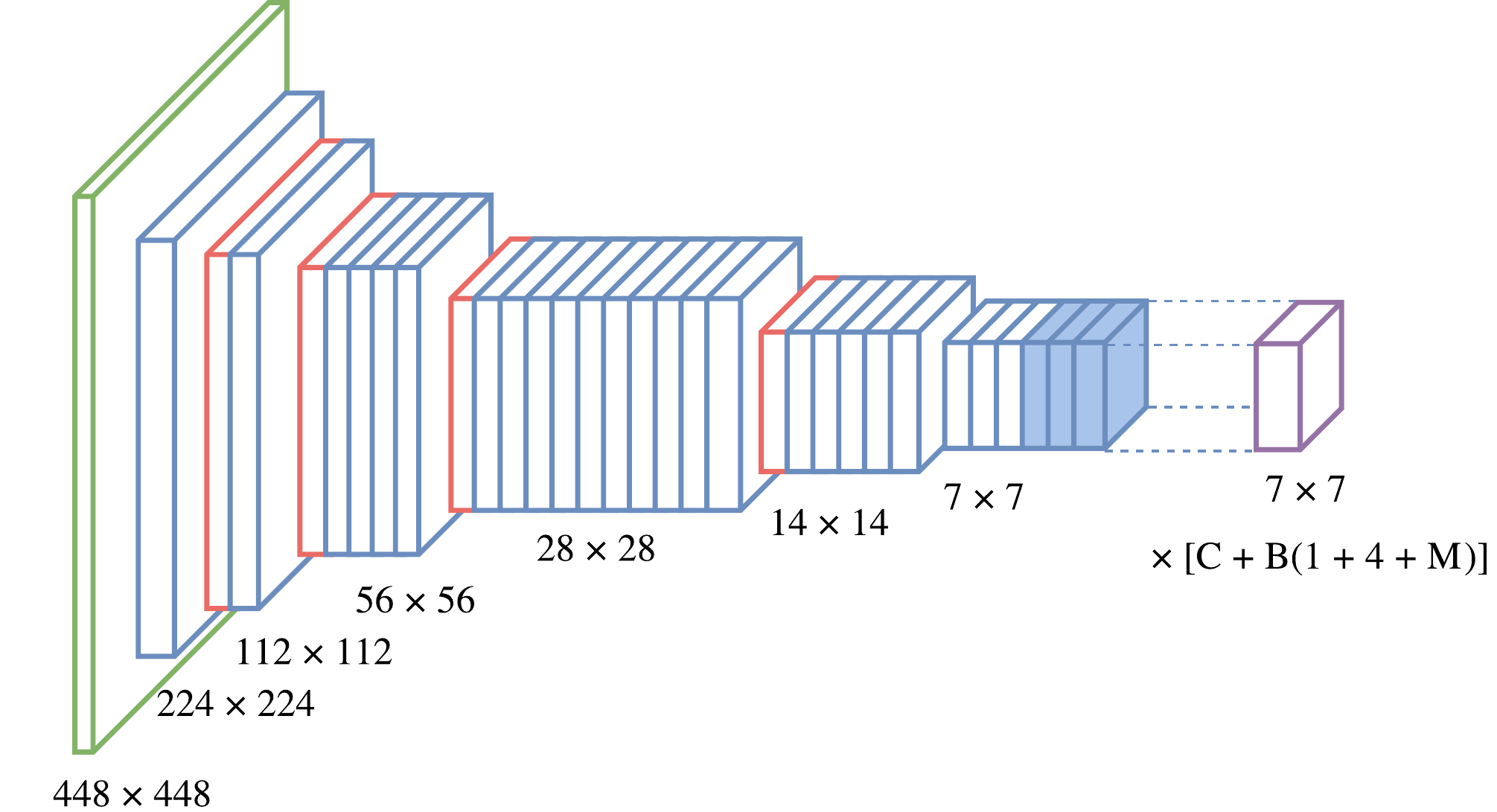}
  \end{center}
  \caption[STS prediction model with shared weights.]
  {
  Fully connected layers are replaced with convolutional layers (filled blue blocks) to build a translation invariant model.
  }
  \label{fig-arc-sts-shared}
\end{figure}

\begin{figure}[t]
  \centering
  \includegraphics[width=1\linewidth]{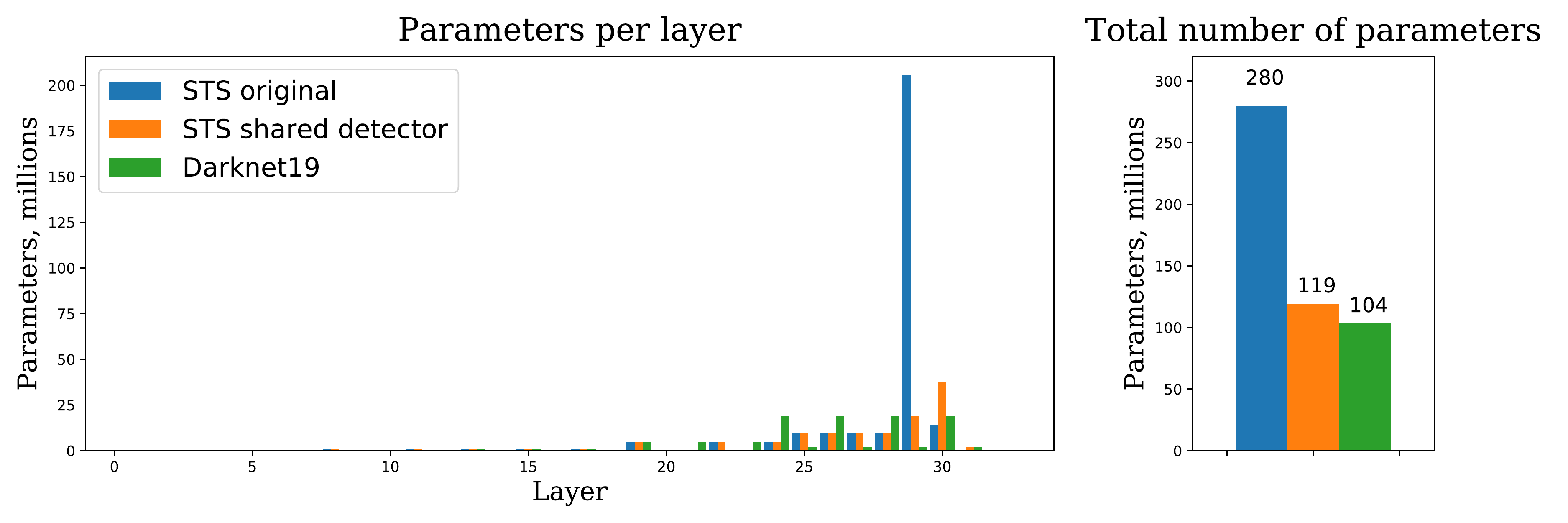}
  \caption[Number of learnable parameters in STS model.]
  {Using shared detection module at different locations significantly reduces number of learnable parameters.}
  %Darknet19 architecture further reduces parameter count and also distributes it more evenly over the layers.}
  \label{fig-params}
\end{figure}

\textbf{Sharing Prediction Weights:}\hspace{2mm}
Original model uses a series of convolutional layers followed by a fully-connected layer to construct a representation of the entire image and predict object instances at different grid locations (see Figure \ref{fig-sts-orig-arc}).
Firstly, such an architecture is not translation invariant by design, and thus has to be presented with sufficient variety of shifted examples to develop this property. % which is a norm of the real world.
Secondly, fully-connected layers contain the majority of learnable network parameters.
Thus, we propose encoding the raw image into a feature map that preserves the spatial information and uses shared convolutional layers to predict object instances at different spatial locations.
%This fully-convolutional architecture is translation invariant by design and memory efficient due to shared weights.
In order not to reduce model's capacity drastically, we introduce a sequence of $3$ convolutional layers ($3\times3\times2048$, $3\times3\times2048$ and $1\times1\times1024$), replacing a single fully-connected layer, before using the shared prediction module (see Figure \ref{fig-arc-sts-shared} for illustration).
Full details of the altered architecture can be found in Table \ref{tbl-arc-sts-shared} in Appendix \ref{ch-architectures}.
%Table 5 of Appendix A.1.
Consequently, total number of parameters is reduced from $280\times10^6$ to $119\times10^6$ (see Figure \ref{fig-params}).

\begin{figure}[t]
  \centering
  \includegraphics[width=1\linewidth]{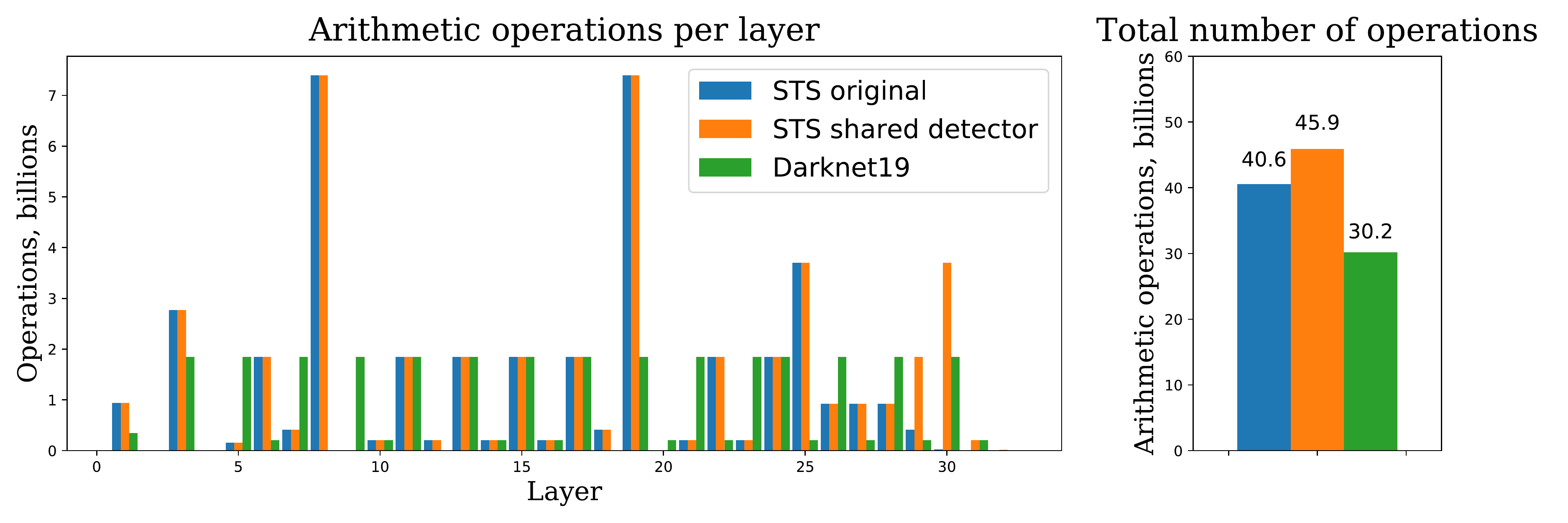}
  \caption[Number of arithmetic operations performed by different networks.]
  {Darknet19 network distributes the computations evenly throughout the network.}
  \label{fig-ops}
\end{figure}

\textbf{Distributing Computations Evenly:}\hspace{2mm}
Finally, we note that the overall computational cost is rather unevenly distributed across the individual layers of the network, see Figure \ref{fig-ops}.
This can slow down the whole network, particularly when the computational resources are limited.
In their subsequent work on YOLO,~\cite{DBLP:journals/corr/RedmonF16} identify this issue and propose the Darknet19 architecture.
They borrow network design ideas from VGG~\citep{Simonyan14c} and Network in Network (NiN)~\citep{lin2013network} architectures.
For example, the number of filters is doubled when the spatial dimension of the feature map is halved, convolutions of size $1\times1$ are used at every other layer to compress the feature representations and reduce overall computational costs.
We upgrade our model to Darknet19 architecture.
Full details of this extended architecture are described in Table \ref{tbl-arc-dn19} in Appendix \ref{ch-architectures}.
%Table 6 in Appendix A.1.

\subsubsection{Changes to Data Preparation Step}
\label{ss-dataprep}
%The data preparation step is analysed in $3$ parts. The first part concerns the raw image pre-processing, where the input frames are individually normalised in size ($448\times448$) and intensity values (scaled to range $[0,1]$), and random visual distortions are introduced.
%Thereafter, we consider the comparison of ground truth object instances to those predicted by the network. This requires establishing the target vector by encoding relevant object-instance information from the underlying image. Finally, we discuss the different shape representations that may describe the object mask in the target vector.

\textbf{Pre-processing Raw Images:}\hspace{2mm}
%Firstly, since images in the dataset come in various sizes and aspect ratios, they are normalised to a fixed $448\times448$ size by image scaling.
%Even though this introduces unwanted distortions (by scaling height and width by unequal factors), this guarantees that any image can be processed in one forward pass through the network at inference.
%Other instance segmentation approaches use image crops of predefined size for model training and a sliding-window method to process images of any size at inference \cite{DBLP:journals/corr/HeGDG17} \cite{dai2015_multitaskcascade}.
Neural networks used on computer vision tasks usually have their parameter count far exceeding the number of data points available for training~\citep{NIPS2012_alexnet}.
Random data distortions are known to help prevent over-fitting and improve overall performance~\citep{DBLP:journals/corr/abs-1003-0358, DBLP:journals/corr/WongGSM16}.
Given also our focus on detecting objects in natural environments, we want to simulate various pose, angle, and lighting conditions via suitable data distortions.
This can be achieved, to a limited extent, using affine transformations of 2D images and by performing per-pixel colour distortions.
We propose to change the data-augmentation paradigm (often favouring more aggressive transformations) as follows:

\hspace{2mm}\emph{$\bullet$ rotation angle}: uniformly from $[-4^{\circ},4^{\circ}]\to[-20^{\circ},20^{\circ}]$

\hspace{2mm}\emph{$\bullet$ translation in $x$ and $y$}: uniformly from $[-0.2,0.2]\to[-0.15,0.15]$

\hspace{2mm}\emph{$\bullet$ scaling factor}: uniformly from $[0.97,1.03]\to$ log-uniformly from $[\frac{1}{1.2},1.2]$

\hspace{2mm}\emph{$\bullet$ random horizontal flip}: from $Bernoulli(0.5)$ (\emph{unchanged})

\hspace{2mm}\emph{$\bullet$ scale intensity value}: uniform from $[0.8,1.2]\to$ log-uniform from $[\frac{1}{1.2},1.2]$

\hspace{2mm}\emph{$\bullet$ additive intensity value}: uniform from $[-10,10]$ (\emph{unchanged})
%\begin{itemize}
%  \item \emph{rotation angle}: uniformly from $[-4^{\circ},4^{\circ}]\to[-20^{\circ},20^{\circ}]$
%  \item \emph{translation in $x$ and $y$}: uniformly from $[-0.2,0.2]\to[-0.15,0.15]$
%  \item \emph{scaling factor}: uniformly from $[0.97,1.03]\to$ log-uniformly from $[\frac{1}{1.2},1.2]$
%  \item \emph{random horizontal flip}: from $Bernoulli(0.5)$ (\emph{unchanged})
%  \item \emph{scale intensity value}: uniform from $[0.8,1.2]\to$ log-uniform from $[\frac{1}{1.2},1.2]$
%  \item \emph{additive intensity value}: uniform from $[-10,10]$ (\emph{unchanged})
%\end{itemize}
%Instance segmentation system is expected to be robust to pose, viewpoint and lighting variations in order to work well in practice, thus, adequate examples must be presented at training.

\textbf{Representing Targets:}\hspace{2mm}
STS~\citep{sts_jetley16} uses the target representation approach similar to YOLO~\citep{DBLP:journals/corr/RedmonF16} and SSD~\citep{DBLP:journals/corr/LiuAESR15}.
A frame is divided into $S \times S$ grid cells.
A cell containing the center of a ground truth bounding box gets assigned all relevant information about the ground truth instance, provided as a vector of size $4+C+M$, as described in \S\ref{original_model}.
A nuance of this assignment, however, is that any grid cell is capable of holding information of at most one ground truth object.
Whenever another object in the frame has its bounding box centered in the cell $i$, it is discarded.
This leads to the model being penalised for predicting such objects, since they are not represented anywhere in the target.
%\begin{figure}
%  \begin{center}
%    \includegraphics[width=0.3\linewidth]{figures/cells}\qquad
%    \includegraphics[width=0.3\linewidth]{figures/anchors}
%  \end{center}
%  \caption[Schematic illustration for generating target data]
%  {
%  (LEFT) Grid cell (red) containing center of the object (blue) is responsible for predicting object's bounding box (green).
%  Center coordinates are given relative to that grid cell, while dimensions are relative to the whole image.
%  (RIGHT) Anchors boxes of 3 different aspect ratios get assigned objects with highest IOU.
%  Distinct anchors specialise on objects of different aspect ratio.
%  Illustration displays several of these boxes at different locations in the image.
%  }
%  \label{fig-cells}
%\end{figure}

%Figure \ref{fig-cells} (LEFT) illustrates how bounding box parameters are assigned.
%The problem with such assignment is that any grid cell is capable of holding information of at most one ground truth object.

We propose changing such a target encoding, and employing \emph{anchor boxes} - a technique used in other detection and segmentation methods such as
Faster R-CNN~\citep{ren2015faster},
Mask R-CNN~\citep{DBLP:journals/corr/HeGDG17},
MNC~\citep{dai2015_multitaskcascade}.
We use 3 such boxes per grid cell corresponding to 3 different aspect ratios, namely, $1\times1$, $1\times2$ and $2\times1$. This allows the object instances of non-standard aspect ratios to find their best match, given that we assign an instance to an anchor box when their IOU is the highest amongst other pairs. In addition, the system can now support the prediction of up to 3 closely placed ground truth objects per location. 
%Thus, in the new proposed scenario up to three objects per location can be represented in the target vector.
% Anchors with different prior aspect ratios specialise in predicting objects of different aspect ratios to their priors.
%Figure \ref{fig-cells}  illustrates (LEFT) the original assignment of a ground truth object to a given grid cell, and (RIGHT) some anchor boxes at random locations in the output grid.

\textbf{Representing Shapes:}\hspace{2mm}
\label{ss-representingshapes}
Several different methods to represent shapes are investigated in the STS paper, namely, down-sampled binary mask representation, radial contour description, and an embedding space learned using a parametric model (in particular, a denoising auto-encoder~\citep{vincent2008extracting}).
We use the learned shape embedding in all of our experiments, due to its low dimensionality, better reconstruction accuracy and robustness to noise.
% For this, a neural convolutional auto-encoder is trained on a separate Caltech-101 silhouettes dataset.
% It is designed to compress original binary masks into a low-dimensional space (20 dimensions) via a sequence of convolutional and max-pooling layers, and to reconstruct them using the reverse set of operations.% with sequence of convolutional and spatial up-sampling layers.

%\paragraph{Distance Transform Shape Representation}

As an alternative, we also propose and investigate the use of the \emph{distance transform} (DT) based representation of binary images~\citep{DBLP:journals/cvgip/Borgefors86}.
DT encodes information in a multi-valued format where each pixel value represents its closest distance to the background (w.r.t. some metric $d$, here we use the Euclidean $l_2$-distance).
This lends DT based representations with a richer structure.
Moreover, a corrupted pixel value in DT representation can be recovered given the surrounding pixels values.
This property provides a more robust shape reconstruction in the presence of prediction noise.

\subsubsection{Changes to Post-processing Step}

%The neural network model described in the sections above predicts decodable instance-proposal parameters at every location of the grid.
%The proposals with too small confidence scores are discarded.
%Remaining are reconstructed from their predicted representations, and finally the overlapping object boxes/masks are discarded.

\textbf{Mask Decoding:}\hspace{2mm}
In order to evaluate the quality of proposed instances, binary masks of size $m\times{m}$ ($m=64$) are reconstructed from their predicted shape representations and re-scaled to fit their corresponding bounding box predictions.
%Whenever a down-sampled binary mask or radial descriptor is modelled, the shape is decoded with respective analytic inverse function.
In the case of learned shape embeddings, a trained neural network decoder is used to reconstruct the mask from the predicted embeddings.
Detailed architecture of the decoder can be found in the Table \ref{tbl-arc-decoder-orig} in Appendix \ref{ch-architectures}. %Table 7 in Appendix A.1.
It is trained separately as part of an auto-encoder pipeline using the per-pixel binary cross-entropy loss:
\begin{equation}\label{eq-bce-decoder}
  BCE=\sum_{k=1}^{m^2}{-{t_k}\log{\hat{t}_k}-(1-t_k)\log{(1-\hat{t}_k)}} \text{,}
\end{equation}
where $t_k$ and $\hat{t}_k$ are ground truth and predicted pixel intensities, respectively.

As one of the experiments, we propose incorporating the shape decoder into the full end-to-end trainable pipeline. This simplifies the overall approach and allows the rich semantic structure of the larger PASCAL VOC dataset to tune the reconstruction process.
%as the shape decoder is trained together with the feature extractor in an end-to-end fashion, and has the rich semantic structure of the large dataset to learn from. Forcing the model to reconstruct shape from a small dimensional embedding holds the same promise of the shape embedding space with rich semantic structure.
The original decoder has relatively small number of learnable parameters ($88$ thousand) and uses hard-coded bi-linear spatial up-sampling, which limits the shape decoder's learning ability.
Hence, we also propose an alternative decoder architecture with increased learning capacity and learnable up-scaling function (via transposed convolutions).
The number of layers and output mask dimensions remains the same, while we increase the number of filters per layer.
More details can be found in the Table \ref{tbl-arc-decoder-prop} of Appendix \ref{ch-architectures}.
%Table 8 of Appendix A.1.
The squared error term in the original loss function design (Eq.~\ref{eq-loss-shape}, Appendix~\ref{ch-lossfunc}), 
%(Eq.~11, Appendix~A.3), 
to regress to the encoded shape representations, is replaced with binary cross-entropy (Eq. \ref{eq-bce-decoder}) during the holistic training of the full pipeline.

\begin{figure}[t]
\centering
\resizebox{0.8\linewidth}{!}{
  \begin{minipage}{0.3\linewidth}
    \centering
    \includegraphics[width=0.8\textwidth]{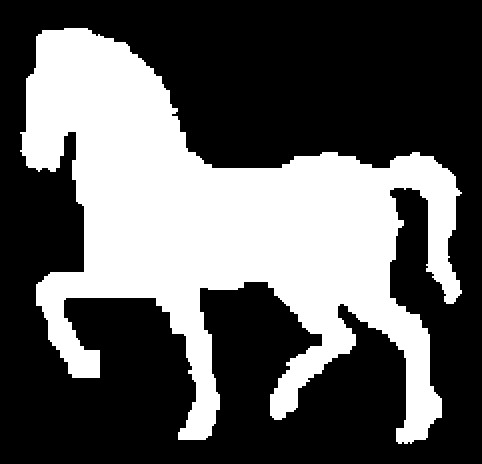}
  \end{minipage}
  \begin{minipage}{0.3\linewidth}
      \centering
    \includegraphics[width=0.8\textwidth]{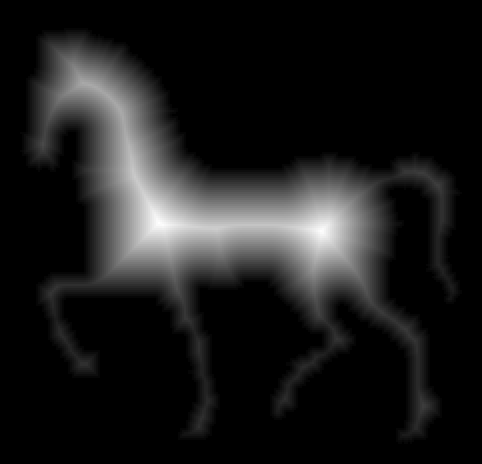}
  \end{minipage}
  \begin{minipage}{0.3\linewidth}
      \centering
    \includegraphics[width=0.8\textwidth]{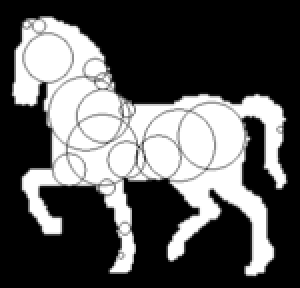}
  \end{minipage}
  }
  \caption[Distance transform based representation and reconstruction.]
  {(Left) Object instance binary mask; (Middle) Distance transform based representation of the binary mask, (Right) Illustration of the process of reconstruction of the binary mask from its distance transform based representation. This is done by superimposing discs, at every pixel position, of radii equal to the underlying distance values. (Images taken from \url{wolfram.com})}
  \label{fig-dt-examples}
  \vspace{-4mm}
\end{figure}

\textbf{Decoding the Distance Transform (DT):}\hspace{2mm}
In order to accommodate DT based shape representations, we implement another differentiable decoder for purposes of reconstruction.
%The unit generates $l$ distinct masks corresponding to the different quantised distances to the background, and deconvolves and combines them to output a final binary shape mask.
Given a low dimensional representation for every instance, a neural decoder, designed as a sequence of transposed and ordinary convolutional layers, generates $l$ binary masks of size ${m}\times{m}$.
Every such mask encodes a specific quantised value representing distance to the background.
In particular, the mask $r\in\{1,\dots,l\}$ has value 1 where the distance is at least $r$, and 0 elsewhere. 
%Thus, each pixel within the object instance boundary must have $1$ value for only a single binary mask and $0$ for all else.
Every pixel in every mask is modelled as a binary variable using logistic units.
%For the mask $l$, it would predict value $1$ whenever distance is at least $l$ at that location.
Once such masks are generated a transposed convolutional (deconvolutional) layer is applied with pre-defined non-learnable filters, where every such filter encodes a disk of radius $r\in\{1,\dots,l\}$.
%(see Figure \ref{fig-dt-rec}).
Essentially, a disk of radius $r$ is drawn at the location where encoded distance value is $r$ (see Figure \ref{fig-dt-examples} for illustration). Noticeably, each pixel value contains information about the distance to the object boundary. This redundancy equips the DT based shape representation to better handle any noise at inference.
%Notice that such disk can always be drawn inside the segment by definition.
Final binary mask is obtained by taking a linear combination of such reconstructed masks and thresholding the output, where linear parameters are learned to optimise the objective.
The whole decoder is fully differentiable and can be learned together with the full network. For more details refer to the Table \ref{tbl-arc-decoder-prop-dt} in Appendix \ref{ch-architectures}. %Table 9 of Appendix A.1. 
We use $l=8$ for our experiments. This is similar to the way that binary masks are reconstructed in the BAIS model by~\cite{DBLP:journals/corr/HayderHS16}.
Finally, given the reconstructed binary masks of valid object proposals, non-maximal suppression is performed to filter the overlapping predictions~\citep{DBLP:conf/icpr/NeubeckG06}.

%\begin{figure}
%  % \includegraphics[width=\textwidth]{figures/dtreconstruction}
%  \includegraphics[width=0.95\linewidth]{figures/dtreconstruction-alt}
%  \caption[Filters for binary mask reconstruction from distance transform]
%  {
%  Transposed convolution filters used for approximate binary mask reconstruction from quantised distance transform values.
%  }
%  \label{fig-dt-rec}
%\end{figure}

%This is a standard technique performed in most object detection and segmentation approaches \cite{Hariharan2014} \cite{DBLP:journals/corr/HeGDG17} \cite{DBLP:journals/corr/HayderHS16}.
%In particular, given two predicted instance masks with IOU greater that the stipulated value ($0.5$, $0.7$ or $0.9$), instance with the lower (non-maximum) confidence score is discarded.

\section{\label{ch:5-experiments}Experiments and Results}

\begin{figure}[t]
  \begin{minipage}{0.49\textwidth}
    \begin{center}
      \includegraphics[width=1\textwidth]{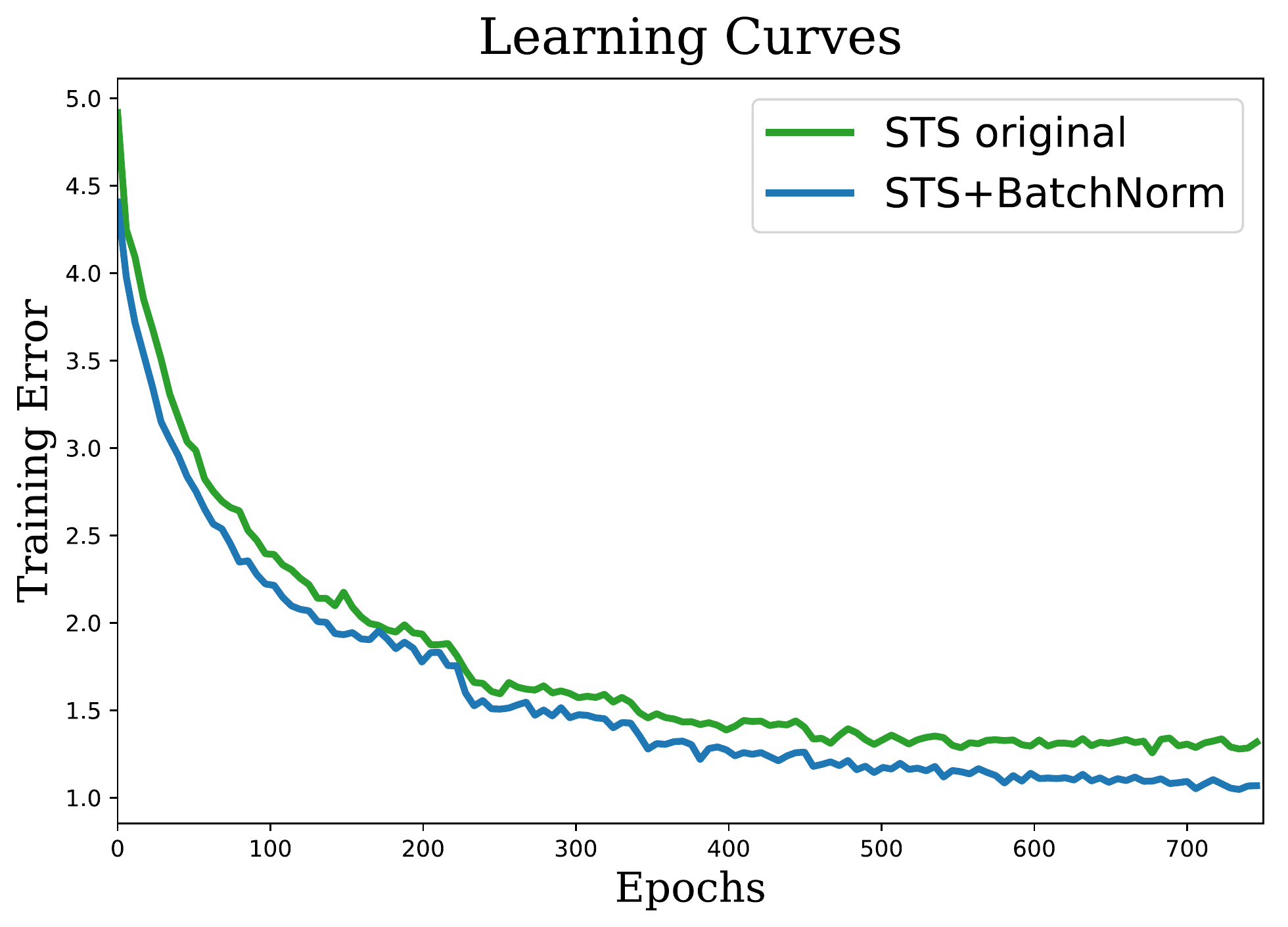}
    \end{center}
    %\caption{Learning curves during training with/without Batch Normalisation.}
    %\label{fig-exp-bn}
  \end{minipage}
  \begin{minipage}{0.49\textwidth}
    \begin{center}
      \includegraphics[width=1\textwidth]{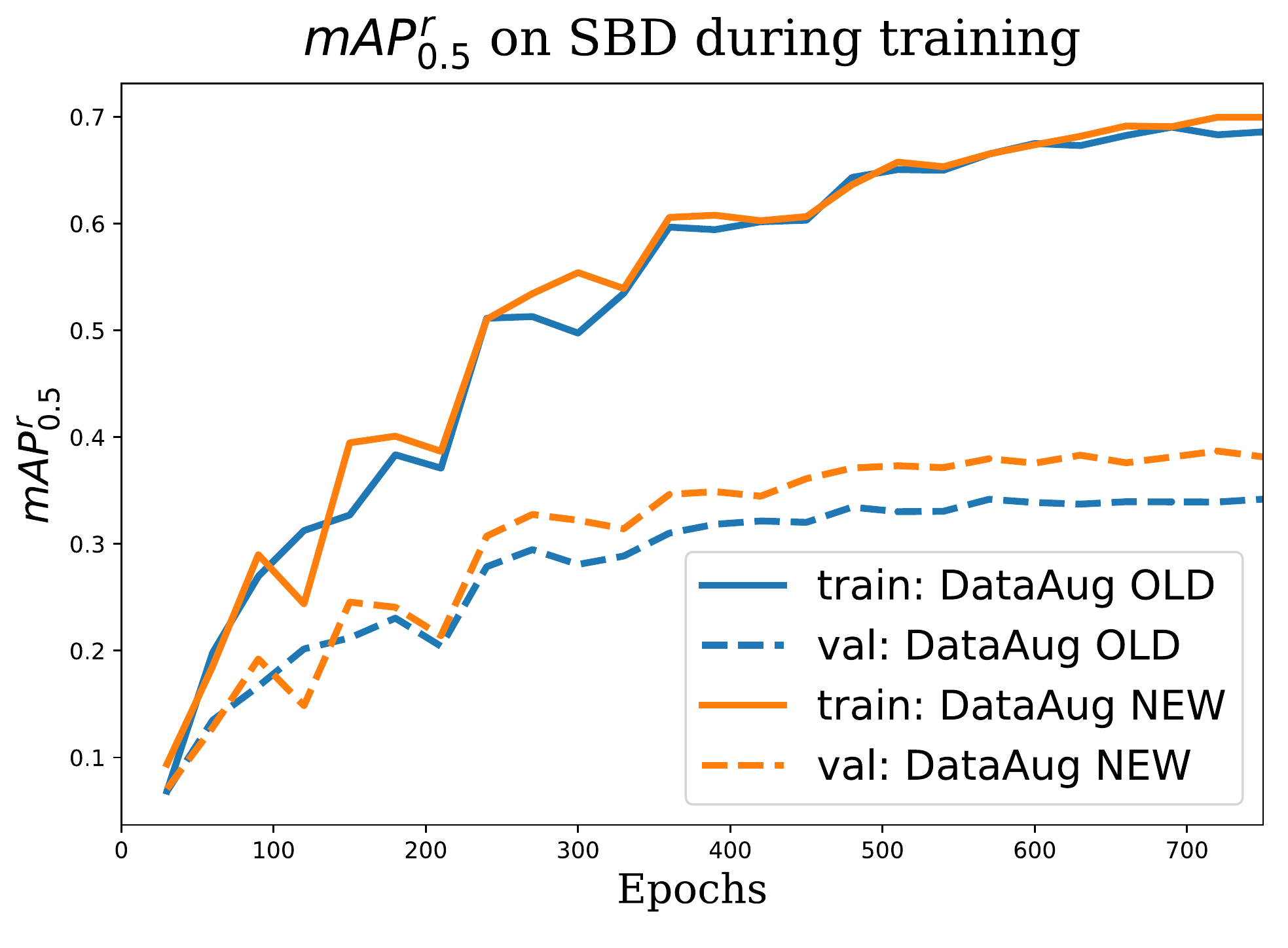}
    \end{center}
    %\caption[Experimental results when training with batch normalisation and aggressive data augmentation]
    %{$mAP^r_{0.5}$ estimates on SBD train and val splits during training with conservative vs. aggressive data augmentation.}
    %\label{fig-exp-da}
  \end{minipage}
  \caption[$mAP^r$ scores on SBD with batch normalisation and data augmentation.]{(Left) Learning curves during training with and without batch normalisation. (Right) The $mAP^r_{0.5}$ estimates on SBD train and val splits during training with conservative vs. aggressive data augmentation.}
  \label{fig-exp-bn-da}
\end{figure}

The details of the experimental setup are provided in Appendix~\ref{ch-experimentalsetup}.
%Appendix~A.2.
Here, we present a quantitative analysis of the effects of the proposed changes on the instance segmentation performance.
For purposes of comparison, we train our models on the SBD train set and evaluate on the SBD validation set~\citep{Hariharan2011}.
%We also report zero-shot segmentation capabilities of the revised architecture.
%Since our approach depends highly on the bounding box predictions, 
We further evaluate the models on the task of object detection alone and analyse the results in terms of the error taxonomy proposed by~\cite{DBLP:conf/eccv/HoiemCD12}.

%\begin{figure}
%\centering
%  \begin{minipage}{0.28\textwidth}
%   \includegraphics[width=1\textwidth]{figures/map-start-point}
%  \end{minipage}\hspace{0.06\textwidth}
%  \begin{minipage}{0.4\textwidth}
%    \resizebox{\linewidth}{!}{
%    \begin{tabular}{l|ccc}
%    	Model & $mAP^r_{0.5}$ & $mAP^r_{0.7}$ & $mAP^r_{vol}$ \\\hline
%    	STS Binary Mask 16x16 & 32.3 & 12.0 & 28.6 \\
%    	STS Radial 256 & 30.0 &  6.5 & 29.0 \\
%    	STS Embedding 50 & 31.1 & 14.1 & 27.9 \\
%    	STS Embedding 20 & 33.5 & 14.9 & 30.1 \\
%    	\hline
%    	%SDS & 49.7 & 25.3 & 41.4 \\
%    	%MNC & 63.5 & 41.5 & - \\
%    	%Arnab & 62.0 & 44.8 & 55.4 \\
%    	%BAIS & 65.7 & 48.3 & - \\\hline
%    \end{tabular}
%    }
%  \end{minipage}
%  \caption[Evaluation results of STS model on SBD val dataset]
%  {$mAP^r$ estimate comparison for STS~\cite{sts_jetley16}, SDS~\cite{Hariharan2014}, MNC~\cite{dai2016instance}, Arnab et al.~\cite{2017cvpr_aarnab} and BAIS~\cite{DBLP:journals/corr/HayderHS16} on SBD val dataset. STS performs best with 20-dimensional learned shape embedding.}
%  \label{fig-exp-start}
%\end{figure}
%\fi

\subsection{Evaluating Instance Segmentation Performance}

We start with the original STS approach that yields top performance using the 20-dimensional learned shape representations and study the effects of the proposed changes that are introduced incrementally. %, as shown in Figure~~\ref{fig-exp-start}. 
%We then introduce incremental changes to the pipeline and report the resulting performance scores for the task of instance segmentation.
%Initial model scores $33.5\%$ $mAP^r_{0.5}$ and $14.9\%$ $mAP^r_{0.7}$ with 20-dimensional shape embeddings and is the best reported performance of the approach (see Figure \ref{fig-exp-start} for STS results with different embeddings).

\textbf{Training with Batch Normalisation:}\hspace{2mm}
As expected and noted in the literature~\citep{DBLP:journals/corr/IoffeS15} batch normalisation improves the convergence speed as well as the performance at convergence, as can be seen in Figure \ref{fig-exp-bn-da}. On the validation set, it boosts the performance by $5.1\%$ ($mAP^r_{0.5}$).
As mentioned before, the processing speed at the time of inference is not affected.

\textbf{Augmenting Training Dataset:}\hspace{2mm}
Presenting the network with more aggressively augmented examples improves the model's generalisation.
As seen in Figure \ref{fig-exp-bn-da}, the model has sufficient capacity to fit the increased variability in the training examples, and at the same time it helps to generalise better on unseen validation samples.
Building on the previous change of batch normalisation, data augmentation improves our model by an additional $3.7\%$ $mAP^r_{0.5}$.

\begin{figure}[t]
	\centering
	\includegraphics[width=0.55\textwidth]{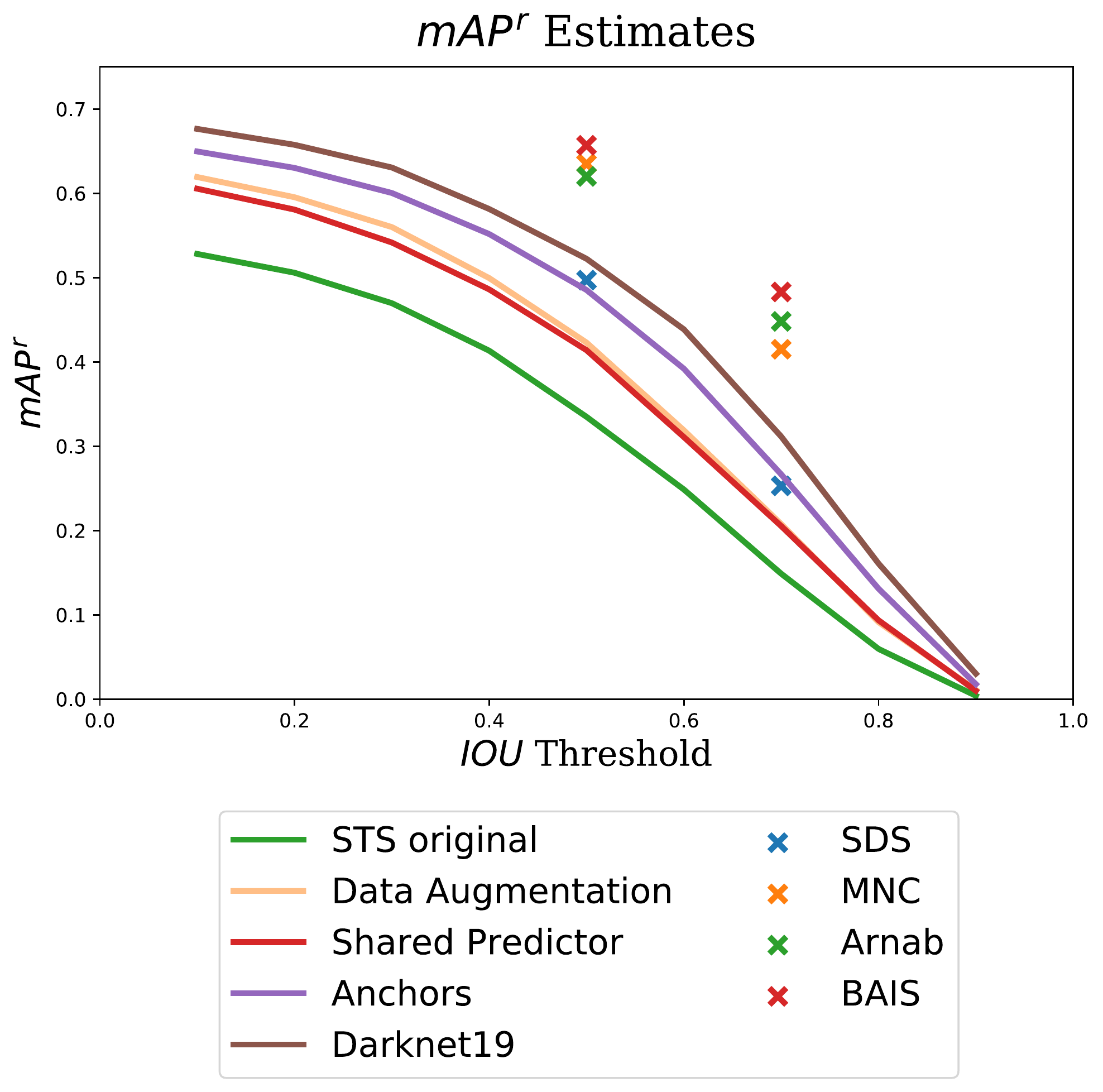}
%\caption[$mAP^r$ scores with weight sharing, anchor boxes and Darknet19.]{$mAP^r$ estimates on SBD validation set for STS models incorporating changes to the network architecture and the target representation. These results are compared here with those for SDS~\citep{Hariharan2014}, MNC~\citep{dai2016instance},~\cite{2017cvpr_aarnab} and BAIS~\citep{DBLP:journals/corr/HayderHS16}.}
%\label{fig-exp-map-conv-anc-dn19}
%\begin{table}
%\end{figure}
%\begin{table}[t]
    \resizebox{0.9\linewidth}{!}{
	\vspace{1cm}
    \begin{tabular}{l|ccccccccc|c}
    	 & \multicolumn{9}{c|}{$mAP^r_{iou}$} & $mAP^r_{vol}$ \\
    	IoU threshold & 0.1 & 0.2 & 0.3 & 0.4 & 0.5 & 0.6 & 0.7 & 0.8 & 0.9 & \\\hline
    	STS*  & 52.8 & 50.6 & 47.0 & 41.3 & 33.5 & 24.9 & 14.9 &  6.0 &  0.4 & 30.1 \\
    	Darknet19  & 67.6 & 65.8 & 63.1 & 58.1 & 52.2 & 43.9 & 31.1 & 16.1 &  3.0 & 44.5 \\\hline\hline
    	End-to-End  & 68.2 & 66.2 & 63.4 & 59.0 & 51.7 & 41.7 & 27.9 & 12.8 &  1.9 & 43.6 \\
    	Large Decoder  & 67.8 & 65.9 & 63.1 & 58.7 & 52.3 & 43.8 & 31.9 & \textbf{17.3} & \textbf{ 3.9} & 45.0 \\\hline\hline
    	Distance Transform (STS++)  & \textbf{68.2} & \textbf{66.3} & \textbf{63.6} & \textbf{59.4} & \textbf{53.2} & \textbf{45.2} & \textbf{32.3} & 16.9 &  3.4 & 45.4 \\
    \end{tabular}
    }
  \caption[$mAP^r$ scores on SBD val for proposed modifications.]
  {${mAP}^{r}$ estimates on SBD  validation set for the proposed models incorporating changes to the network architecture and target representations. (Top) $mAP^r$ scores for models with weight sharing, anchor boxes and Darknet19 are compared here with those for SDS~\citep{Hariharan2014}, MNC~\citep{dai2016instance},~\cite{2017cvpr_aarnab} and BAIS~\citep{DBLP:journals/corr/HayderHS16}. (Bottom) Distance transform based shape reconstruction offers a marginal advantage over both pre-trained and end-to-end trained decoders for reconstructing learned shape representations. Note: The model marked with a * has been retrained for this work.}
  \label{tbl-exp-e2e-dt}
%\end{table}
\end{figure}

\textbf{Sharing Prediction Weights:}\hspace{2mm}
This change in architecture, from fully-connected to convolutional, introduces translation invariance into the model and reduces the total number of parameters.
This has a small negative effect on accuracy decreasing it by $0.9\%$ $mAP^r_{0.5}$, see Figure \ref{tbl-exp-e2e-dt}~(top).

\textbf{Anchors as Bounding-box Priors:}\hspace{2mm}
This impacts how adjacent or possibly overlapping object instances are presented to the model. %increases the resolution at which ground truth instances 
In the previous setup, the model was discouraged from predicting objects appearing close to each other in the image (by selecting only the ground truth bounding box with the highest IoU when multiple such boxes were centered in the same grid cell). In comparison, our approach provides three anchor boxes per grid cell location with which to match the ground truth instance hypotheses. 
This leads to a significant $7.1\%$ improvement in $mAP^r_{0.5}$ as can be seen in the performance plots of Figure \ref{tbl-exp-e2e-dt}.
%This raises a question, that of designing the most efficient way of representing ground truth instances and its effect on overall pipeline performance.
%More work needs to be done on this topic.

\textbf{Distributing Computations Evenly:}\hspace{2mm}
These architectural changes not only reduce the parameter count and the total computational effort, but also result in a more robust model, offering an overall performance gain of $3.7\%$ in terms of $mAP^r_{0.5}$, refer to Figure \ref{tbl-exp-e2e-dt}~(top).

\begin{table*}[t]
    \begin{center}
    \resizebox{1\linewidth}{!}{
    \begin{tabular}{l|ccccccccc|c|c|c|c|c|c|}
    	 & \pbox{0.08\linewidth}{STS\\~\citep{sts_jetley16}} & & & & & & & & STS++ & \pbox{0.075\linewidth}{SDS\\~\citep{Hariharan2014}} & \pbox{0.06\linewidth}{naive MNC\\~\citep{DBLP:journals/corr/LiQDJW16}} & \pbox{0.06\linewidth}{MNC\\~\citep{dai2016instance}} & \pbox{0.06\linewidth}{\cite{2017cvpr_aarnab}} & \pbox{0.07\linewidth}{BAIS\\~\citep{DBLP:journals/corr/HayderHS16}} & \pbox{0.06\linewidth}{FCIS\\~\citep{DBLP:journals/corr/LiQDJW16}}\\
 & &  &  &  &  &  & & & & & & &  & &\\
 & (darknet) &  &  &  &  &  & & & & (alexnet) & (resnet-101) & (vgg-16) &  & & (resnet-101)\\
\hline
    	 Learned Embedding (20) & + & + & + & + & + & + & & & & & & & & & \\
    	 Batch Normalisation & & + & + & + & + & + & + & + & + & & & & & & \\
    	 Data Augmentation & & & + & + & + & + & + & + & + & & & & & & \\
    	 Shared Predictor & & & & + & + & + & + & + & + & & & & & & \\
    	 Anchor Boxes & & & & & + & + & + & + & + & & & & & & \\
    	 Darknet19 & & & & & & + & + & + & + & & & & & & \\
    	 End-to-End & & & & & & & + & + & + & & & & & & \\
    	 Large Decoder & & & & & & & & + & + & & & & & & \\
    	 Distance Transform (STS++) & & & & & & & & & + & & & & & & \\
         \hline
    	 mAP$_{0.5}^r$ & 34.6 & 38.6& 42.3& 41.4& 48.5& 52.2& 51.7& 52.3 & \textbf{53.2} & 49.7 & 59.1 & 63.5 & 62.0 & 65.7 & \textbf{65.7} \\
    	 mAP$_{0.7}^r$ & 15.0 & 17.4& 20.8& 20.5& 26.7& 31.1& 27.9& 31.9& \textbf{32.3} & 25.3 & 36.0 & 41.5 & 44.8 & 48.3 & \textbf{52.1} \\
    	 mAP$_{vol}^r$ & 31.5 & 34.3& 37.0& 36.1& 41.4& 44.5& 43.6& 45.0& \textbf{45.4} & 41.4 & - & - & 55.4 & - & - \\
         \hline
         runtime/frame (s) & $0.028$ & - & - & - & - & - & - & - & - & $48$ & $0.36$ & $1.5$ & $1.37$ & $0.78$ & $\mathbf{0.24}$\\
         frames/sec & $\sim35$ & - & - & - & - & - & - & - & $\mathbf{\sim35}$ & $0.02$ & $2.8$ & $0.67$ & $0.73$ & $1.28$ & $\mathbf{4.17}$\\
         \hline
    \end{tabular}}
  \end{center}
\caption[Ablative performance analysis of all proposed changes on SBD val.]{Effects of the proposed changes on the instance segmentation performance measured in terms of $mAP^r$ on SBD val~\citep{Hariharan2011}. The revised STS++ is compared against existing instance segmentation methods in terms of performance accuracy and the processing speed (measured as both runtime/frame in sec. and frames/sec).}
\label{tab-exp-allresults}
\end{table*}

\textbf{Training End-to-End:}\hspace{2mm}
Training the shape decoder in a unified end-to-end way simplifies the overall optimisation procedure.
Moreover, learning the shape decoder in an independent optimisation step can lead to a solution that is sub-optimal overall.
Thus, we first train the decoder from a random initialisation using the original architecture discussed in STS~\citep{sts_jetley16}.
This leads to a degradation in the performance, especially, in the high IoU range (see Figure \ref{tbl-exp-e2e-dt}).
This suggests that the model has difficulty in reproducing fine details of the instance shape masks.
We then train the decoder with an increased learning capacity (i.e. an increased number of parameters), and call it the Large Decoder model.
This model is able to recover from the drop in prediction quality noted at high IoU values, and surpass the Darknet19 model with a pre-trained decoder by a small margin.

% \begin{figure}/
%   \begin{minipage}{0.47\textwidth}
%     \begin{center}
%       \includegraphics[width=\textwidth]{figures/experiment-e2e}
%     \end{center}
%     \caption{SBD $mAP^r_{0.5}$ estimates for end-to-end decoder training}
%     \label{fig-exp-e2e}
%   \end{minipage}\hspace{0.06\textwidth}
%   \begin{minipage}{0.47\textwidth}
%     \begin{center}
%       \includegraphics[width=\textwidth]{figures/experiment-dt}
%     \end{center}
%     \caption{SBD $mAP^r_{0.5}$ estimates for distance transfrorm representation}
%     \label{fig-exp-dt}
%   \end{minipage}
% \end{figure}

\textbf{Distance Transform based Shape Encoding:}\hspace{2mm}
The alternative shape representation making use of quantised distance transform values offers a slight improvement (up to $1\%$ in terms of $mAP^{r}$ accuracy) for low IoU values ($\sim 0.5$), as can be seen in Figure~\ref{tbl-exp-e2e-dt}).
However, the model still fails to predict the fine details of the object shape masks near the object boundaries, a flaw that manifests itself as low $mAP^{r}$ scores at high IoU values.

\begin{figure}[H]
\centering
\resizebox{0.75\linewidth}{!}{
  \begin{minipage}{0.32\linewidth}
      \centering
      \mbox{Gnd. Truth}
      \includegraphics[width=\textwidth]{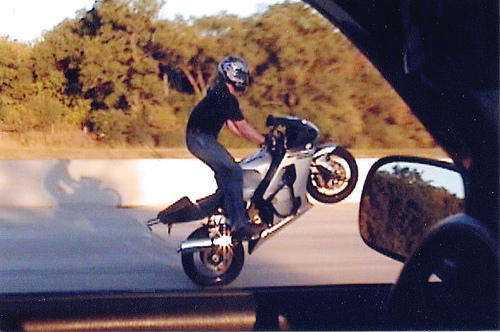}
  \end{minipage}
  \begin{minipage}{0.32\linewidth}
  \centering
      \mbox{STS}
      \includegraphics[width=\textwidth]{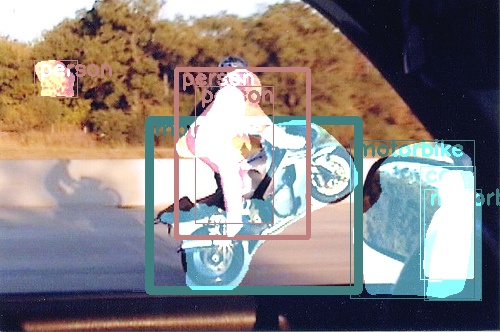}
  \end{minipage}
  \begin{minipage}{0.32\linewidth}
  \centering
      \mbox{STS++}
      \includegraphics[width=\textwidth]{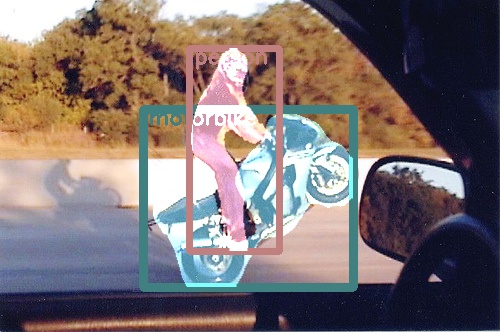}
  \end{minipage}}

\resizebox{0.75\linewidth}{!}{
  \begin{minipage}{0.32\linewidth}
    \includegraphics[width=\textwidth]{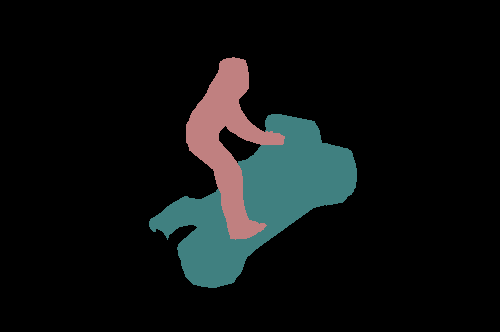}
  \end{minipage}
  \begin{minipage}{0.32\linewidth}
    \includegraphics[width=\textwidth]{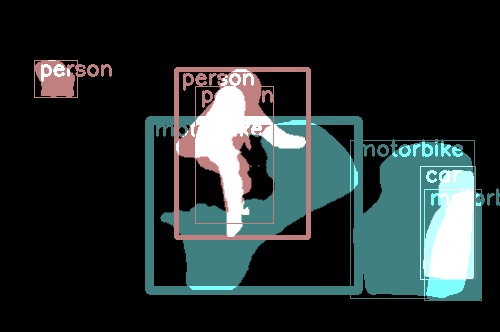}
  \end{minipage}
  \begin{minipage}{0.32\linewidth}
    \includegraphics[width=\textwidth]{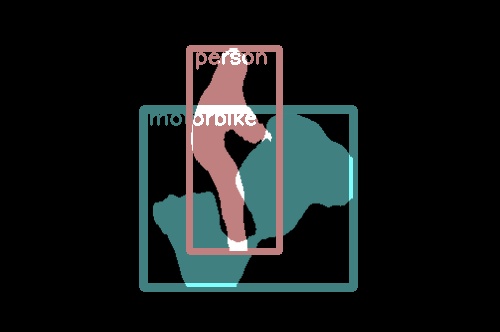}
  \end{minipage}
  }
  
  \vspace{1mm}
  
  \resizebox{0.75\linewidth}{!}{
  \begin{minipage}{0.32\linewidth}
      \centering
    \includegraphics[width=\textwidth]{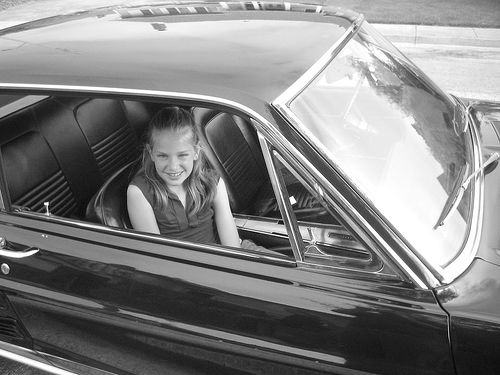}
  \end{minipage}
  \begin{minipage}{0.32\linewidth}
      \centering
    \includegraphics[width=\textwidth]{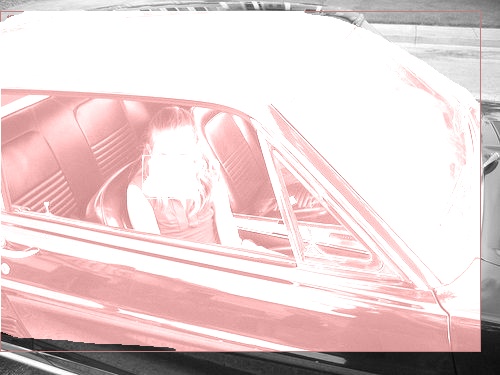}
  \end{minipage}
  \begin{minipage}{0.32\linewidth}
      \centering
    \includegraphics[width=\textwidth]{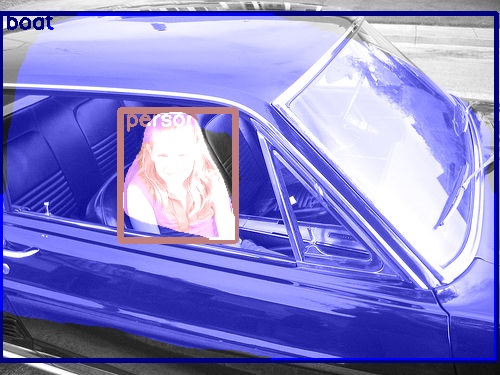}
  \end{minipage}
  }
  
  \resizebox{0.75\linewidth}{!}{
  \begin{minipage}{0.32\linewidth}
    \includegraphics[width=\textwidth]{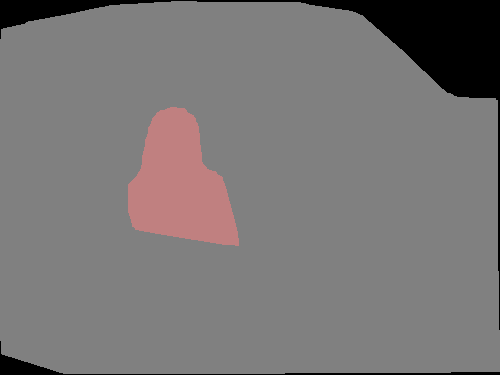}
  \end{minipage}
  \begin{minipage}{0.32\linewidth}
    \includegraphics[width=\textwidth]{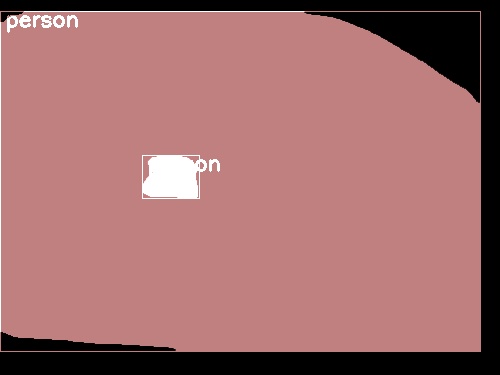}
  \end{minipage}
  \begin{minipage}{0.32\linewidth}
    \includegraphics[width=\textwidth]{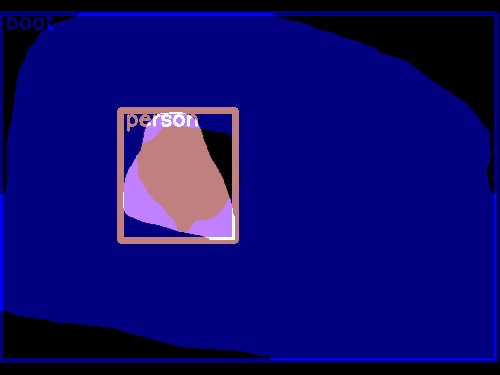}
  \end{minipage}
  }
  \caption[Qualitative instance segmentation results.]
  {Qualitative instance segmentation results: STS (column 2) predicts objects around the rear-view mirror (row 2) and misses the person in the car (row 4).
  Our proposed model (column 3) provides a more detailed prediction of shape masks for the motorcycle and the rider (row 1 and 2) and correctly delineates the person sitting inside the car (row 3 and 4). }
  \label{fig-qual-results}
\end{figure}

Table \ref{tab-exp-allresults} traces the incremental growth in performance, over the full range of IoU values, for the above discussed series of changes to the network architecture. For a qualitative demonstration of the above results, refer to Figure~\ref{fig-qual-results}.

%\begin{figure*}
%  \begin{minipage}{0.47\textwidth}
%    \begin{center}
%      \includegraphics[width=0.7\textwidth]{figures/experiment-bn-da}
%    \end{center}
%    \caption[SBD $mAP^r$ estimates, training with Batch Normalisation and Data Augmentation]
%    {
%    SBD $mAP^r$ estimates, for model trained with Batch Normalisation and Data Augmentation.
%    Neither of implemented changes affect model's architecture or run-time at inference.
%    }
%    \label{fig-exp-map-bn-da}
%  \end{minipage}\hspace{0.06\textwidth}
%  \begin{minipage}{0.47\textwidth}

%\begin{figure*}
%    \begin{minipage}{1\textwidth}

%\subsection{Evaluating Zero-shot Segmentation}

\subsection{Evaluating Object Detection Performance}

%The quality of the predicted instance masks relies heavily on the correct prediction of the corresponding object bounding box parameters.
During training, the target masks of object instances are always generated with respect to the ground truth bounding boxes rather than the predicted ones. This prevents the shape predictions from adjusting themselves to the errors in the prediction of the bounding boxes at the time of training. Thus, the accuracy of the overall instance segmentation systems rests on the quality of the object detections. We therefore proceed to analyse how the quality of the detector unit alone evolves as a function of the proposed changes. We also make a quantitative comparison with the STS model proposed by~\cite{sts_jetley16}. 
We perform the evaluation on \pascal~2007 test set~\citep{DBLP:journals/ijcv/EveringhamEGWWZ15} (which is disjoint from our training set) and use the error analysis methodology and toolkit\footnote{MATLAB code available at~\url{http://dhoiem.web.engr.illinois.edu/projects/detectionAnalysis/}} developed by~\cite{DBLP:conf/eccv/HoiemCD12}.
%All detection predictions are classified either into true positives or four categories of false positive errors: \emph{localisation}, \emph{confusion with similar objects}, \emph{confusion with dissimilar objects} and \emph{confusion with background}.

%\begin{figure}
%\begin{minipage}{0.5\textwidth}
\begin{table}
    \centering
    \resizebox{\linewidth}{!}{
    \begin{tabular}{l|c|cccccccccc}
    	Approach & $mAP^r_{0.5}$ & aeroplane & bicycle & bird & boat & bottle & bus & car & cat & chair & cow\\\hline
    	STS original & 33.5 & 58.6 & 33.7 & 31.2 & 17.6 & 11.0 & 65.5 & 37.0 & 62.6 &  6.2 & 26.1\\
    	Darknet19 & 52.2 & 72.0 & 52.7 & 54.6 & 30.6 & \textbf{28.0} & 74.8 & \textbf{56.8} & 81.8 & 21.1 & \textbf{56.9}\\
    	Large Decoder & 52.3 & \textbf{73.0} & 52.5 & 55.1 & \textbf{34.8} & 24.9 & \textbf{75.3} & 55.3 & \textbf{82.0} & 20.7 & 54.6\\
    	Distance Transform (STS++) & \textbf{53.2} & 72.9 & \textbf{53.6} & \textbf{58.5} & 32.4 & 25.8 & 74.9 & 56.5 & 81.8 & \textbf{22.8} & 55.0\\
    \hline
    	SDS~\citep{Hariharan2014} & 49.7 & 68.4 & 49.4 & 52.1 & 32.8 & 33.0 & 67.8 & 53.6 & 73.9 & 19.9 & 43.7\\
    	\citep{2017cvpr_aarnab} & 62.0 & 80.3 & 52.8 & 68.5 & 47.4 & 39.5 & 79.1 & 61.5 & 87.0 & 28.1 & 68.3\\\hline
    \hline 	 &  & dining table & dog & horse & motorbike & person & potted plant & sheep & sofa & train & tv monitor\\\hline
    	STS original & 33.5 & 13.5 & 49.7 & 31.0 & 37.9 & 37.7 &  7.0 & 31.2 & 20.0 & 62.7 & 29.2\\
    	Darknet19 & 52.2 & 26.0 & 71.4 & 61.6 & \textbf{62.4} & 54.4 & 19.4 & 53.2 & 36.2 & 76.1 & 54.5\\
    	Large Decoder & 52.3 & 25.2 & 71.9 & \textbf{64.8} & 58.6 & 54.0 & 19.6 & 54.9 & \textbf{36.5} & 76.9 & 55.7\\
    	Distance Transform (STS++)& \textbf{53.2} & \textbf{26.2} & \textbf{74.1} & 62.5 & 59.8 & \textbf{57.7} & \textbf{22.7} & \textbf{56.1} & 36.4 & \textbf{78.5} & \textbf{56.4}\\
    \hline
    	SDS~\citep{Hariharan2014} & 49.7 & 25.7 & 60.6 & 55.9 & 58.9 & 56.7 & 28.5 & 55.6 & 32.1 & 64.7 & 60.0\\
        \cite{2017cvpr_aarnab} & 62.0 & 35.5 & 86.1 & 73.9 & 66.1 & 63.8 & 32.9 & 65.3 & 50.4 & 81.4 & 71.4\\\hline
    \end{tabular}}
  \caption[Per category instance segmentation results on SBD val]
  {$AP^r_{0.5}$ scores for individual \pascal~categories on SBD val set.}
  \label{tbl-cat-best-05}
%\end{minipage}
%\end{figure}
\end{table}

\begin{figure}
  \begin{minipage}{0.39\textwidth}
      \includegraphics[width=\textwidth]{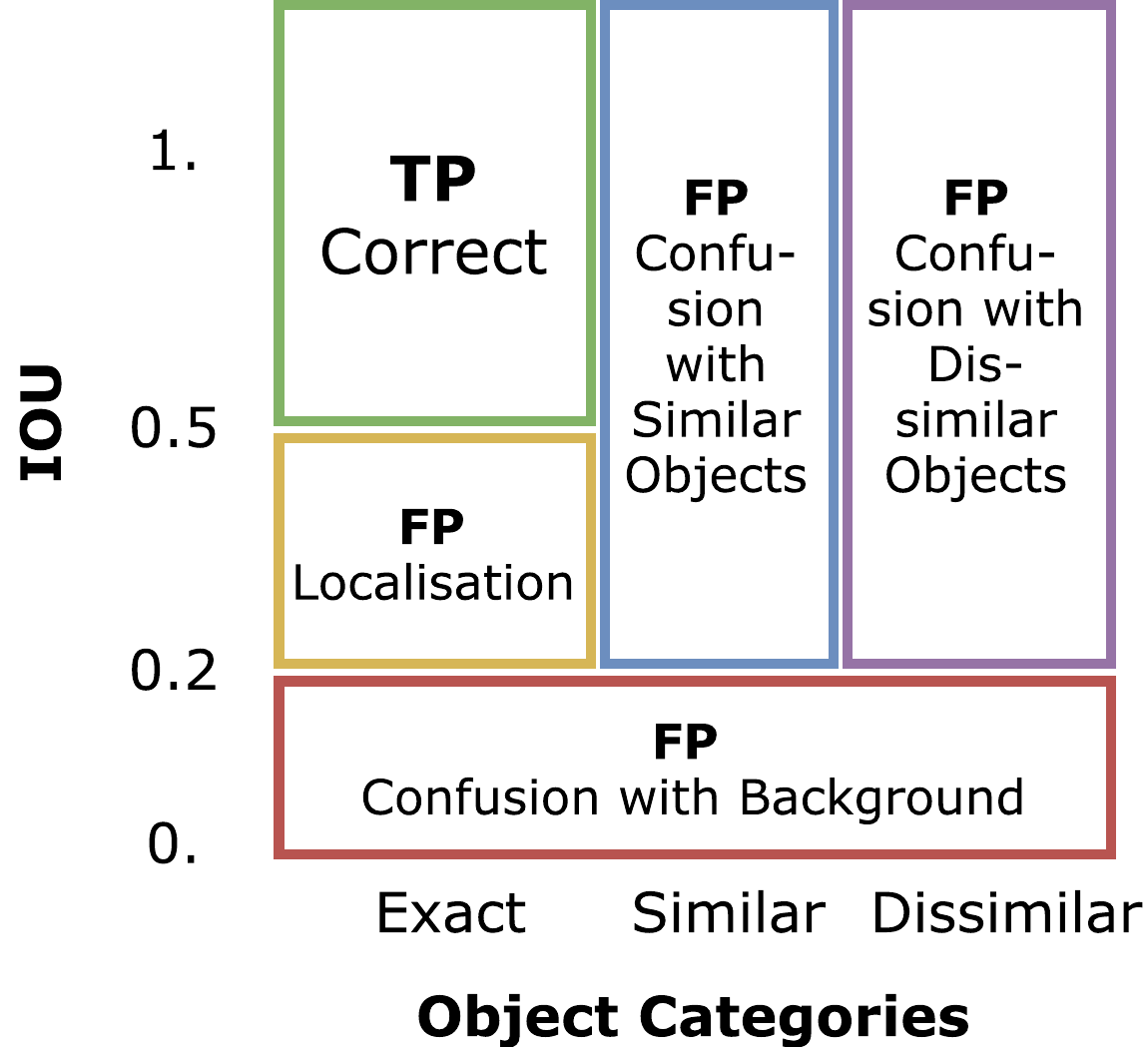}
%      \caption[Schematic illustration for categorising detection predictions]
%      {Five types of objects detections as per~\cite{DBLP:conf/eccv/HoiemCD12}.}
      \label{fig-det-errors-method}
  \end{minipage}
  %\begin{minipage}{0.14\textheight}
  %\centering
  %\resizebox{0.8\linewidth}{!}{
  %\begin{minipage}{0.23\linewidth}
  %     \centering
  %    \mbox{\tiny{GT}}
  %    \includegraphics[width=\textwidth]{figures/segmentations/gt/2008_001514}
  %\end{minipage}
  %\begin{minipage}{0.23\linewidth}
  %     \centering
  %    \mbox{\tiny{STS+ (w/o DT)}}
  %    \includegraphics[width=\textwidth]{figures/segmentations/dn19/2008_001514}
  %\end{minipage}
  %\begin{minipage}{0.23\linewidth}
  %     \centering
  %    \mbox{\tiny{STS+}}
  %    \includegraphics[width=\textwidth]{figures/segmentations/dt/2008_001514}
  %\end{minipage}}
  %\centering
  %\resizebox{0.8\linewidth}{!}{
  %\begin{minipage}{0.23\linewidth}
  %  \includegraphics[width=\textwidth]{figures/segmentations/gt/2008_001514_mask}
  %\end{minipage}
  %\begin{minipage}{0.23\linewidth}
  %  \includegraphics[width=\textwidth]{figures/segmentations/dn19/2008_001514_mask}
  %\end{minipage}
  %\begin{minipage}{0.23\linewidth}
  %  \includegraphics[width=\textwidth]{figures/segmentations/dt/2008_001514_mask}
  %\end{minipage}}
%  \caption[Instance segmentation examples]
%  {Our proposed model fails to capture the complete bird inside the bounding box and overlooks the fine details of its wings.}
  %\label{fig-seg-examples}
  %\end{minipage}
  \begin{minipage}{0.6\textwidth}
    \includegraphics[width=\textwidth]{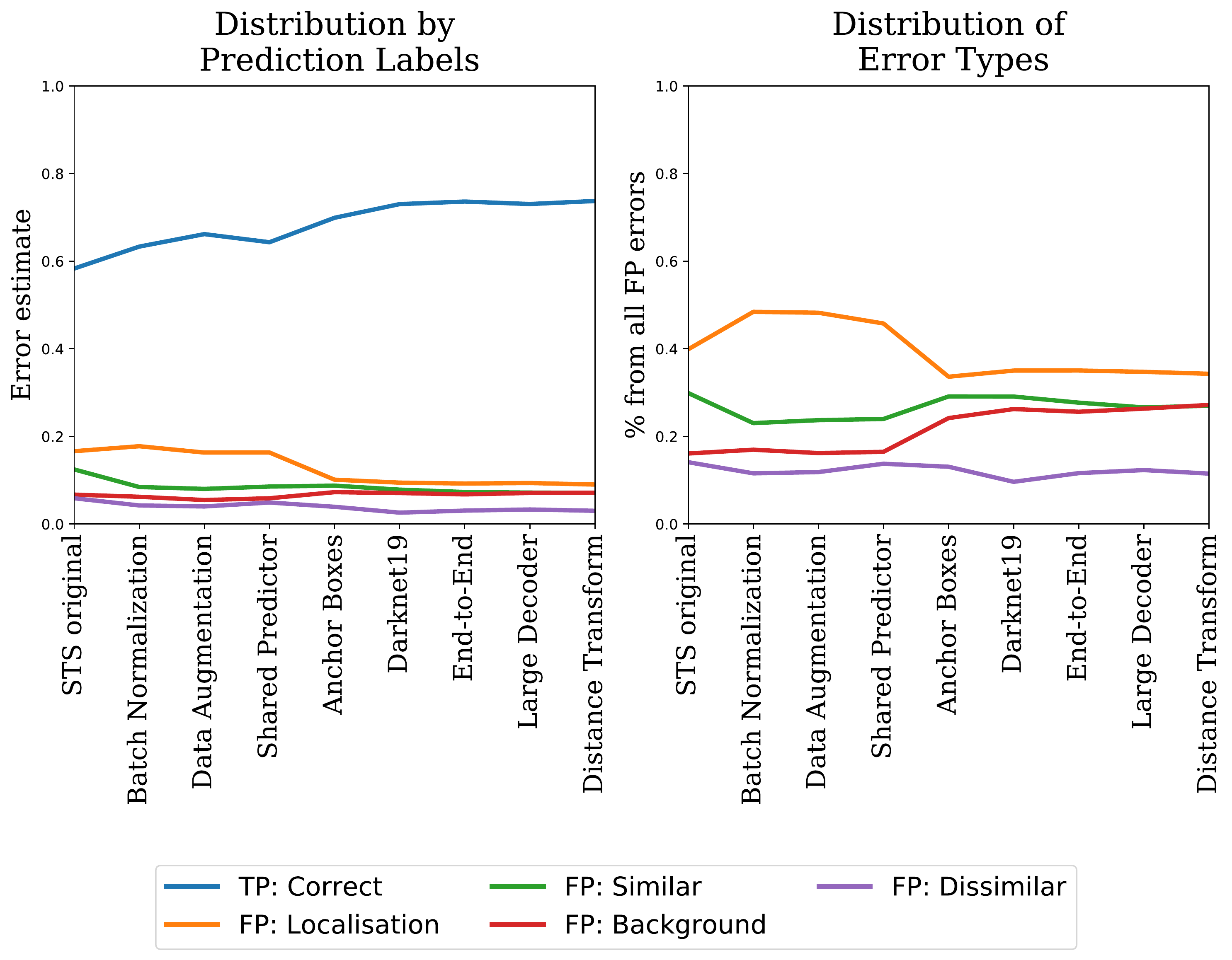}
%    \caption[Object detection error by type on VOC 2007 \emph{test}]
%    {Analysis of detections on \pascal~2007 \emph{test}; (Left) Composition of the total $mAP$ estimate; (Right) $\%$-age contribution of each error type to the total number of false positives. Localisation error persists after a drop due to 'anchor boxes', while background error increases due to incomplete annotations.}
    %Detection metric $mAP$ has improved, however, the quality gap remains persistent due to localisation errors ($\sim 9\%$).
    %(RIGHT) \emph{False positive} errors split by error type.
    %Background error increases due to imperfect annotations in the dataset.}
    \label{fig-det-errors}
  \end{minipage}
  \caption[Types of object detections and their error contributions.]{(Left) Five types of object detections on \pascal~2007 test as per~\cite{DBLP:conf/eccv/HoiemCD12}; (Middle) Typewise break-up of the total $mAP$ error for different architectural modifications; (Right) $\%$-age contribution of each detection type to the total number of false positives.} %Localisation error persists after a drop due to `anchor boxes', background error increases due to imperfect annotations.
  \label{fig-error-analysis}
\end{figure}

Figure~\ref{fig-error-analysis} defines the taxonomy of prediction errors for a detector unit and compares these errors over the different modifications that are studied in this work.
As noted before, the most notable improvement comes from using textit{anchor boxes} instead of simply the grid cells for target representation during network training. This has the biggest impact on the quality of object localisation.
Despite the improvement, the localisation error still remains a main contributing factor, making up more than $9\%$ of the total $mAP$ error. %, and needs to be addressed further.
For details of the detection performance on distinct object categories see Tables \ref{tbl-detection-errors} through \ref{tbl-dlt-errors-07} in Appendix \ref{ch-errors}.
%Appendix A.5 Tables 11-13.
Notice also that the number of background errors is strongly inversely correlated to the localisation error (Figure \ref{fig-error-analysis}).
This implies that as the model becomes more precise in localising objects, it is also more prone to detecting background regions as likely object instances. Anecdotally, it appears that the model is detecting true objects in the input images which are not actually annotated in the dataset, see Figure~\ref{fig:extradetection} in Appendix~\ref{ch-segmentations}. This is, however, a drawback of the evaluation dataset rather than that of the prediction model. 
%Few examples supporting this claim are shown in Figure~\ref{}. %%SJ

\section{Discussion and Conclusion}

In this work we tackle the problem of real-time instance segmentation.
As shown in Figure~\ref{fig-perf-illus}, our revised STS++ model sets a new performance benchmark at real-time processing rates for the task of multiple object instance segmentation. 
The changes we implement improve the overall model performance by almost $20\%$ in terms of $mAP^r_{0.5}$ and reduce the total number of parameters by $60\%$ via the reuse of network parameters over different spatial locations.
Tables \ref{tab-exp-allresults} \& \ref{tbl-cat-best-05} summarise the effects of the various atomic changes made to the pipeline measured in terms of the mean and individual $AP^r$ scores over $20$ distinct object categories of \pascal dataset.
The instance mask visualisations in Appendix \ref{ch-segmentations} %Appendix A.4
demonstrate that the new model is better at localising instances and inferring the general body shapes.
For more details of the error measurements, refer to Tables \ref{tbl-dlt-errors-05} and \ref{tbl-dlt-errors-07} in Appendix \ref{ch-errors}.
%Appendix A.5 Tables 14 \& 15.

%The biggest increase ($7.1\%$ in $mAP^r_{0,5}$) comes from changing the methodology of how model gets assigned its targets.
Significant improvements have been made to the real-time instance segmentation solution proposed by~\cite{sts_jetley16} making it more attractive for practical use. Yet there remains an accuracy gap in comparison to the state of the art. We believe that the two main challenges in building more accurate models are as follows.
Firstly, models demonstrate poor quality in terms of $mAP^r_t$ for low IoU values (i.e. $IoU \in[0.1,0.4]$) which indicates issues with correctly localising instance-level object bounding boxes. The analysis of detection errors on \pascal~dataset confirms erroneous localisation as the biggest contributor to the total number of false positives.
Moreover, under the current model, instance mask prediction is decoupled from bounding box estimation and has no mechanism to adjust or recover if there is an error in the prediction of the bounding box.
Increasing the capacity to represent ground truth instances with different sizes and aspect ratios was demonstrated to yield a great improvement in the results.
A further enhancement in the encoding of ground truth information must be sought.
Secondly, a poor performance in the high IoU range ($IoU\in[0.7,1]$) indicates the model's inability to capture intricate details of objects' boundaries.
This can be addressed by training pixel-level bottom-up segmentation models such as those making use of conditional random fields (CRFs).
The instance segmentation masks obtained from the proposed model can be used as priors for an additional CRF based boundary refinement stage.
A similar approach was taken by~\cite{Arnab2016}, however, bounding box priors were used instead of segmentation masks.
Taking more precise priors in the form of instance masks could further benefit the CRF based post-processing operation.

\clearpage
\begin{subappendices}
\section{Appendix}

\subsection{Architectures of Models}
\label{ch-architectures}

Tables~\ref{tbl-arc-sts-orig} through~\ref{tbl-arc-decoder-prop-dt}, describe the different neural network architectures used in this project.
Every feed-forward network is defined as a sequence of layers.
Each layer is described by providing its type, number of output filters (feature maps), filter (kernel) size dimensions, spatial stride parameter, spatial dimensions of the output layer, number of arithmetic operations performed and the total number of learnable parameters.
The layer types include convolutional layer (CONV), transposed convolutional layer (TCONV), max-pooling layer (MAXPOOL) and spatial up-sampling layer (UPSAMPLE).
All the convolutional layers are padded with zeros in order to maintain the spatial dimensions of the output feature maps.
A single step of addition, multiplication or \textsf{max} comparison is considered to be a single arithmetic operation.

\clearpage

\begin{table}[H]
  \begin{center}
    \resizebox{0.9\linewidth}{!}{
    \begin{tabular}{cc|c|cc|c|cc}
      & Type & Filters & Size & Stride & Output & Ops, $\times10^6$ & Params, $\times10^6$ \\\hline
      1: & CONV & 64 & $7\times7$ & 2 & $224\times224\times64$ &   944 &  0.01\\
      2: & MAXPOOL & 1 & $2\times2$ & 2 & $112\times112\times64$ &     0 &  0.00\\
      3: & CONV & 192 & $3\times3$ & 1 & $112\times112\times192$ & 2,775 &  0.11\\
      4: & MAXPOOL & 1 & $2\times2$ & 2 & $56\times56\times192$ &     0 &  0.00\\
      5: & CONV & 128 & $1\times1$ & 1 & $56\times56\times128$ &   154 &  0.02\\
      6: & CONV & 256 & $3\times3$ & 1 & $56\times56\times256$ & 1,850 &  0.30\\
      7: & CONV & 256 & $1\times1$ & 1 & $56\times56\times256$ &   411 &  0.07\\
      8: & CONV & 512 & $3\times3$ & 1 & $56\times56\times512$ & 7,399 &  1.18\\
      9: & MAXPOOL & 1 & $2\times2$ & 2 & $28\times28\times512$ &     0 &  0.00\\
      10: & CONV & 256 & $1\times1$ & 1 & $28\times28\times256$ &   206 &  0.13\\
      11: & CONV & 512 & $3\times3$ & 1 & $28\times28\times512$ & 1,850 &  1.18\\
      12: & CONV & 256 & $1\times1$ & 1 & $28\times28\times256$ &   206 &  0.13\\
      13: & CONV & 512 & $3\times3$ & 1 & $28\times28\times512$ & 1,850 &  1.18\\
      14: & CONV & 256 & $1\times1$ & 1 & $28\times28\times256$ &   206 &  0.13\\
      15: & CONV & 512 & $3\times3$ & 1 & $28\times28\times512$ & 1,850 &  1.18\\
      16: & CONV & 256 & $1\times1$ & 1 & $28\times28\times256$ &   206 &  0.13\\
      17: & CONV & 512 & $3\times3$ & 1 & $28\times28\times512$ & 1,850 &  1.18\\
      18: & CONV & 512 & $1\times1$ & 1 & $28\times28\times512$ &   411 &  0.26\\
      19: & CONV & 1024 & $3\times3$ & 1 & $28\times28\times1024$ & 7,399 &  4.72\\
      20: & MAXPOOL & 1 & $2\times2$ & 2 & $14\times14\times1024$ &     0 &  0.00\\
      21: & CONV & 512 & $1\times1$ & 1 & $14\times14\times512$ &   206 &  0.52\\
      22: & CONV & 1024 & $3\times3$ & 1 & $14\times14\times1024$ & 1,850 &  4.72\\
      23: & CONV & 512 & $1\times1$ & 1 & $14\times14\times512$ &   206 &  0.52\\
      24: & CONV & 1024 & $3\times3$ & 1 & $14\times14\times1024$ & 1,850 &  4.72\\
      25: & CONV & 1024 & $3\times3$ & 1 & $14\times14\times1024$ & 3,699 &  9.44\\
      26: & CONV & 1024 & $3\times3$ & 2 & $7\times7\times1024$ &   925 &  9.44\\
      27: & CONV & 1024 & $3\times3$ & 1 & $7\times7\times1024$ &   925 &  9.44\\
      28: & CONV & 1024 & $3\times3$ & 1 & $7\times7\times1024$ &   925 &  9.44\\
      29: & CONV & 4096 & $7\times7$ & 7 & $1\times1\times4096$ &   411 & 205.52\\
      30: & CONV & 3430 & $1\times1$ & 1 & $1\times1\times3430$ &    28 & 14.05\\
      \hline
      &  &  &  &  & \textbf{Total:} & \textbf{40,586} & \textbf{279.7}
    \end{tabular}
    }
  \end{center}
  \caption[Details of architecture of original STS model.]{Details of the architecture of the original STS model.
NOTE: Number of parameters and operations required are given in \textbf{millions} ($10^6$).}
  \label{tbl-arc-sts-orig}
\end{table}

\begin{table}[H]
  \begin{center}
    \resizebox{0.9\linewidth}{!}{
    \begin{tabular}{cc|c|cc|c|cc}
      & Type & Filters & Size & Stride & Output & Ops, $\times10^6$ & Params, $\times10^6$ \\\hline
      1: & CONV & 64 & $7\times7$ & 2 & $224\times224\times64$ &   944 &  0.01\\
      2: & MAXPOOL & 1 & $2\times2$ & 2 & $112\times112\times64$ &     0 &  0.00\\
      3: & CONV & 192 & $3\times3$ & 1 & $112\times112\times192$ & 2,775 &  0.11\\
      4: & MAXPOOL & 1 & $2\times2$ & 2 & $56\times56\times192$ &     0 &  0.00\\
      5: & CONV & 128 & $1\times1$ & 1 & $56\times56\times128$ &   154 &  0.02\\
      6: & CONV & 256 & $3\times3$ & 1 & $56\times56\times256$ & 1,850 &  0.30\\
      7: & CONV & 256 & $1\times1$ & 1 & $56\times56\times256$ &   411 &  0.07\\
      8: & CONV & 512 & $3\times3$ & 1 & $56\times56\times512$ & 7,399 &  1.18\\
      9: & MAXPOOL & 1 & $2\times2$ & 2 & $28\times28\times512$ &     0 &  0.00\\
      10: & CONV & 256 & $1\times1$ & 1 & $28\times28\times256$ &   206 &  0.13\\
      11: & CONV & 512 & $3\times3$ & 1 & $28\times28\times512$ & 1,850 &  1.18\\
      12: & CONV & 256 & $1\times1$ & 1 & $28\times28\times256$ &   206 &  0.13\\
      13: & CONV & 512 & $3\times3$ & 1 & $28\times28\times512$ & 1,850 &  1.18\\
      14: & CONV & 256 & $1\times1$ & 1 & $28\times28\times256$ &   206 &  0.13\\
      15: & CONV & 512 & $3\times3$ & 1 & $28\times28\times512$ & 1,850 &  1.18\\
      16: & CONV & 256 & $1\times1$ & 1 & $28\times28\times256$ &   206 &  0.13\\
      17: & CONV & 512 & $3\times3$ & 1 & $28\times28\times512$ & 1,850 &  1.18\\
      18: & CONV & 512 & $1\times1$ & 1 & $28\times28\times512$ &   411 &  0.26\\
      19: & CONV & 1024 & $3\times3$ & 1 & $28\times28\times1024$ & 7,399 &  4.72\\
      20: & MAXPOOL & 1 & $2\times2$ & 2 & $14\times14\times1024$ &     0 &  0.00\\
      21: & CONV & 512 & $1\times1$ & 1 & $14\times14\times512$ &   206 &  0.52\\
      22: & CONV & 1024 & $3\times3$ & 1 & $14\times14\times1024$ & 1,850 &  4.72\\
      23: & CONV & 512 & $1\times1$ & 1 & $14\times14\times512$ &   206 &  0.52\\
      24: & CONV & 1024 & $3\times3$ & 1 & $14\times14\times1024$ & 1,850 &  4.72\\
      25: & CONV & 1024 & $3\times3$ & 1 & $14\times14\times1024$ & 3,699 &  9.44\\
      26: & CONV & 1024 & $3\times3$ & 2 & $7\times7\times1024$ &   925 &  9.44\\
      27: & CONV & 1024 & $3\times3$ & 1 & $7\times7\times1024$ &   925 &  9.44\\
      28: & CONV & 1024 & $3\times3$ & 1 & $7\times7\times1024$ &   925 &  9.44\\
      29: & CONV & 2048 & $3\times3$ & 1 & $7\times7\times2048$ & 1,850 & 18.88\\
      30: & CONV & 2048 & $3\times3$ & 1 & $7\times7\times2048$ & 3,699 & 37.75\\
      31: & CONV & 1024 & $1\times1$ & 1 & $7\times7\times1024$ &   206 &  2.10\\
      32: & CONV & 135 & $1\times1$ & 1 & $7\times7\times135$ &    14 &  0.14\\
      \hline
      &  &  &  &  & \textbf{Total:} & \textbf{45,915} & \textbf{119.0}
    \end{tabular}
    }
  \end{center}
  \caption[Details of architecture of STS model with shared detection layer.]{Details of the architecture of the STS model with a shared detection layer.
NOTE: Number of parameters and operations required are given in \textbf{millions} ($10^6$).}
  \label{tbl-arc-sts-shared}
\end{table}

\begin{table}[H]
  \begin{center}
    \resizebox{0.9\linewidth}{!}{
    \begin{tabular}{cc|c|cc|c|cc}
      & Type & Filters & Size & Stride & Output & Ops, $\times10^6$ & Params, $\times10^6$ \\\hline
      1: & CONV & 32 & $3\times3$ & 1 & $448\times448\times32$ &   347 &  0.00\\
      2: & MAXPOOL & 1 & $2\times2$ & 2 & $224\times224\times32$ &     0 &  0.00\\
      3: & CONV & 64 & $3\times3$ & 1 & $224\times224\times64$ & 1,850 &  0.02\\
      4: & MAXPOOL & 1 & $2\times2$ & 2 & $112\times112\times64$ &     0 &  0.00\\
      5: & CONV & 128 & $3\times3$ & 1 & $112\times112\times128$ & 1,850 &  0.07\\
      6: & CONV & 64 & $1\times1$ & 1 & $112\times112\times64$ &   206 &  0.01\\
      7: & CONV & 128 & $3\times3$ & 1 & $112\times112\times128$ & 1,850 &  0.07\\
      8: & MAXPOOL & 1 & $2\times2$ & 2 & $56\times56\times128$ &     0 &  0.00\\
      9: & CONV & 256 & $3\times3$ & 1 & $56\times56\times256$ & 1,850 &  0.30\\
      10: & CONV & 128 & $1\times1$ & 1 & $56\times56\times128$ &   206 &  0.03\\
      11: & CONV & 256 & $3\times3$ & 1 & $56\times56\times256$ & 1,850 &  0.30\\
      12: & MAXPOOL & 1 & $2\times2$ & 2 & $28\times28\times256$ &     0 &  0.00\\
      13: & CONV & 512 & $3\times3$ & 1 & $28\times28\times512$ & 1,850 &  1.18\\
      14: & CONV & 256 & $1\times1$ & 1 & $28\times28\times256$ &   206 &  0.13\\
      15: & CONV & 512 & $3\times3$ & 1 & $28\times28\times512$ & 1,850 &  1.18\\
      16: & CONV & 256 & $1\times1$ & 1 & $28\times28\times256$ &   206 &  0.13\\
      17: & CONV & 512 & $3\times3$ & 1 & $28\times28\times512$ & 1,850 &  1.18\\
      18: & MAXPOOL & 1 & $2\times2$ & 2 & $14\times14\times512$ &     0 &  0.00\\
      19: & CONV & 1024 & $3\times3$ & 1 & $14\times14\times1024$ & 1,850 &  4.72\\
      20: & CONV & 512 & $1\times1$ & 1 & $14\times14\times512$ &   206 &  0.52\\
      21: & CONV & 1024 & $3\times3$ & 1 & $14\times14\times1024$ & 1,850 &  4.72\\
      22: & CONV & 512 & $1\times1$ & 1 & $14\times14\times512$ &   206 &  0.52\\
      23: & CONV & 1024 & $3\times3$ & 1 & $14\times14\times1024$ & 1,850 &  4.72\\
      24: & CONV & 2048 & $3\times3$ & 2 & $7\times7\times2048$ & 1,850 & 18.88\\
      25: & CONV & 1024 & $1\times1$ & 1 & $7\times7\times1024$ &   206 &  2.10\\
      26: & CONV & 2048 & $3\times3$ & 1 & $7\times7\times2048$ & 1,850 & 18.88\\
      27: & CONV & 1024 & $1\times1$ & 1 & $7\times7\times1024$ &   206 &  2.10\\
      28: & CONV & 2048 & $3\times3$ & 1 & $7\times7\times2048$ & 1,850 & 18.88\\
      29: & CONV & 1024 & $1\times1$ & 1 & $7\times7\times1024$ &   206 &  2.10\\
      30: & CONV & 2048 & $3\times3$ & 1 & $7\times7\times2048$ & 1,850 & 18.88\\
      31: & CONV & 1024 & $1\times1$ & 1 & $7\times7\times1024$ &   206 &  2.10\\
      32: & CONV & 70 & $1\times1$ & 1 & $7\times7\times70$ &     7 &  0.07\\
      \hline
      &  &  &  &  & \textbf{Total:} & \textbf{30,155} & \textbf{103.8}
    \end{tabular}
    }
  \end{center}
  \caption[Details of architecture of modified Darknet19 model.]{Details of the architecture of the modified Darknet19 model.
  NOTE: Number of parameters and operations required are given in \textbf{millions} ($10^6$).}
  \label{tbl-arc-dn19}
\end{table}

\begin{table}[H]
  \begin{center}
    \resizebox{0.9\linewidth}{!}{
    \begin{tabular}{cc|c|cc|c|cc}
      & Type & Filters & Size & Stride & Output & Ops, $\times10^3$ & Params, $\times10^3$ \\\hline
      1: & TCONV & 100 & $4\times4$ & 1 & $4\times4\times100$ &    64 & 32.10\\
      2: & UPSAMPLE & 100 & $2\times2$ & 1 & $8\times8\times100$ &     6 &  0.00\\
      3: & CONV & 50 & $3\times3$ & 1 & $8\times8\times50$ & 5,760 & 45.05\\
      4: & UPSAMPLE & 50 & $2\times2$ & 1 & $16\times16\times50$ &    13 &  0.00\\
      5: & CONV & 20 & $3\times3$ & 1 & $16\times16\times20$ & 4,608 &  9.02\\
      6: & UPSAMPLE & 20 & $2\times2$ & 1 & $32\times32\times20$ &    20 &  0.00\\
      7: & CONV & 10 & $3\times3$ & 1 & $32\times32\times10$ & 3,686 &  1.81\\
      8: & UPSAMPLE & 10 & $2\times2$ & 1 & $64\times64\times10$ &    41 &  0.00\\
      9: & CONV & 1 & $3\times3$ & 1 & $64\times64\times1$ &   737 &  0.09\\
      \hline
      &  &  &  &  & \textbf{Total:} & \textbf{14,936} & \textbf{ 88.1}
    \end{tabular}
    }
  \end{center}
  \caption[Details of architecture of learned shape decoder (in STS).]{Details of the architecture of the learned shape decoder in STS model.
  NOTE: Number of parameters and operations required are given in \textbf{thousands} ($10^3$).}
  \label{tbl-arc-decoder-orig}
\end{table}

\begin{table}[H]
  \begin{center}
    \resizebox{0.9\linewidth}{!}{
    \begin{tabular}{cc|c|cc|c|cc}
     & Type & Filters & Size & Stride & Output & Ops, $\times10^6$ & Params, $\times10^6$ \\\hline
    	1: & TCONV & 1024 & $5\times5$ & 1 & $5\times5\times1024$ &     7 &  3.28\\
    	2: & TCONV & 256 & $3\times3$ & 2 & $11\times11\times256$ &   118 &  2.36\\
    	3: & CONV & 192 & $3\times3$ & 1 & $11\times11\times192$ &   107 &  0.44\\
    	4: & TCONV & 128 & $3\times3$ & 2 & $23\times23\times128$ &    54 &  0.22\\
    	5: & CONV & 96 & $3\times3$ & 1 & $23\times23\times96$ &   117 &  0.11\\
    	6: & TCONV & 64 & $3\times3$ & 2 & $47\times47\times64$ &    59 &  0.06\\
    	7: & CONV & 1 & $3\times3$ & 1 & $47\times47\times1$ &     3 &  0.00\\
    \hline
     &  &  &  &  & \textbf{Total:} & \textbf{  463} & \textbf{  6.5}
    \end{tabular}
    }
  \end{center}
  \caption[Details of architecture of proposed large shape decoder.]{Details of the architecture of the proposed large shape decoder.
  NOTE: Number of parameters and operations required are given in \textbf{millions} ($10^6$).}
  \label{tbl-arc-decoder-prop}
\end{table}

\begin{table}[H]
  \begin{center}
    \resizebox{0.9\linewidth}{!}{
    \begin{tabular}{cc|c|cc|c|cc}
     & Type & Filters & Size & Stride & Output & Ops, $\times10^6$ & Params, $\times10^6$ \\\hline
    	1: & TCONV & 1024 & $5\times5$ & 1 & $5\times5\times1024$ &     7 &  3.28\\
    	2: & TCONV & 256 & $3\times3$ & 2 & $11\times11\times256$ &   118 &  2.36\\
    	3: & CONV & 192 & $3\times3$ & 1 & $11\times11\times192$ &   107 &  0.44\\
    	4: & TCONV & 128 & $3\times3$ & 2 & $23\times23\times128$ &    54 &  0.22\\
    	5: & CONV & 96 & $3\times3$ & 1 & $23\times23\times96$ &   117 &  0.11\\
    	6: & TCONV & 64 & $3\times3$ & 2 & $47\times47\times64$ &    59 &  0.06\\
    	7: & CONV & 8 & $3\times3$ & 1 & $47\times47\times8$ &    20 &  0.00\\
    	8: & DT & 8 & $15\times15$ & 1 & $47\times47\times8$ &    64 &  0.00\\
    	9: & CONV & 1 & $1\times1$ & 1 & $47\times47\times1$ &     0 &  0.00\\
    \hline
     &  &  &  &  & \textbf{Total:} & \textbf{  545} & \textbf{  6.5}
    \end{tabular}
    }
  \end{center}
  \caption[Details of architecture of distance transform based neural shape decoder.]{Details of the architecture of the distance transform based neural shape decoder.  NOTE: Number of parameters and operations required are given in \textbf{millions} ($10^6$).}
  \label{tbl-arc-decoder-prop-dt}
\end{table}

\subsection{Experimental Setup}
\label{ch-experimentalsetup}
The original STS model makes use of the Darknet model~\citep{Darknet2013} as its training and inference workhorse (implemented in plain \emph{C}).
The vanilla version of the Darknet model is updated with shape prediction capabilities and additional software layers (in \emph{C++}) for dataset loading and manipulation, shape mask reconstruction, and for evaluating the model on the task of instance segmentation~\footnote{The project source code can be found at \url{https://github.com/torrvision/straighttoshapes}.}.
We ran all our experiments on a single desktop machine with Intel Core i7-4960X CPU (3.6GHz , 6 cores) and NVidia GeForce GTX Titan X GPU (12 GB RAM).
SBD dataset~\citep{Hariharan2011} is chosen as the experimental dataset to benchmark our models. This dataset is divided into $5623$ \emph{train} and $5732$ validation images for training and evaluation respectively.
The performance accuracy is measured in terms of $mAP_{0.5}^r$, $mAP_{0.7}^r$ and $mAP_{vol}^r$ scores.

Due to memory constraints, the data is processed in the form of mini-batches of 8 training examples each. However, the parameters are only updated after accumulating gradients from 8 such mini-batches.
In particular, batch-normalisation parameters are computed over 8 training examples, while gradient descent is performed over 64 such examples.
The model is trained for $750$ epochs ($\sim 65000$ mini-batches) using Stochastic Gradient Descent (SGD) with momentum ($0.9$) and weight decay ($5\times10^{-4}$).
Leaky ReLU with an $\alpha=0.1$ is the non-linearity used throughout our networks.
The parameters in the initial layers of all the neural networks are borrowed from the pre-training of the Darknet model on ImageNet dataset~\citep{ILSVRC15} for the task of image classification~\footnote{Pre-trained Darknet model can be downloaded from \url{https://pjreddie.com/darknet/yolo/}.}.
The learning rate schedule for the training of the network on the task of instance segmentation is as follows:

\begin{center}
\resizebox{0.4\columnwidth}{!}{
    \begin{tabular}{cc}\hline
      \textbf{Batch number} & \textbf{Learning rate} \\\hline
      $1 - 200$ & $1\times10^{-3}$ \\
      $201 - 400$ & $2.5\times10^{-3}$ \\
      $401 - 20~000$ & $5\times10^{-3}$ \\
      $20~001 - 30~000$ & $2.5\times10^{-3}$ \\
      $30~001 - 40~000$ & $1.25\times10^{-3}$ \\
      $40~001 - 50~000$ & $6.25\times10^{-4}$ \\
      $50~001 - 60~000$ & $3.16\times10^{-4}$ \\
      $60~001 - 65~000$ & $1.56\times10^{-4}$ \\\hline
    \end{tabular}
  }
\end{center}

The learning rate is kept low in the beginning in order to preserve the pre-trained weights. It is subsequently increased in order to speed up convergence and then reduced once more as we fine-tune the solution in the later stages of network training.

\subsection{Loss Function Design}
\label{ch-lossfunc}
The model is trained solely as a regression problem and the loss is evaluated depending on where the ground truth objects appear in the input image.

Given a cell $1\leq{i}\leq{S^2}$ containing an object, let $\hat{\mathbf{b}}_{ij}$ denote the parameters of the predicted bounding box at that cell location.
We compute $iou_{ij}$ as the IoU between the ground truth box and the predicted box $j$ and assign $b_i=\argmax_{1\leq{j}\leq{B}}{iou_{ij}}$ as the index of the best prediction in cell $i$.
Further, we define indicators $\nu_{ij}=\mathds{1}\{j=b_i\}$ and $\mu_{i}=\mathds{1}\{\textnormal{cell $i$ contains an object}\}$ to capture the truth of the predicted box being the best fit and the cell containing the ground truth object respectively.

\begin{figure}[ht]
  \begin{center}
    \includegraphics[width=0.3\linewidth]{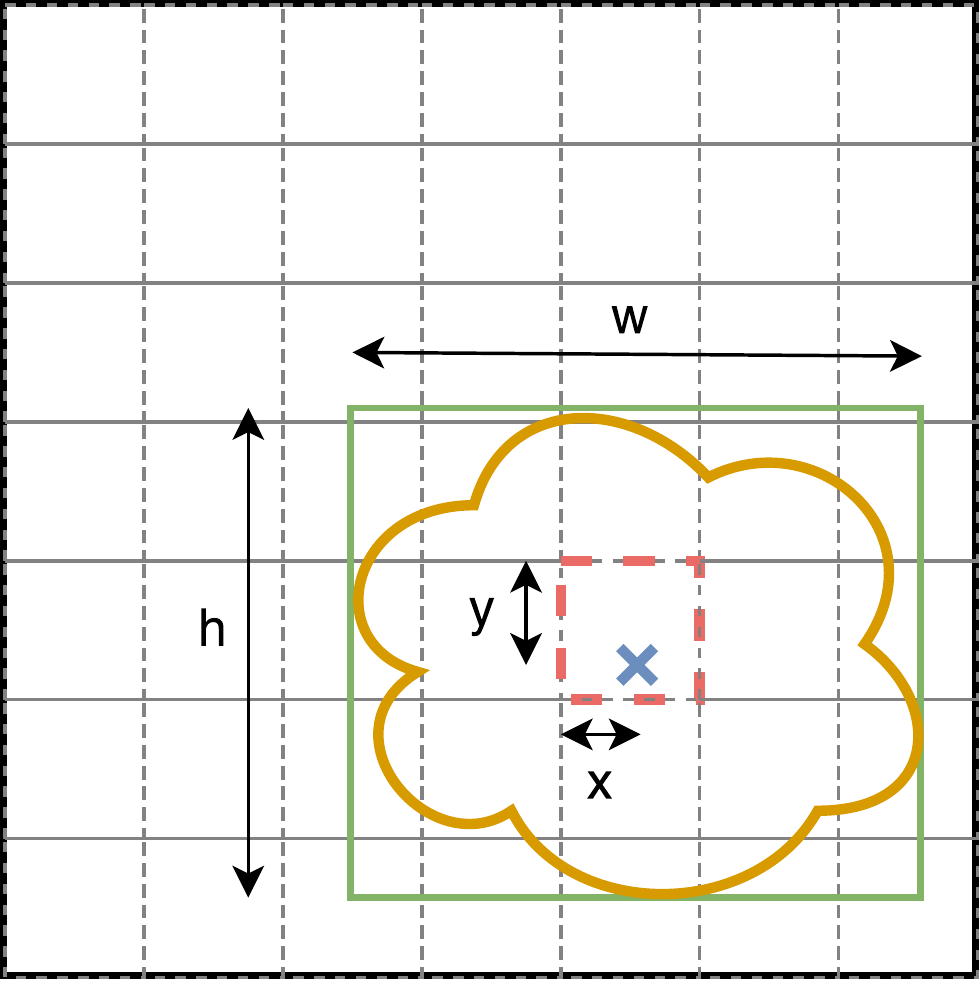}\qquad
    \includegraphics[width=0.3\linewidth]{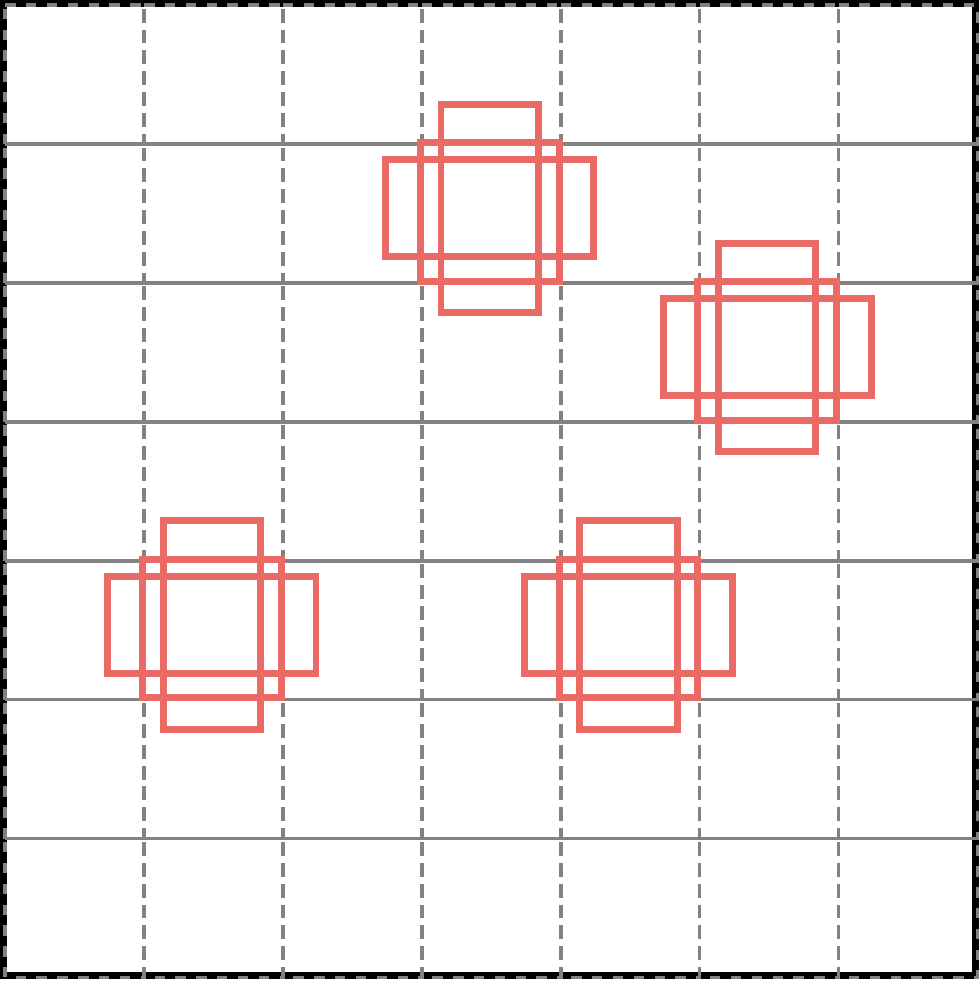}
  \end{center}
  \caption[Generation of bounding-box ground truth data for object localisation.]
  {
  (Left) Grid cell (red) containing the center of the object (marked as a blue cross) is responsible for predicting the object's bounding box (green).
  The center coordinates are normalised relative to the grid cell, while the dimensions of the bounding-box are normalised relative to the dimensions of the complete image.
  (Right) There are 3 anchors boxes of different aspect ratios per grid cell location. The illustration displays several of these boxes at different locations in the image. Each anchor box gets assigned the ground truth object that shares the highest IoU with the anchor. Clearly, different anchors specialise in predicting objects of different aspect ratios.
  }
  \label{fig-cells}
\end{figure}

Then the different terms of the objective function can be defined as follows:
\begin{equation}
%\begin{aligned}[b]
  L_{coord}(\hat{\mathbf{y}}, \mathbf{y}) = \sum_{i=1}^{S^2}\mu_{i}\sum_{j=1}^B\nu_{ij} \left[(\hat{x}_{ij}-x_{i})^2 + (\hat{y}_{ij}-y_{i})^2 + (\hat{\psi}_{ij}-\psi_{i})^2+(\hat{\omega}_{ij}-\omega_{i})^2\right]
%\end{aligned}
\label{eq-loss-coord}
\end{equation}
where $(x_i,y_i)\in[0,1)^2$ describes the center coordinates of the ground truth bounding box (see Figure~\ref{fig-cells} above and %\S\ref{ss-dataprep} 
\S3.2.2 for how target values are constructed).
Note that the model predicts the square root of the bounding box dimensions, i.e. $\psi_i=\sqrt{h_i}$, $\omega_i=\sqrt{w_i}$, for reasons described in \S3.1.%\ref{original_model}.
The model is penalised for having high confidence predictions when (i) there is no ground truth object in the cell location, or (ii) its bounding box prediction is not the current best for that cell location:
\begin{equation}
  L_{conf}^{noobj}(\hat{\mathbf{y}},\mathbf{y})=
  \sum_{i=1}^{S^2}\sum_{j=1}^B(1-\mu_{i}\nu_{ij})(\hat{c}_{ij}-0)^2
\end{equation}
In contrast, when cell $i$ does contain a ground truth object, the confidence score of the best overlapping bounding box is penalised for deviating from the associated IoU as follows:
\begin{equation}
  L_{conf}^{obj}(\hat{\mathbf{y}},\mathbf{y})=
  \sum_{i=1}^{S^2}\mu_{i}\sum_{j=1}^B\nu_{ij}(\hat{c}_{ij}-iou_{ij})^2
\end{equation}
The conditional class probabilities are modelled independently from the bounding box predictions and also independently for each class (as $C$ binary random variables),
\begin{equation}
  L_{class}(\hat{\mathbf{y}},\mathbf{y})=
  \sum_{i=1}^{S^2}\mu_{i}\sum_{j=1}^B\nu_{ij}\sum_{k=1}^C(\hat{p}_{ik}-p_{ik})^2
\end{equation}
All the loss terms defined up until this point are used for the task of detection and are exactly as those used for training the YOLO detector.
In order to address the task of segmentation the STS model additionally regresses to shape representation values as follows:
\begin{equation}\label{eq-loss-shape}
  L_{shape}(\hat{\mathbf{y}},\mathbf{y})=
  \sum_{i=1}^{S^2}\mu_{i}\sum_{j=1}^B\nu_{ij}\sum_{k=1}^M(\hat{s}_{ijk}-s_{ik})^2
\end{equation}
, where $s_{ik}$ denotes the target shape representation (see \S3.2.2) while $\hat{s}_{ijk}$ denotes the predicted shape mask.

The overall objective of the optimisation problem is then expressed as a weighted sum of the above terms (Eq. \ref{eq-loss-coord})-(\ref{eq-loss-shape}):
\begin{equation}
%\begin{aligned}[b]
  L= \lambda_{coord}L_{coord}+ \lambda_{noobj}L_{conf}^{noobj}+ \lambda_{obj}L_{conf}^{obj}+ \lambda_{class}L_{class}+ \lambda_{shape}L_{shape}
%\end{aligned}
\end{equation}
, where the values
$\lambda_{coord}=5.0$,
$\lambda_{noobj}=0.5$,
$\lambda_{obj}=1.0$,
$\lambda_{class}=1.0$, and
$\lambda_{shape}=0.15$, 
are chosen via cross-validation.

% \textbf{[TODO] Discuss NLL and modelling class distribution; Mention L1 loss experiments}

Remark:
Note that the prediction of the object bounding box and shape masks is decoupled in the STS model.
This is to say that whenever the former introduces a discrepancy in the bounding box location, the latter is not corrected or adjusted in any way.
This is in contrast to other state-of-the-art instance segmentation models~\citep{dai2015_multitaskcascade, DBLP:journals/corr/HeGDG17}, where the target mask is adjusted to the predicted bounding box and thus the error therein.

\subsection{More Qualitative Results}\label{ch-segmentations}

The section contains some example images from the SBD validation set and their corresponding instance segmentation results. 
These results as visualised in the following $8$ ways: the top row contains images overlaid with predicted instance masks, while the bottom row simply contains these masks on a black background.
The segmentation masks (left to right) include those from the ground truth, original STS model, proposed Darknet19 with pre-trained shape decoder, and proposed Darknet19 trained end-to-end using distance transform based shape representations.

\begin{figure*}[h]
  \begin{minipage}{0.23\textwidth}
    \begin{center}
      \mbox{}{Ground Truth}
      \includegraphics[width=\textwidth]{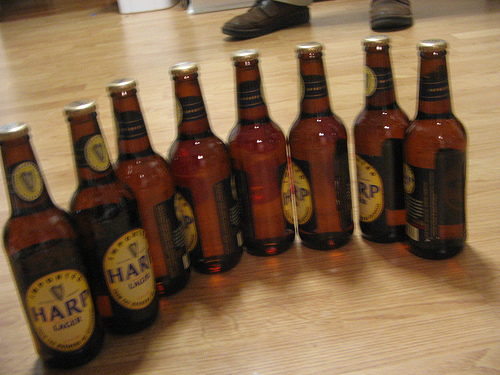}
    \end{center}
  \end{minipage}\hspace{0.015\textwidth}
  \begin{minipage}{0.23\textwidth}
    \begin{center}
      \mbox{}{STS original}
      \includegraphics[width=\textwidth]{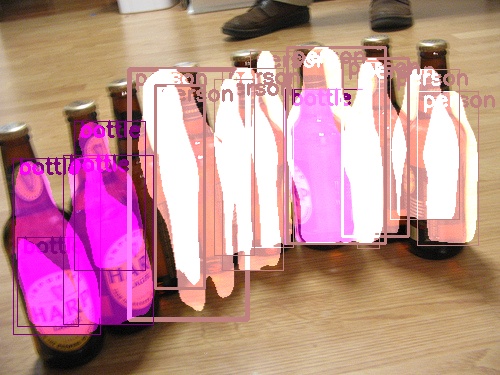}
    \end{center}
  \end{minipage}\hspace{0.015\textwidth}
  \begin{minipage}{0.23\textwidth}
    \begin{center}
      \mbox{}{STS (Darknet19)}
      \includegraphics[width=\textwidth]{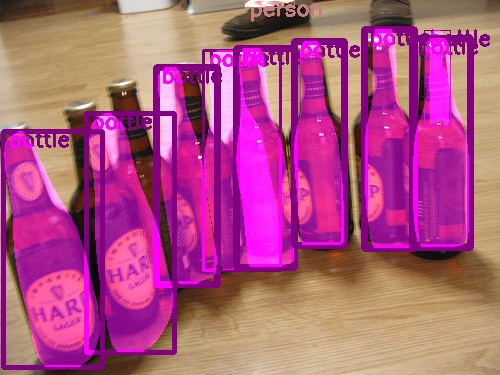}
    \end{center}
  \end{minipage}\hspace{0.015\textwidth}
  \begin{minipage}{0.23\textwidth}
    \begin{center}
      \mbox{}{STS++}
      \includegraphics[width=\textwidth]{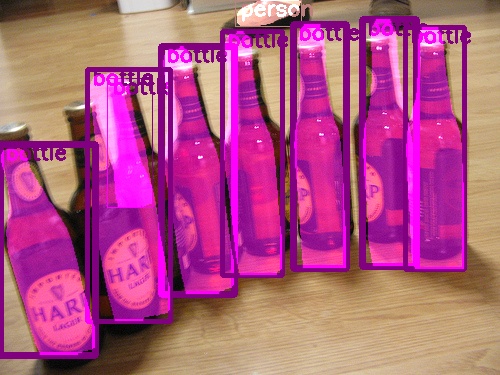}
    \end{center}
  \end{minipage}

\vspace{2pt}

  \begin{minipage}{0.23\textwidth}
    \includegraphics[width=\textwidth]{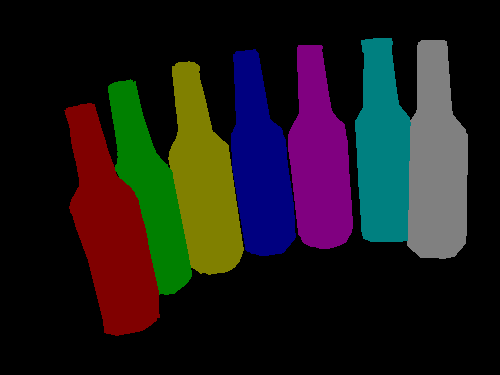}
  \end{minipage}\hspace{0.015\textwidth}
  \begin{minipage}{0.23\textwidth}
    \includegraphics[width=\textwidth]{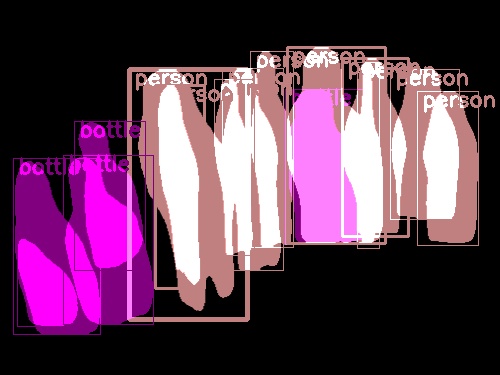}
  \end{minipage}\hspace{0.015\textwidth}
  \begin{minipage}{0.23\textwidth}
    \includegraphics[width=\textwidth]{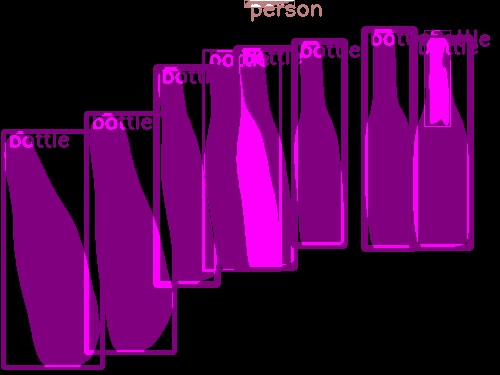}
  \end{minipage}\hspace{0.015\textwidth}
  \begin{minipage}{0.23\textwidth}
    \includegraphics[width=\textwidth]{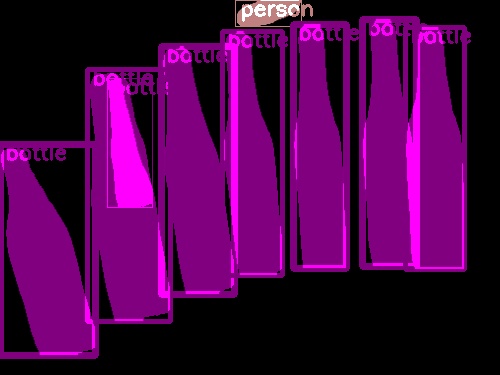}
  \end{minipage}
  \caption[Proposed models do a better job of segmenting overlapping objects.]
  {
  The original model confuses tall and narrow bottles with humans in standing pose.
  Our proposed methods do a better job at delineating overlapping object instances.
Note also that the STS++ model is able to segment out the pair of shoes in the background, labelling it of \textit{person} class.
These objects are, however, not annotated in the dataset and the model is penalised for segmenting them.
  }
  \label{fig:extradetection}
\end{figure*}

\begin{figure*}[h]
\centering
  \begin{minipage}{0.23\textwidth}
    \centering
    \mbox{}{Ground Truth}
      \includegraphics[width=0.9\textwidth]{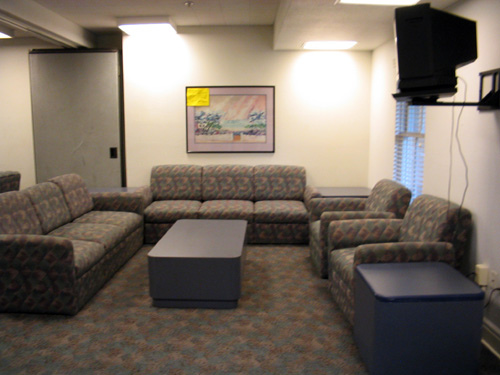}
  \end{minipage}\hspace{0.01\textwidth}
  \begin{minipage}{0.23\textwidth}
    \centering
     \mbox{}{STS original}
      \includegraphics[width=0.9\textwidth]{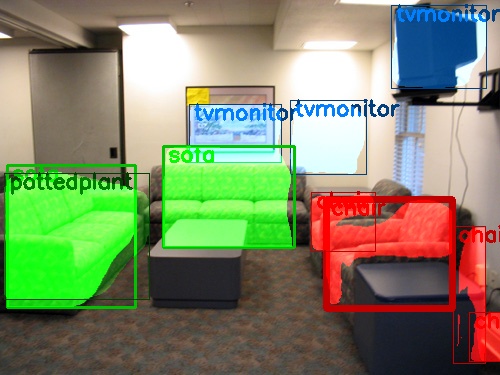}
  \end{minipage}\hspace{0.01\textwidth}
  \begin{minipage}{0.23\textwidth}
    \centering
      \mbox{}{STS (Darknet19)}
      \includegraphics[width=0.9\textwidth]{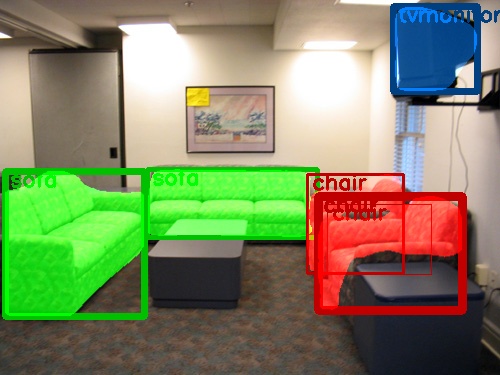}
  \end{minipage}\hspace{0.01\textwidth}
  \begin{minipage}{0.23\textwidth}
    \centering
      \mbox{}{STS++}
      \includegraphics[width=0.9\textwidth]{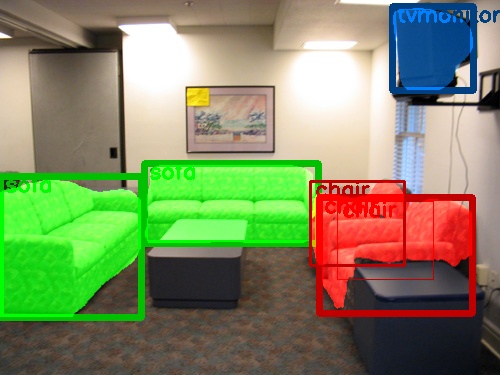}
  \end{minipage}

  \begin{minipage}{0.23\textwidth}
      \centering
    \includegraphics[width=0.9\textwidth]{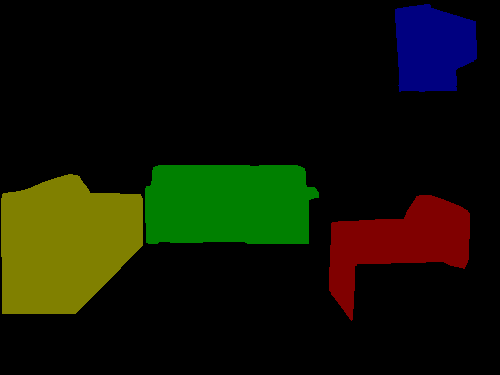}
  \end{minipage}\hspace{0.01\textwidth}
  \begin{minipage}{0.23\textwidth}
      \centering
    \includegraphics[width=0.9\textwidth]{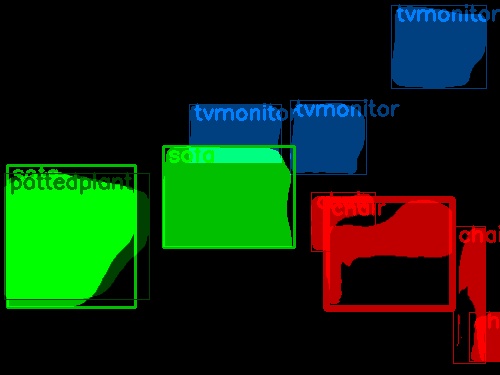}
  \end{minipage}\hspace{0.01\textwidth}
  \begin{minipage}{0.23\textwidth}
      \centering
    \includegraphics[width=0.9\textwidth]{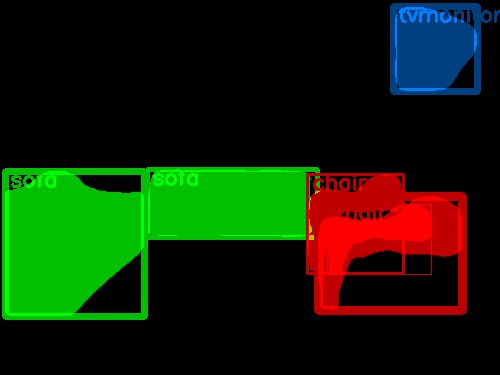}
  \end{minipage}\hspace{0.01\textwidth}
  \begin{minipage}{0.23\textwidth}
      \centering
    \includegraphics[width=0.9\textwidth]{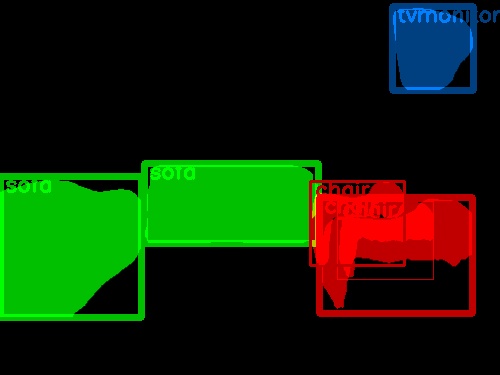}
  \end{minipage}

\vspace{2mm}

  \begin{minipage}{0.23\textwidth}
      \centering
    \includegraphics[width=0.9\textwidth]{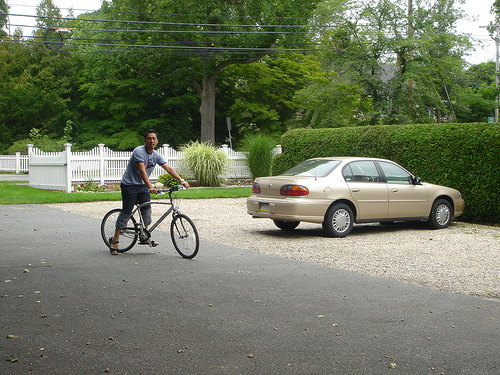}
  \end{minipage}\hspace{0.01\textwidth}
  \begin{minipage}{0.23\textwidth}
      \centering
    \includegraphics[width=0.9\textwidth]{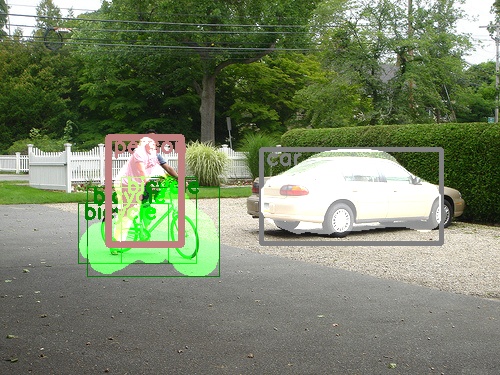}
  \end{minipage}\hspace{0.01\textwidth}
  \begin{minipage}{0.23\textwidth}
      \centering
    \includegraphics[width=0.9\textwidth]{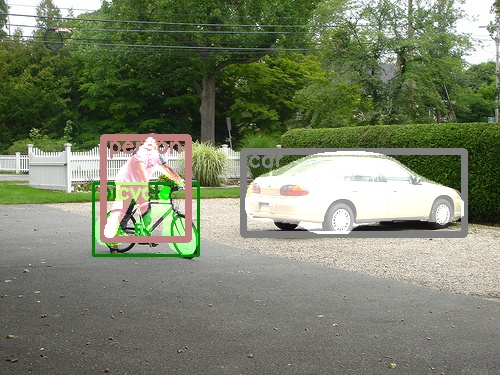}
  \end{minipage}\hspace{0.01\textwidth}
  \begin{minipage}{0.23\textwidth}
      \centering
    \includegraphics[width=0.9\textwidth]{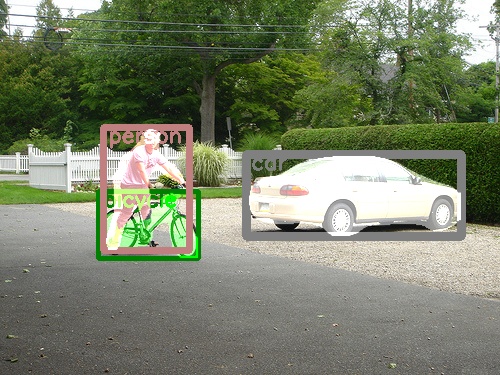}
  \end{minipage}

  \begin{minipage}{0.23\textwidth}
      \centering
    \includegraphics[width=0.9\textwidth]{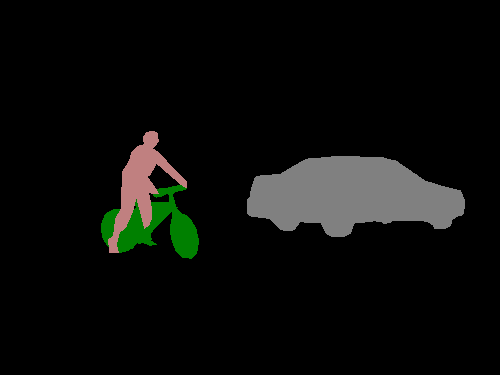}
  \end{minipage}\hspace{0.01\textwidth}
  \begin{minipage}{0.23\textwidth}
      \centering
    \includegraphics[width=0.9\textwidth]{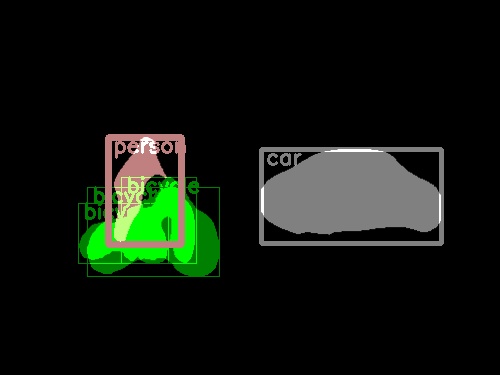}
  \end{minipage}\hspace{0.01\textwidth}
  \begin{minipage}{0.23\textwidth}
      \centering
    \includegraphics[width=0.9\textwidth]{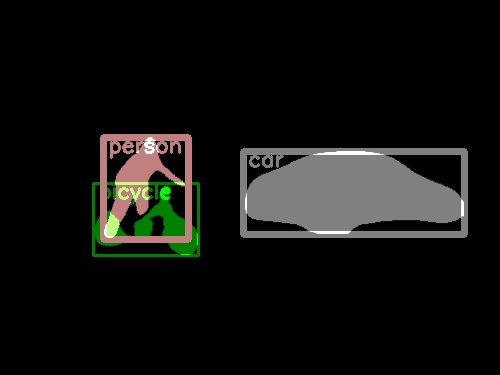}
  \end{minipage}\hspace{0.01\textwidth}
  \begin{minipage}{0.23\textwidth}
      \centering
    \includegraphics[width=0.9\textwidth]{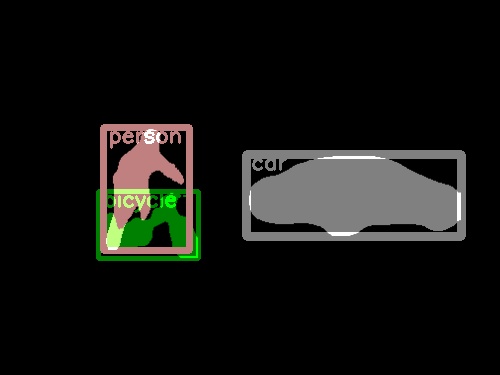}
  \end{minipage}

\vspace{2mm}

  \begin{minipage}{0.23\textwidth}
      \centering
    \includegraphics[width=0.9\textwidth]{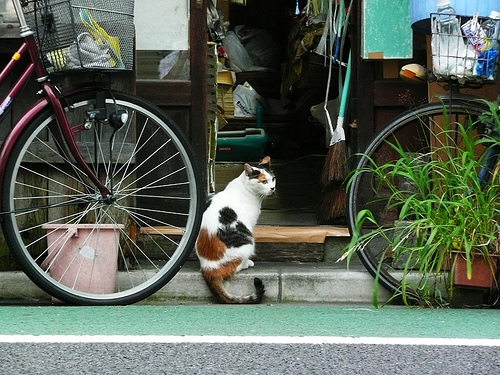}
  \end{minipage}\hspace{0.01\textwidth}
  \begin{minipage}{0.23\textwidth}
      \centering
    \includegraphics[width=0.9\textwidth]{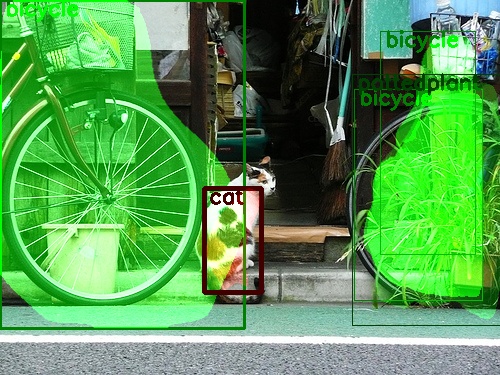}
  \end{minipage}\hspace{0.01\textwidth}
  \begin{minipage}{0.23\textwidth}
      \centering
    \includegraphics[width=0.9\textwidth]{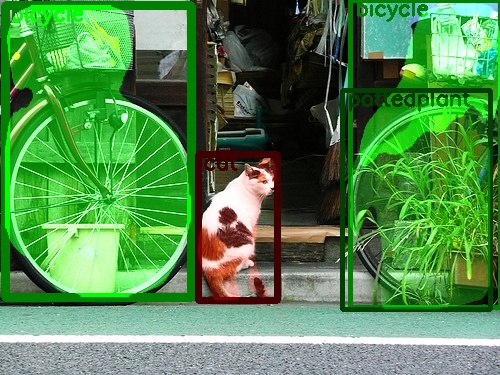}
  \end{minipage}\hspace{0.01\textwidth}
  \begin{minipage}{0.23\textwidth}
      \centering
    \includegraphics[width=0.9\textwidth]{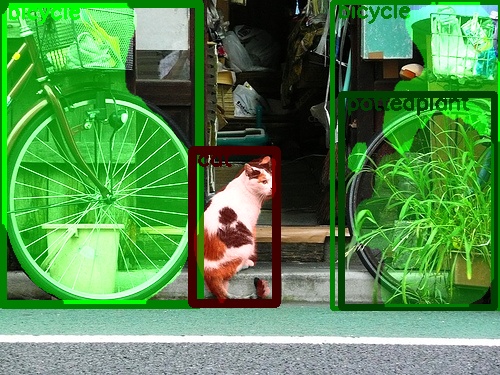}
  \end{minipage}

  \begin{minipage}{0.23\textwidth}
      \centering
    \includegraphics[width=0.9\textwidth]{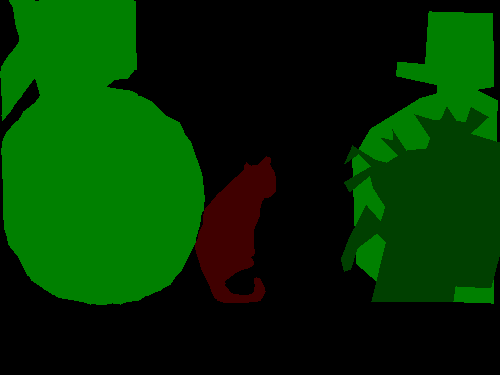}
  \end{minipage}\hspace{0.01\textwidth}
  \begin{minipage}{0.23\textwidth}
      \centering
    \includegraphics[width=0.9\textwidth]{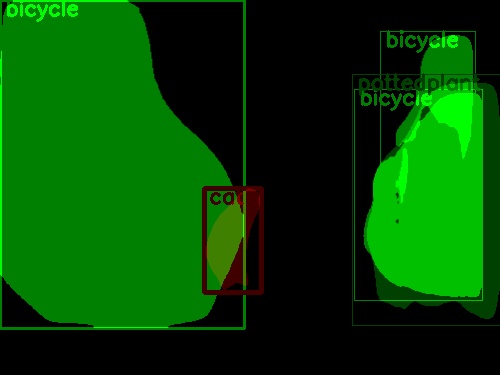}
  \end{minipage}\hspace{0.01\textwidth}
  \begin{minipage}{0.23\textwidth}
      \centering
    \includegraphics[width=0.9\textwidth]{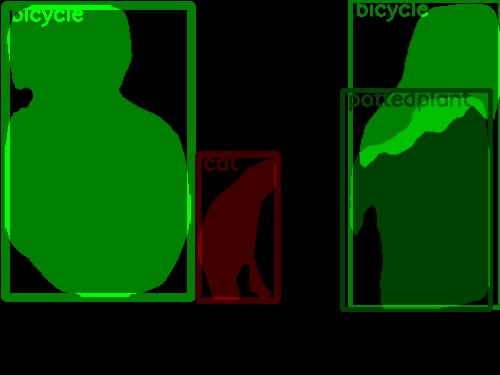}
  \end{minipage}\hspace{0.01\textwidth}
  \begin{minipage}{0.23\textwidth}
      \centering
    \includegraphics[width=0.9\textwidth]{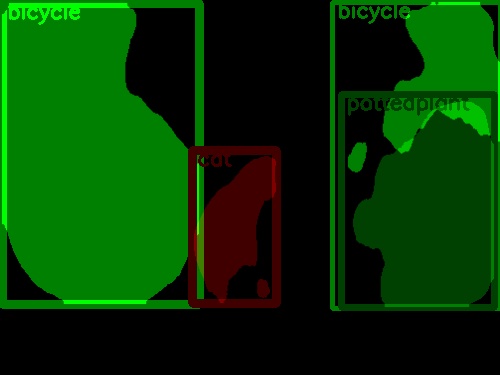}
  \end{minipage}

  \caption[More instance segmentation results.]
  {  More instance segmentation results.  }
  \label{fig-more-segs}
\end{figure*}

These examples, contained in Figures~\ref{fig:extradetection} to~\ref{fig-more-segs} demonstrate the instance segmentation capabilities and frequent drawbacks, for example, failure to model object confidence scores, detect particular instances, or predict precise object boundaries.

~\\
~\\

\begin{figure*}[h]
  \begin{minipage}{0.23\textwidth}
  \centering
    \mbox{}{Ground Truth}
    \includegraphics[width=1\textwidth]{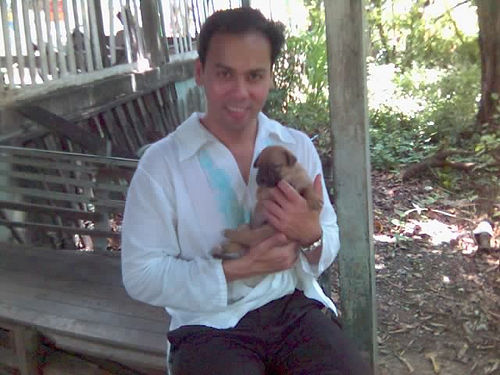}
  \end{minipage}\hspace{0.015\textwidth}
  \begin{minipage}{0.23\textwidth}
   \centering
      \mbox{}{STS original}
      \includegraphics[width=1\textwidth]{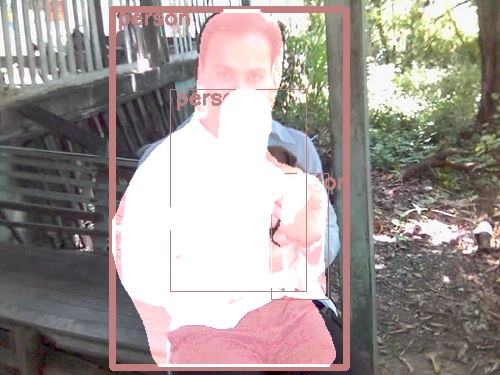}
  \end{minipage}\hspace{0.015\textwidth}
  \begin{minipage}{0.23\textwidth}
  \centering
       \mbox{}{STS (Darknet19)}
      \includegraphics[width=1\textwidth]{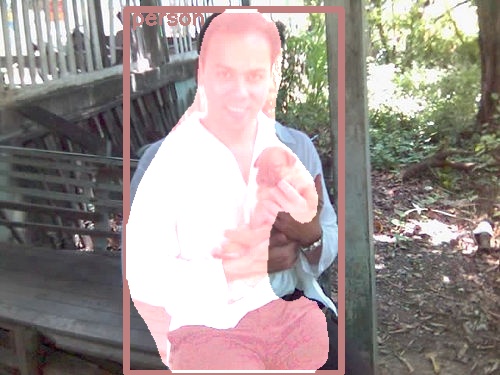}
  \end{minipage}\hspace{0.015\textwidth}
  \begin{minipage}{0.23\textwidth}
  \centering
      \mbox{}{STS++}
      \includegraphics[width=1\textwidth]{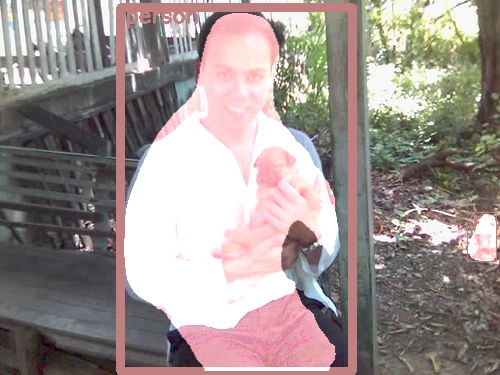}
  \end{minipage}

\vspace{2pt}

  \begin{minipage}{0.23\textwidth}
    \includegraphics[width=1\textwidth]{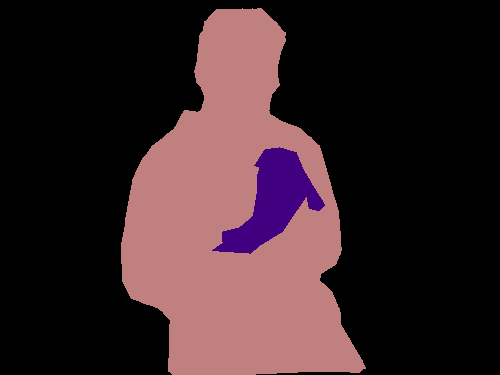}
  \end{minipage}\hspace{0.015\textwidth}
  \begin{minipage}{0.23\textwidth}
    \includegraphics[width=1\textwidth]{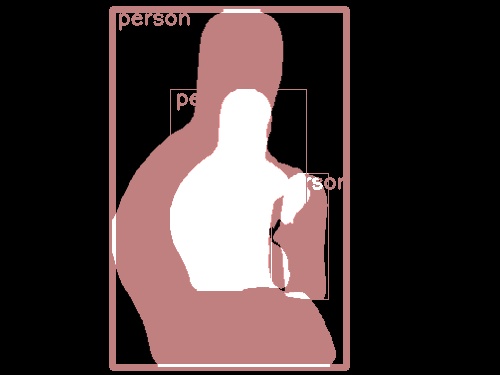}
  \end{minipage}\hspace{0.015\textwidth}
  \begin{minipage}{0.23\textwidth}
    \includegraphics[width=1\textwidth]{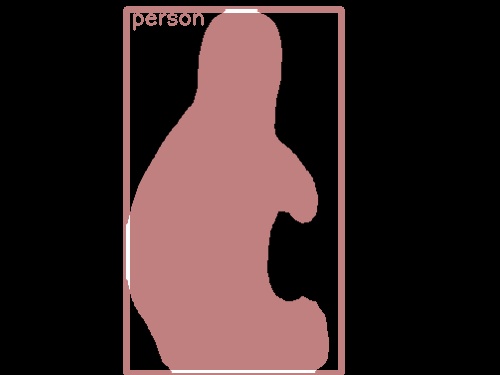}
  \end{minipage}\hspace{0.015\textwidth}
  \begin{minipage}{0.23\textwidth}
    \includegraphics[width=\textwidth]{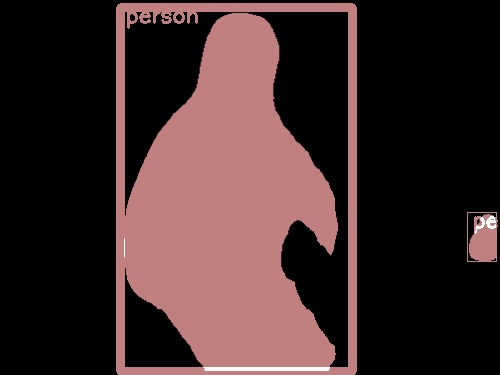}
  \end{minipage}
  \caption[Error analysis: Models fail to detect an object surrounded by another.]
  {
  The original STS model predicts two high confidence bounding boxes for the same ground truth object.
  Although our proposed models overcome this flaw, they still fail to detect the puppy in the hands of the person.
  }
\end{figure*}

\begin{figure*}[h]
  \begin{minipage}{0.23\textwidth}
    \begin{center}
     \mbox{}{Ground Truth}
    \end{center}
    \includegraphics[width=0.92\textwidth]{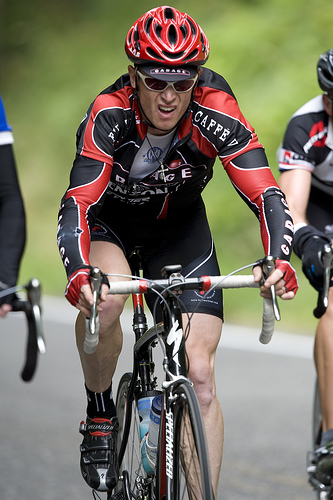}
  \end{minipage}\hspace{0.015\textwidth}
  \begin{minipage}{0.23\textwidth}
    \begin{center}
     \mbox{}{STS original}
    \end{center}
    \includegraphics[width=0.92\textwidth]{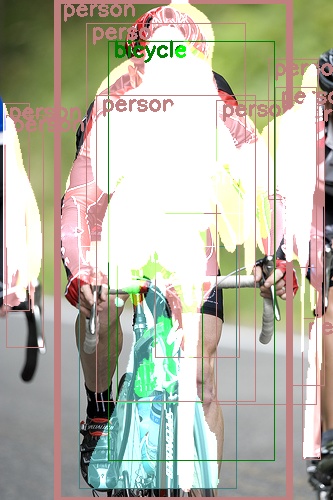}
  \end{minipage}\hspace{0.015\textwidth}
  \begin{minipage}{0.23\textwidth}
    \begin{center}
       \mbox{}{STS (Darknet19)}
    \end{center}
    \includegraphics[width=0.92\textwidth]{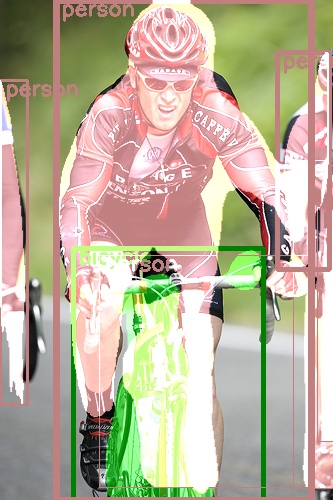}
  \end{minipage}\hspace{0.015\textwidth}
  \begin{minipage}{0.23\textwidth}
    \begin{center}
      \mbox{}{STS++}
    \end{center}
    \includegraphics[width=0.92\textwidth]{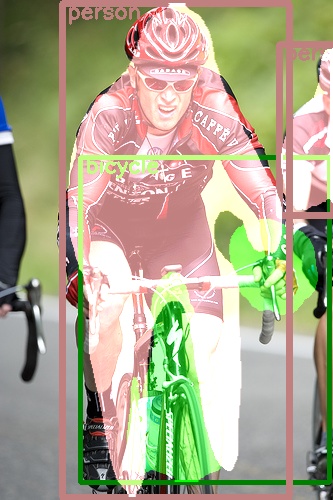}
  \end{minipage}

\vspace{2pt}

  \begin{minipage}{0.23\textwidth}
    \includegraphics[width=0.92\textwidth]{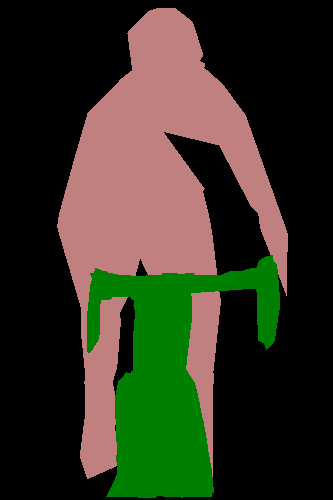}
  \end{minipage}\hspace{0.015\textwidth}
  \begin{minipage}{0.23\textwidth}
    \includegraphics[width=0.92\textwidth]{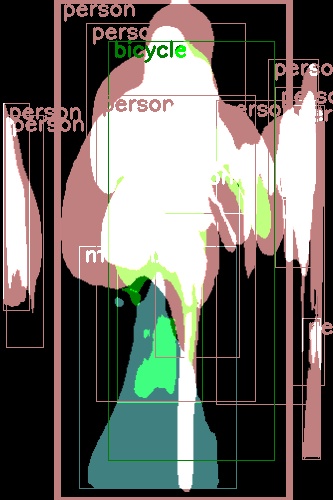}
  \end{minipage}\hspace{0.015\textwidth}
  \begin{minipage}{0.23\textwidth}
    \includegraphics[width=0.92\textwidth]{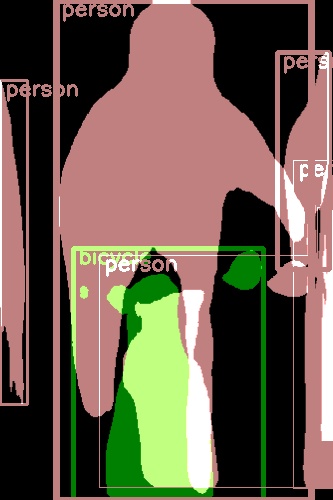}
  \end{minipage}\hspace{0.015\textwidth}
  \begin{minipage}{0.23\textwidth}
    \includegraphics[width=0.92\textwidth]{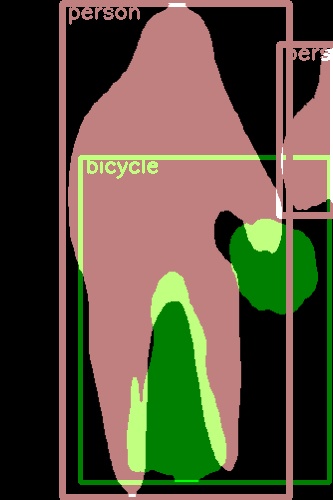}
  \end{minipage}
  \caption[Error analysis: Models struggle to capture pixel level details.]
  {
The proposed models only get the coarse location of the bicycle but do a poor job at capturing the pixel level details.
  }
\end{figure*}

\begin{figure*}[h]
  \begin{minipage}{0.23\textwidth}
    \begin{center}
      \mbox{}{Ground Truth}
      \includegraphics[width=\textwidth]{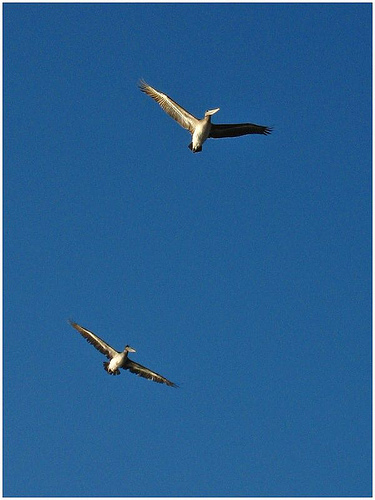}
    \end{center}
  \end{minipage}\hspace{0.01\textwidth}
  \begin{minipage}{0.23\textwidth}
    \begin{center}
     \mbox{}{STS original}
      \includegraphics[width=\textwidth]{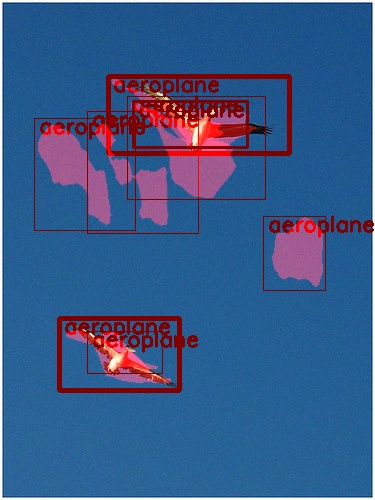}
    \end{center}
  \end{minipage}\hspace{0.01\textwidth}
  \begin{minipage}{0.23\textwidth}
    \begin{center}
      \mbox{}{STS (Darknet19)}
      \includegraphics[width=\textwidth]{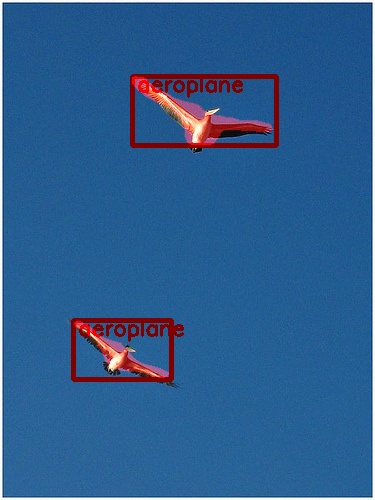}
    \end{center}
  \end{minipage}\hspace{0.01\textwidth}
  \begin{minipage}{0.23\textwidth}
    \begin{center}
      \mbox{}{STS++}
      \includegraphics[width=\textwidth]{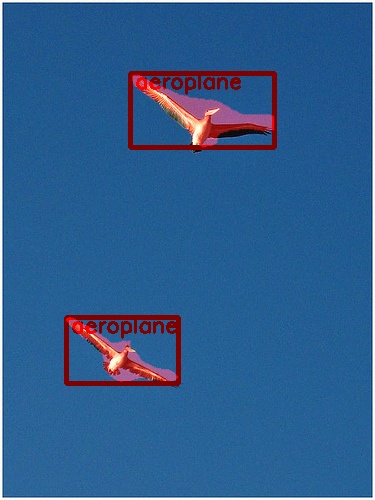}
    \end{center}
  \end{minipage}

\vspace{2pt}

  \begin{minipage}{0.23\textwidth}
    \includegraphics[width=\textwidth]{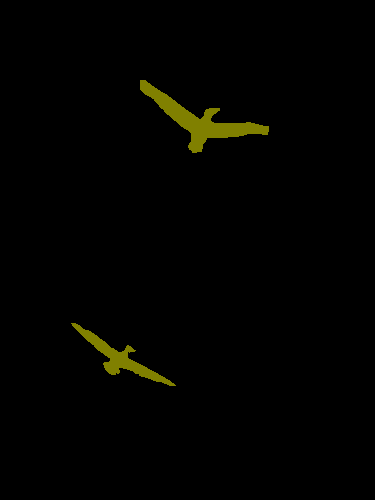}
  \end{minipage}\hspace{0.01\textwidth}
  \begin{minipage}{0.23\textwidth}
    \includegraphics[width=\textwidth]{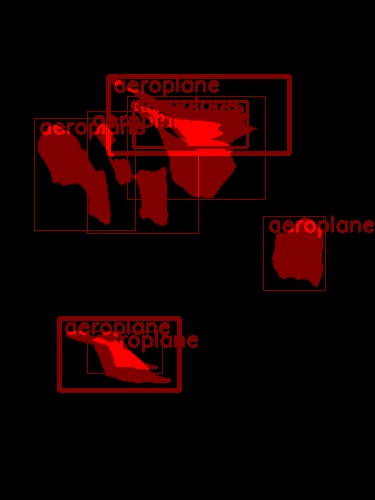}
  \end{minipage}\hspace{0.01\textwidth}
  \begin{minipage}{0.23\textwidth}
    \includegraphics[width=\textwidth]{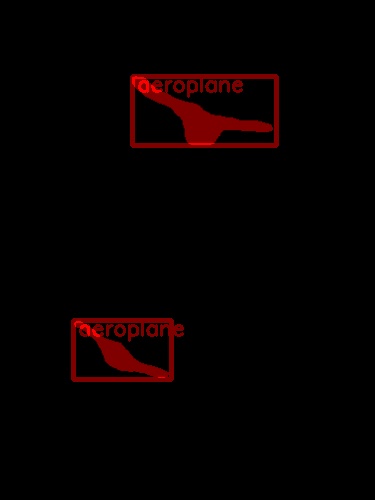}
  \end{minipage}\hspace{0.01\textwidth}
  \begin{minipage}{0.23\textwidth}
    \includegraphics[width=\textwidth]{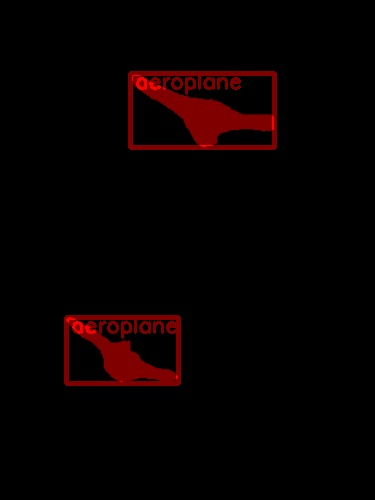}
  \end{minipage}
  \caption[Error analysis: Models fail to predict the correct category.]
  {
  The models do better at delineating object boundaries, however, fail to correctly identify the objects as birds and confuse them with aeroplanes.
  }
\end{figure*}

\begin{figure*}[h]
  \begin{minipage}{0.23\textwidth}
    \begin{center}
      \mbox{}{Ground Truth}
      \includegraphics[width=\textwidth]{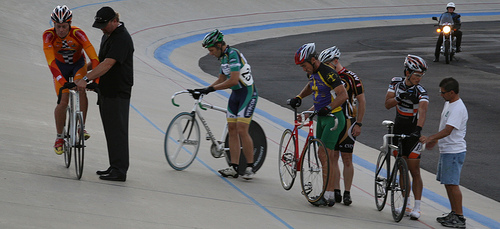}
    \end{center}
  \end{minipage}\hspace{0.015\textwidth}
  \begin{minipage}{0.23\textwidth}
    \begin{center}
     \mbox{}{STS original}
      \includegraphics[width=\textwidth]{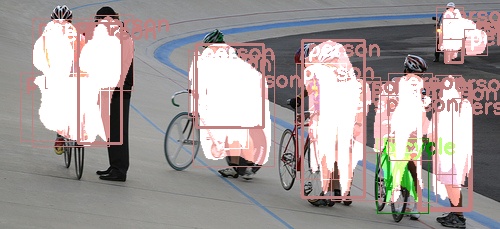}
    \end{center}
  \end{minipage}\hspace{0.015\textwidth}
  \begin{minipage}{0.23\textwidth}
    \begin{center}
       \mbox{}{STS (Darknet19)}
      \includegraphics[width=\textwidth]{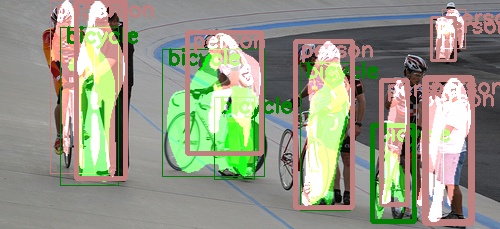}
    \end{center}
  \end{minipage}\hspace{0.015\textwidth}
  \begin{minipage}{0.23\textwidth}
    \begin{center}
      \mbox{}{STS++}
      \includegraphics[width=\textwidth]{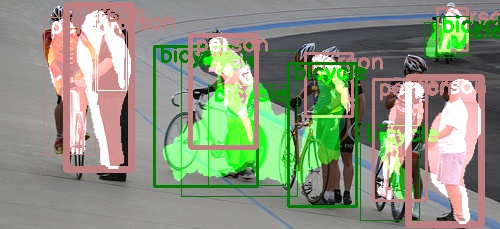}
    \end{center}
  \end{minipage}

\vspace{2pt}

  \begin{minipage}{0.23\textwidth}
    \includegraphics[width=\textwidth]{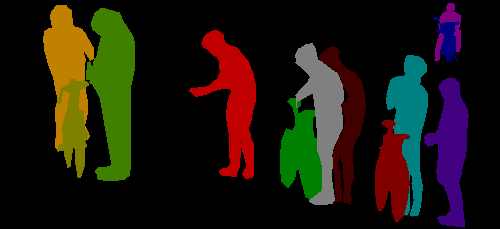}
  \end{minipage}\hspace{0.015\textwidth}
  \begin{minipage}{0.23\textwidth}
    \includegraphics[width=\textwidth]{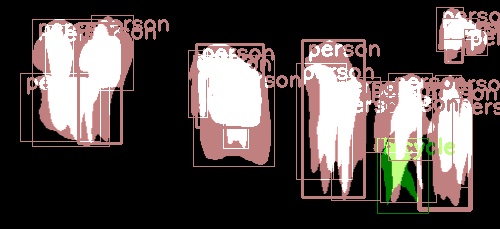}
  \end{minipage}\hspace{0.015\textwidth}
  \begin{minipage}{0.23\textwidth}
    \includegraphics[width=\textwidth]{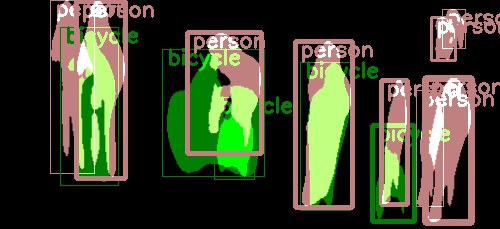}
  \end{minipage}\hspace{0.015\textwidth}
  \begin{minipage}{0.23\textwidth}
    \includegraphics[width=\textwidth]{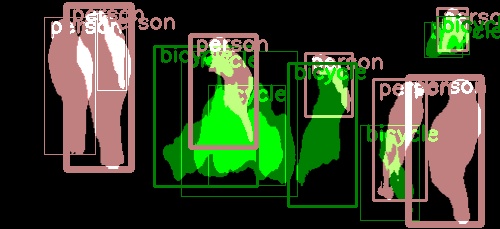}
  \end{minipage}
  \caption[Error analysis: Models fail to correctly predict object bounding boxes.]
  {
  In some locations the models output more bounding boxes than there are ground truth objects, while in other places they miss the objects entirely.
  }
\end{figure*}

\clearpage
\subsection{Detailed Error Analysis}\label{ch-errors}

\begin{table}[H]
  \begin{center}
    \resizebox{0.8\linewidth}{!}{
    \begin{tabular}{l|ccccc|cccc}
    	 & \multicolumn{5}{c|}{Prediction label} & \multicolumn{4}{c}{\% of all FP error}\\
    	Approach & Corr & Loc & Sim & Dissim & Backgr & Loc & Sim & Dissim & Backgr \\\hline
    	STS original & 58.3 & 16.6 & 12.5 &  5.9 &  6.7 & 39.9 & 29.9 & 14.1 & 16.1 \\
    	Batch Normalisation & 63.3 & 17.8 &  8.4 &  4.2 &  6.2 & 48.4 & 23.0 & 11.6 & 17.0 \\
    	Data Augmentation & 66.2 & 16.3 &  8.0 &  4.0 &  5.5 & 48.2 & 23.7 & 11.9 & 16.2 \\
    	Shared Predictor & 64.3 & 16.3 &  8.6 &  4.9 &  5.9 & 45.8 & 24.0 & 13.8 & 16.5 \\
    	Anchor Boxes & 69.9 & 10.1 &  8.8 &  3.9 &  7.3 & 33.6 & 29.1 & 13.1 & 24.2 \\
    	Darknet19 & 73.0 &  9.4 &  7.8 &  2.6 &  7.1 & 35.0 & 29.1 &  9.6 & 26.2 \\
    	End-to-End & 73.6 &  9.2 &  7.3 &  3.1 &  6.8 & 35.0 & 27.7 & 11.6 & 25.6 \\
    	Large Decoder & 73.1 &  9.4 &  7.2 &  3.3 &  7.1 & 34.7 & 26.6 & 12.3 & 26.3 \\
    	Distance Transform (STS++) & 73.7 &  9.0 &  7.1 &  3.0 &  7.1 & 34.3 & 27.0 & 11.5 & 27.2 \\
    \hline\end{tabular}
    }
  \end{center}
  \caption[Detection errors of proposed models on Pascal VOC 2007 test.]
  {
  Detection errors of our proposed models on Pascal VOC 2007 test dataset. The errors are grouped as per the methodology presented in~\cite{DBLP:conf/eccv/HoiemCD12}.
  }
  \label{tbl-detection-errors}
\end{table}

\begin{table}[H]
  \begin{center}
    \resizebox{1\linewidth}{!}{
    \begin{tabular}{l|c|cccccccccc}
    	STS original & Overall & aeroplane & bicycle & bird & boat & bottle & bus & car & cat & chair & cow\\\hline
    	TP: correct & 58.3  & 59.2 & 61.7 & 54.5 & 57.0 & 39.6 & 61.0 & 65.9 & 66.2 & 58.8 & 63.2\\
    	FP: localisation & 16.6  & 23.8 & 15.9 & 20.5 & 23.7 & 19.0 &  7.9 & 19.1 &  7.6 & 14.0 & 15.2\\
    	FP: similar & 12.5  & 11.6 & 12.3 & 10.4 &  7.9 &  0.0 & 24.0 &  7.1 & 23.8 &  7.5 & 19.5\\
    	FP: dissimilar &  5.9  &  1.0 &  8.7 &  4.0 &  2.5 & 20.9 &  0.8 &  2.2 &  0.8 &  9.0 &  0.0\\
    	FP: background &  6.7  &  4.5 &  1.3 & 10.6 &  8.9 & 20.5 &  6.3 &  5.6 &  1.6 & 10.6 &  2.1\\\hline\hline
    	 &  & dining table & dog & horse & motorbike & person & potted plant & sheep & sofa & train & tv monitor\\\hline
    	TP: correct & 58.3  & 53.5 & 62.3 & 60.8 & 59.3 & 56.6 & 37.0 & 61.4 & 68.7 & 65.9 & 53.7\\
    	FP: localisation & 16.6  & 15.1 &  6.6 & 13.7 & 14.1 & 36.1 & 28.5 & 15.4 &  7.3 & 12.6 & 16.3\\
    	FP: similar & 12.5  & 10.4 & 29.6 & 22.0 & 17.1 &  1.9 &  0.0 & 22.2 &  7.8 & 14.2 &  0.0\\
    	FP: dissimilar &  5.9  & 12.4 &  0.2 &  1.8 &  8.7 &  2.5 & 14.5 &  0.0 & 12.6 &  0.7 & 14.4\\
    	FP: background &  6.7  &  8.7 &  1.3 &  1.8 &  0.8 &  2.9 & 19.9 &  1.0 &  3.5 &  6.6 & 15.5\\\hline
    \end{tabular}
    }
  \end{center}
  \caption[Category-level detection errors of STS model on Pascal VOC 2007 test]
  {
  Category-level detection errors of the original STS model on Pascal VOC 2007 test dataset.
  }
  \label{tbl-detection-errors-cat-sts-orig}
\end{table}

\begin{table}[H]
  \begin{center}
    \resizebox{1\linewidth}{!}{
    \begin{tabular}{l|c|cccccccccc}
    	Darknet19 & Overall & aeroplane & bicycle & bird & boat & bottle & bus & car & cat & chair & cow\\\hline
    	TP: correct & 73.0  & 70.1 & 71.0 & 73.1 & 68.2 & 49.8 & 76.4 & 79.4 & 78.9 & 72.9 & 79.3\\
    	FP: localisation &  9.4  & 11.9 & 13.6 & 14.6 & 17.3 & 17.2 &  3.9 &  9.3 &  6.2 &  9.9 &  5.8\\
    	FP: similar &  7.8  &  7.7 &  9.0 &  7.5 &  5.6 &  0.0 & 12.2 &  3.4 & 14.6 &  4.4 & 14.0\\
    	FP: dissimilar &  2.6  &  1.9 &  4.1 &  0.3 &  0.8 &  8.1 &  0.4 &  1.0 &  0.0 &  3.5 &  0.0\\
    	FP: background &  7.1  &  8.4 &  2.3 &  4.5 &  8.1 & 25.0 &  7.1 &  6.7 &  0.3 &  9.3 &  0.9\\\hline\hline
    	 &  & dining table & dog & horse & motorbike & person & potted plant & sheep & sofa & train & tv monitor\\\hline
    	TP: correct & 73.0  & 77.6 & 75.7 & 79.2 & 75.6 & 71.4 & 49.0 & 75.9 & 84.6 & 80.8 & 71.7\\
    	FP: localisation &  9.4  &  3.0 &  5.1 &  5.6 &  7.6 & 20.6 & 20.1 &  4.5 &  1.3 &  5.6 &  5.8\\
    	FP: similar &  7.8  &  3.7 & 17.9 & 14.4 &  8.9 &  1.4 &  0.0 & 16.7 &  6.6 &  8.9 &  0.0\\
    	FP: dissimilar &  2.6  &  8.4 &  0.0 &  0.3 &  3.8 &  1.6 &  6.2 &  0.3 &  4.8 &  0.3 &  6.1\\
    	FP: background &  7.1  &  7.4 &  1.3 &  0.5 &  4.1 &  5.0 & 24.7 &  2.6 &  2.8 &  4.3 & 16.3\\\hline
    \end{tabular}
    }
  \end{center}
  \caption[Category-level detection errors of Darknet19 on Pascal VOC 2007 test]
  {
  Category-level detection errors of our proposed Darknet19 model on Pascal VOC 2007 test dataset.
  }
  \label{tbl-detection-errors-cat-dn19}
\end{table}

\begin{table}[H]
  \begin{center}
    \resizebox{1\linewidth}{!}{
    \begin{tabular}{l|c|cccccccccc}
    	Large Decoder & Overall & aeroplane & bicycle & bird & boat & bottle & bus & car & cat & chair & cow\\\hline
    	TP: correct & 73.1  & 70.7 & 74.0 & 70.7 & 67.7 & 50.7 & 78.3 & 79.4 & 80.0 & 73.2 & 77.5\\
    	FP: localisation &  9.4  & 14.1 & 10.0 & 16.0 & 15.8 & 16.9 &  3.1 & 10.3 &  7.0 &  9.6 &  4.9\\
    	FP: similar &  7.2  &  6.4 &  6.7 &  8.2 &  4.1 &  0.0 &  9.1 &  2.7 & 12.2 &  4.3 & 16.1\\
    	FP: dissimilar &  3.3  &  2.6 &  4.4 &  0.5 &  1.8 &  9.7 &  1.6 &  0.9 &  0.5 &  4.5 &  0.0\\
    	FP: background &  7.1  &  6.1 &  4.9 &  4.7 & 10.7 & 22.7 &  7.9 &  6.6 &  0.3 &  8.4 &  1.5\\\hline\hline
    	 &  & dining table & dog & horse & motorbike & person & potted plant & sheep & sofa & train & tv monitor\\\hline
    	TP: correct & 73.1  & 78.9 & 77.5 & 78.2 & 74.0 & 71.5 & 49.3 & 75.9 & 83.8 & 77.5 & 72.0\\
    	FP: localisation &  9.4  &  3.3 &  4.5 &  6.3 &  8.7 & 21.1 & 15.7 &  6.8 &  2.0 &  6.3 &  4.7\\
    	FP: similar &  7.2  &  3.0 & 16.2 & 14.7 &  8.9 &  1.5 &  0.0 & 14.8 &  5.3 &  9.3 &  0.0\\
    	FP: dissimilar &  3.3  &  7.0 &  0.2 &  0.0 &  6.5 &  1.4 & 10.8 &  0.3 &  6.8 &  0.7 &  6.1\\
    	FP: background &  7.1  &  7.7 &  1.5 &  0.8 &  1.9 &  4.5 & 24.2 &  2.3 &  2.0 &  6.3 & 17.2\\\hline
    \end{tabular}
    }
  \end{center}
  \caption[Category-level detection errors of Large Decoder on Pascal VOC 2007 test.]
  {
  Category-level detection errors of our proposed Large Decoder model on Pascal VOC 2007 test dataset.
  }
  \label{tbl-detection-errors-cat-e2e-large}
\end{table}

\begin{table}[H]
  \begin{center}
    \resizebox{1\linewidth}{!}{
    \begin{tabular}{l|c|cccccccccc}
    	Approach & $mAP^r_{0.5}$ & aeroplane & bicycle & bird & boat & bottle & bus & car & cat & chair & cow\\\hline
    	STS original & 33.5 & 58.6 & 33.7 & 31.2 & 17.6 & 11.0 & 65.5 & 37.0 & 62.6 &  6.2 & 26.1\\
    	Batch Normalisation & 38.6 & 58.9 & 39.3 & 38.2 & 18.6 & 15.1 & 67.9 & 41.4 & 70.6 &  7.6 & 34.6\\
    	Data Augmentation & 42.3 & 64.7 & 38.4 & 41.0 & 23.6 & 14.5 & 71.0 & 42.4 & 74.3 &  8.7 & 44.4\\
    	Shared Predictor & 41.4 & 63.0 & 42.2 & 41.9 & 22.4 & 14.8 & 69.3 & 39.8 & 73.9 &  8.1 & 40.3\\
    	Anchors & 48.5 & 69.6 & 47.0 & 51.9 & 29.9 & 22.9 & 71.9 & 50.5 & 77.5 & 14.9 & 49.3\\
    	Darknet19 & 52.2 & 72.0 & 52.7 & 54.6 & 30.6 & \textbf{28.0} & 74.8 & \textbf{56.8} & 81.8 & 21.1 & \textbf{56.9}\\
    	End-to-End & 51.7 & 65.3 & 52.6 & 54.0 & 31.3 & 26.5 & \textbf{77.7} & 55.3 & 80.7 & 19.0 & 55.4\\
    	Large Decoder & 52.3 & \textbf{73.0} & 52.5 & 55.1 & \textbf{34.8} & 24.9 & 75.3 & 55.3 & \textbf{82.0} & 20.7 & 54.6\\
    	Distance Transform (STS++) & \textbf{53.2} & 72.9 & \textbf{53.6} & \textbf{58.5} & 32.4 & 25.8 & 74.9 & 56.5 & 81.8 & \textbf{22.8} & 55.0\\
    \hline
    	SDS~\citep{Hariharan2014} & 49.7 & 68.4 & 49.4 & 52.1 & 32.8 & 33.0 & 67.8 & 53.6 & 73.9 & 19.9 & 43.7\\
    	\cite{2017cvpr_aarnab} & 62.0 & 80.3 & 52.8 & 68.5 & 47.4 & 39.5 & 79.1 & 61.5 & 87.0 & 28.1 & 68.3\\\hline
    \hline 	 &  & dining table & dog & horse & motorbike & person & potted plant & sheep & sofa & train & tv monitor\\\hline
    	STS original & 33.5 & 13.5 & 49.7 & 31.0 & 37.9 & 37.7 &  7.0 & 31.2 & 20.0 & 62.7 & 29.2\\
    	Batch Normalisation & 38.6 & 11.9 & 58.6 & 45.5 & 45.6 & 36.4 & 14.1 & 34.4 & 23.9 & 69.8 & 39.8\\
    	Data Augmentation & 42.3 & 15.9 & 64.1 & 47.1 & 53.1 & 40.9 & 15.7 & 44.1 & 27.0 & 72.0 & 43.1\\
    	Shared Predictor & 41.4 & 19.3 & 60.4 & 50.3 & 51.2 & 37.5 & 13.4 & 41.0 & 28.1 & 69.0 & 41.9\\
    	Anchors & 48.5 & 24.2 & 68.4 & 55.0 & 59.7 & 49.7 & 18.6 & 53.7 & 31.1 & 74.5 & 50.5\\
    	Darknet19 & 52.2 & 26.0 & 71.4 & 61.6 & \textbf{62.4} & 54.4 & 19.4 & 53.2 & 36.2 & 76.1 & 54.5\\
    	End-to-End & 51.7 & 25.1 & 72.9 & 57.9 & 58.6 & 51.9 & 22.5 & \textbf{56.6} & \textbf{37.5} & 77.9 & 56.2\\
    	Large Decoder & 52.3 & 25.2 & 71.9 & \textbf{64.8} & 58.6 & 54.0 & 19.6 & 54.9 & 36.5 & 76.9 & 55.7\\
    	Distance Transform (STS++)& \textbf{53.2} & \textbf{26.2} & \textbf{74.1} & 62.5 & 59.8 & \textbf{57.7} & \textbf{22.7} & 56.1 & 36.4 & \textbf{78.5} & \textbf{56.4}\\
    \hline
    	SDS~\citep{Hariharan2014} & 49.7 & 25.7 & 60.6 & 55.9 & 58.9 & 56.7 & 28.5 & 55.6 & 32.1 & 64.7 & 60.0\\
    	\cite{2017cvpr_aarnab} & 62.0 & 35.5 & 86.1 & 73.9 & 66.1 & 63.8 & 32.9 & 65.3 & 50.4 & 81.4 & 71.4\\\hline
    \end{tabular}
    }
  \end{center}
  \caption[Comparison of average precision scores on SBD val at IoU of $0.5$.]
  {
  Average precision ($AP^r_{0.5}$) estimates for the task of detection on SBD validation set for all the proposed methods compared with state-of-the-art models.
  }
  \label{tbl-dlt-errors-05}
\end{table}

\begin{table}[H]
  \begin{center}
    \resizebox{1\linewidth}{!}{
    \begin{tabular}{l|c|cccccccccc}
    	Approach & $mAP^r_{0.7}$ & aeroplane & bicycle & bird & boat & bottle & bus & car & cat & chair & cow\\\hline
    	STS original & 14.9 & 20.2 & 10.9 &  7.8 &  5.2 &  5.5 & 52.3 & 20.8 & 41.3 &  0.6 &  7.5\\
    	Batch Normalisation & 17.4 & 22.4 & 13.5 & 12.3 &  5.3 &  6.0 & 55.5 & 21.6 & 46.8 &  0.4 & 11.4\\
    	Data Augmentation & 20.8 & 27.9 & 11.9 & 14.2 &  8.8 &  6.9 & 57.3 & 25.5 & 53.4 &  0.8 & 15.9\\
    	Shared Predictor & 20.5 & 28.5 & 15.2 & 15.0 &  9.6 &  7.0 & 55.5 & 24.1 & 51.1 &  0.8 & 17.9\\
    	Anchors & 26.7 & 30.2 & 19.1 & 18.8 & 10.9 & 11.4 & 65.0 & 35.9 & 61.0 &  2.2 & 24.8\\
    	Darknet19 & 31.1 & 40.3 & \textbf{24.9} & 22.7 & 11.6 & 14.5 & 66.7 & 39.5 & 67.2 & \textbf{ 5.9} & 26.3\\
    	End-to-End & 27.9 & 23.4 & 19.3 & 18.6 & 13.8 & 13.4 & \textbf{70.3} & 39.8 & 60.4 &  3.2 & 23.3\\
    	Large Decoder & 31.9 & \textbf{42.3} & 22.9 & 24.5 & \textbf{18.1} & \textbf{15.2} & 67.5 & 40.3 & \textbf{68.8} &  5.8 & 31.1\\
    	Distance Transform (STS++)& \textbf{32.3} & 39.1 & 23.8 & \textbf{24.9} & 15.8 & 12.8 & 65.4 & \textbf{41.3} & 66.8 &  5.6 & \textbf{31.3}\\
    \hline
    	\cite{2017cvpr_aarnab} & 44.8 & 69.0 & 27.4 & 52.7 & 26.4 & 22.4 & 70.3 & 46.0 & 74.7 &  9.6 & 46.8\\\hline
    \hline 	 &  & dining table & dog & horse & motorbike & person & potted plant & sheep & sofa & train & tv monitor\\\hline
    	STS original & 14.9 &  2.0 & 23.7 &  3.9 & 13.7 &  8.9 &  1.4 &  5.5 &  9.7 & 45.8 & 11.0\\
    	Batch Normalisation & 17.4 &  2.1 & 24.9 &  7.4 & 15.9 &  8.4 &  2.2 &  9.0 & 12.4 & 55.6 & 14.6\\
    	Data Augmentation & 20.8 &  3.3 & 33.6 & 10.4 & 21.8 & 11.2 &  3.7 & 11.9 & 15.2 & 57.8 & 23.7\\
    	Shared Predictor & 20.5 &  5.9 & 33.2 & 12.0 & 22.6 & 11.5 &  3.0 & 10.0 & 13.9 & 54.3 & 19.4\\
    	Anchors & 26.7 &  7.9 & 42.5 & 17.9 & 26.7 & 17.9 &  4.3 & 19.9 & 19.2 & 63.8 & 33.8\\
    	Darknet19 & 31.1 & 10.5 & 48.6 & \textbf{24.3} & 35.1 & 23.7 &  4.6 & 26.8 & 24.5 & 64.8 & 40.5\\
    	End-to-End & 27.9 & 11.7 & 39.6 & 14.5 & 25.0 & 18.8 & \textbf{ 6.6} & 24.0 & \textbf{29.1} & 61.6 & 40.6\\
    	Large Decoder & 31.9 &  8.5 & 48.8 & 22.1 & \textbf{35.8} & 24.7 &  5.7 & 25.8 & 25.4 & 63.1 & \textbf{41.9}\\
    	Distance Transform (STS++)& \textbf{32.3} & \textbf{13.2} & \textbf{50.8} & 22.6 & 35.7 & \textbf{27.4} &  6.4 & \textbf{29.0} & 26.8 & \textbf{65.4} & 41.0\\
    \hline
    	\cite{2017cvpr_aarnab} & 44.8 & 16.9 & 71.6 & 48.4 & 46.3 & 40.3 & 14.8 & 47.6 & 36.5 & 69.7 & 58.2\\\hline
    \end{tabular}
    }
  \end{center}
  \caption[Comparison of average precision scores on SBD val at IoU of $0.7$.]
  {
  Average precision ($AP^r_{0.7}$) estimates for the task of detection on SBD validation set for all the proposed methods compared with state-of-the-art models.
  }
  \label{tbl-dlt-errors-07}
\end{table}

\end{subappendices}

\end{document}